%% file: soumission.tex
\RequirePackage{fix-cm}

\documentclass[twocolumn]{svjour3}

\smartqed

\usepackage{array}
\usepackage{hhline}
\usepackage{multirow}
\usepackage{color}
\usepackage{amsmath}
\usepackage{amssymb}
\usepackage{graphics}
\graphicspath{{FIGURES/},{FIGURES/},{FIGURES/},{FIGURES/SV/},{FIGURES/SV_P/},{FIGURES/MO/},{FIGURES/CT/}}
\usepackage{epsfig}
\usepackage{graphicx}
\usepackage{soul}

\usepackage[pagebackref=true,breaklinks=false,letterpaper=true,colorlinks,bookmarks=false]{hyperref}

\usepackage{import}
\usepackage{diagbox}
\usepackage{bm}

\let\oldFootnote\footnote
\newcommand\nextToken\relax

\renewcommand\footnote[1]{%
    \oldFootnote{#1}\futurelet\nextToken\isFootnote}

\newcommand\isFootnote{%
    \ifx\footnote\nextToken\textsuperscript{,}\fi}

\usepackage{ifthen}
\newboolean{ABR}
\setboolean{ABR}{false}
\newboolean{FTR}
\setboolean{FTR}{false}

\usepackage{mathtools}

\def\*#1{\mathbf{#1}}

\newcommand{\off}[1]{}

\newenvironment{itemizepoint}{%
 \begin{itemize}}{\end{itemize}}

\title{{Variational Methods for Normal Integration}}

\author{Yvain Qu\'eau \and Jean-Denis Durou \and Jean-Fran\c cois Aujol}

\institute{Y. Qu\'eau \at
	Technical University Munich, Germany \\
	\email{yvain.queau@tum.de}
	\and 
	J.-D. Durou \at
	IRIT, Universit\'{e} de Toulouse, France
	\and
	J.-F. Aujol \at
	IMB, Universit\'{e} de Bordeaux, Talence, France \\
	Institut Universitaire de France
}

\begin{document}

\maketitle

\begin{abstract}
The need for an efficient method of integration of a dense normal field is inspired by several computer vision tasks, such as shape-from-shading, photometric stereo, deflectometry, etc. 
Inspired by edge-preserving methods from image processing, {we study in this paper several variational approaches for normal integration, with a focus on non-rectangular domains, free boundary and depth discontinuities.}
We first introduce a new discretization for quadratic integration, which is designed to ensure both fast recovery and the ability to handle non-rectangular domains {with a free boundary}. Yet, with this solver, discontinuous surfaces can be handled only if the scene is first segmented into pieces without discontinuity. 
{Hence, we then discuss} several discontinuity-preserving {strategies. Those inspired, respectively, by the Mumford-Shah segmentation method and by ani\-so\-tro\-pic diffusion, are shown}
to be the most effective for recovering discontinuities.
\end{abstract}



\section{Introduction}
\label{sec:5}

In this paper, we study several methods for numerical integration of a gradient field over a 2D-grid. Our aim is to estimate the values of a function $z:\,\mathbb{R}^2 \to \mathbb{R}$, over a set $\Omega \subset \mathbb{R}^2$ {(reconstruction domain)} where an estimate $\mathbf{g} = [p,q]^\top:\,\Omega \to \mathbb{R}^2$ of its gradient $\nabla z$ is available. Formally, we want to {solve} the following equation in the unknown \emph{depth map} $z$:
\begin{equation}
	\nabla z(u,v) = \underbrace{\left[p(u,v),q(u,v) \right]^\top}_{\mathbf{g}(u,v)},\, \forall (u,v) \in \Omega
\label{eq:1}
\end{equation}


In {a companion} survey paper~\cite{Durou:2016a}, we have shown that an ideal numerical tool for solving Equation~\eqref{eq:1} should satisfy the following properties, appart accuracy:
\begin{itemizepoint}
	\item $\mathcal{P}_{\text{Fast}}$: be as \emph{fast} as possible;

	\item $\mathcal{P}_{\text{Robust}}$: be \emph{robust} to a noisy gradient field;

	\item $\mathcal{P}_{\text{FreeB}}$: be able to handle a \emph{free boundary};

	\item $\mathcal{P}_{\text{Disc}}$: preserve the \emph{depth discontinuities};

	\item $\mathcal{P}_{\text{NoRect}}$: be able to work on a \emph{non-rectangular} domain $\Omega$;

	\item $\mathcal{P}_{\text{NoPar}}$: have \emph{no critical parameter} to tune.
\end{itemizepoint}

\paragraph{{Contributions.}}

{
This paper builds upon the previous conference papers~\cite{Durou:2009a,Durou:2007a,Queau:2015b} to clarify the building blocks of variational approaches to the integration problem, with a view to meeting the largest subset of these requirements. As discussed in Section~\ref{sec:variational_methods}, the variational framework is well-adapted to this task, thanks to its flexibility. However, these properties are difficult, if not impossible, to satisfy simultaneously. In particular, $\mathcal{P}_{\text{Disc}}$ seems hardly compatible with $\mathcal{P}_{\text{Fast}}$ and $\mathcal{P}_{\text{NoPar}}$.  
}

{
Therefore, we first focus in Section~\ref{sec:smooth} on the properties $\mathcal{P}_{\text{FreeB}}$ and $\mathcal{P}_{\text{NoRect}}$. A new discretization strategy for normal integration is presented, which is independent from the shape of the domain and assumes no particular boundary condition. When used within a quadratic variational approach, this discretization strategy allows to ensure all the desired properties except $\mathcal{P}_{\text{Disc}}$. In particular, the numerical solution comes down to solving a symmetric, diagonally dominant linear system, which can be achieved very efficiently using preconditioning techniques. In comparison with our previous work~\cite{Durou:2007a} which considered only forward finite differences and standard Jacobi iterations, the properties  $\mathcal{P}_{\text{Robust}}$ and $\mathcal{P}_{\text{Fast}}$  are better satisfied.
}

{In Section~\ref{sec:nonsmooth}, we focus more specifically on the integration problem in the presence of discontinuities. Several variations of well-known models from image processing are empirically compared, while suggesting for each of them the appropriate state-of-the-art minimization method. Besides the approaches based on total variation and non-convex regularization, which we already presented, respectively, in~\cite{Queau:2015b} and~\cite{Durou:2009a}, two new methods inspired by the Mumford-Shah segmentation method and by anisotropic diffusion are introduced. They are shown to be particularly effective for handling $\mathcal{P}_{\text{Disc}}$, although $\mathcal{P}_{\text{Fast}}$ and $\mathcal{P}_{\text{NoPar}}$ are lost.}

{These variational methods for normal integration are based on the same variational framework, which is detailed in the next section.}


\section{{From Variational Image Restoration to Variational Normal Integration}}
\label{sec:variational_methods}

In view of the $\mathcal{P}_{\text{Robust}}$ property, variational methods, which aim at estimating the surface by minimization of a well-chosen criterion, are particularly suited for the integration problem. Hence, we choose the variational framework as basis for the design of new methods. This choice is also motivated by the fact that the property which is the most difficult to ensure is probably $\mathcal{P}_{\text{Disc}}$. Numerous variational methods have been designed for edge-preserving image processing: such methods may thus be a natural source of inspiration for designing discontinuity-preserving integration methods.

\subsection{Variational Methods in Image Processing}

For a comprehensive introduction to this literature, we refer the reader to \cite{AubertKornprobst} and to pioneering papers such as \cite{Catte_etal,Charbonnier,Kornprobst_Aubert,Nikolova}. Basically, the idea in edge-preserving image restoration is that edges need to be processed in a particular way. This is usually achieved by choosing an appropriate energy to minimize, formulating the inverse problem as the recovery of a restored image $z:\,\Omega \subset \mathbb{R}^2 \to \mathbb{R}$ minimizing the energy:
\begin{equation}
	\mathcal{E}(z) = \mathcal{F}(z) + \mathcal{R}(z)
\label{eq:Ez}
\end{equation}
where:
\begin{itemizepoint}
	\item $\mathcal{F}(z)$ is a \emph{fidelity} term penalizing the difference between a corrupted image $z^0$ and the restored image:
	\begin{equation}
		\mathcal{F}(z) = \iint\displaylimits_{(u,v) \in \Omega} \mathrm{\Phi}\left( z(u,v) - z^0(u,v)  \right) \mathrm{d}u\,\mathrm{d}v
		\label{eq:4bis}
	\end{equation}
\item $\mathcal{R}(z)$ is a \emph{regularization} term, which usually penalizes the gradient of the restored image:
	\begin{equation}
		\mathcal{R}(z) = \iint\displaylimits_{(u,v) \in \Omega} \lambda(u,v) \, \mathrm{\Psi}\left( \left\| \nabla z(u,v) \right\|\right) \mathrm{d}u\,\mathrm{d}v
	\label{eq:4}
	\end{equation}	
\end{itemizepoint}
	
	In~\eqref{eq:4bis}, $\mathrm{\Phi}$ is chosen accordingly to the type of corruption the original image $z^0$ is affected by. For instance, $\mathrm{\Phi}_{L_2}(s) = s^2$ is the natural choice in the presence of additive, zero-mean, Gaussian noise, while $\mathrm{\Phi}_{L_1}(s) = |s|$ can be used in the presence of bi-exponential (Laplacian) noise, which is a rather good model when outliers come into play (e.g., ``salt \& pepper" noise). 
	
	In~\eqref{eq:4}, $\lambda \geq 0$ is a field of weights which control the respective influence of the fidelity and the regularization terms. It can be either manually tuned beforehand (if $\lambda(u,v) \equiv \lambda$, $\lambda$ can be seen as a ``hyper-parameter"), or defined as a function of $\|\nabla z(u,v)\|$. 
	
	The choice of $\mathrm{\Psi}$ must be made accordingly to a {desired smoothness of the restored image}. The \emph{quadratic penalty} $\mathrm{\Psi}_{L_2}(s) = s^2$ will produce ``smooth" images, while piecewise-constant images are obtained when choosing the \emph{sparsity penalty} $\mathrm{\Psi}_{L_0}(s) = 1-\delta(s)$, with $\delta(s) = 1$ if $s=0$ and $\delta(s) = 0$ otherwise. The latter approach preserves the edges, but the numerical solving is much more difficult, since the regularization term is non-smooth and non-convex. Hence, several choices of regularizers ``inbetween'' the quadratic and the sparsity ones have been suggested. 
	
	For instance, the total variation (TV) regularizer is obtained by setting $\mathrm{\Psi}(s) = |s|$. Efficient numerical methods exist for solving this non-smooth, yet convex, problem. Examples include primal-dual methods~\cite{Chambolle:2010a}, augmented Lagrangian approaches~\cite{Goldstein:2014a}, and forward-backward splittings~\cite{Ochs:2015a}. The latter can also be adapted to the case where the regularizer $\mathrm{\Psi}$ is non-convex, but smooth~\cite{Ochs:2014a}. Such non-convex regularization terms were shown to be particularly effective for edge-preserving image restoration~\cite{Geman_Reynolds,Lanza:2016,Nikolova}. 
	
	Another strategy is to {stick to quadratic regularization ($\Psi = \Psi_{L_2}$), but apply it} in a non-uniform manner {by tuning the field of weights $\lambda$ in~\eqref{eq:4}}. For instance, setting $\lambda(u,v)$ in~\eqref{eq:4} inversely proportional to $\|\nabla z(u,v)\|$ yields the ``anisotropic diffusion" model by Perona and Malik~\cite{Perona_Malik}. The discontinuity {set $K$} can also be automatically estimated and discarded {by setting $\lambda(u,v) \equiv \
0$ over $K$ and $\lambda(u,v) \equiv \lambda$ over $\Omega \backslash K$}, in the spirit of Mumford and Shah's segmentation method~\cite{Mumford_Shah}.

\subsection{Notations}

Although we chose for simplicity to write the variational problems in a continuous form, we are overall interested in solving discrete problems. Two different discretization strategies exist. The first one consists in using variational calculus to derive the (continuous) necessary optimality condition, then discretize it by finite differences, and eventually solve the discretized optimality condition. The alternative method
 is to discretize the functional itself by finite differences, before solving the optimality condition associated to the discrete problem. 
 
 As shown in~\cite{Durou:2007a}, the latter approach eases the handling of the boundary of $\Omega$, hence we use it as discretization strategy. The variational models hereafter will be presented using the continuous notations, because we find them more readable. The discrete notations will be used only when presenting the numerical solving. Yet, to avoid confusion, we will use caligraphic letters for the continuous energies (e.g., $\mathcal{E}$), and capital letters for their discrete counterparts (e.g., $E$). With these conventions, it should be clear whether an optimization problem is discrete or continuous. Hence, we will use the same notation $\nabla z = \left[ \partial_u z,\partial_v z \right]^\top$ both for the gradient of $z$ and its finite differences approximation.

\subsection{Proposed Variational Framework}

In this work, we show how to adapt the aforementioned variational models, originally designed for image restoration, to normal integration. Although both these inverse problems are formally very similar, they are somehow different, for the following reasons:
\begin{itemizepoint}
  \item The concept of edges in an image to restore is replaced by those of depth discontinuities and kinks. 
  \item Unlike image processing functionals, our data consist in an estimate $\mathbf{g}$ of the \emph{gradient} of the unknown $z$, in lieu of a corrupted version $z^0$ of $z$. Therefore, the fidelity term $\mathcal{F}(z)$ will apply to the difference between $\nabla z$ and $\mathbf{g}$, and it is the choice of this term which will or not allow depth discontinuities.
  \item Regularization terms are optional here: all the methods we discuss basically work even with $\mathcal{R}(z)\equiv 0$, but we may use this regularization term to allow introducing, if available, a prior on the surface (e.g., user-defined control points~\cite{Horovitz:2004a,Kimmel:2003} or a rough depth estimate obtained using a low-resolution depth sensor~\cite{Kadambi:2015a}). Such feature ``is appreciable, although not required"~\cite{Durou:2016a}.
\end{itemizepoint}   

We will discuss methods seeking the depth $z$ as the minimizer of an energy $\mathcal{E}(z)$ in the form~\eqref{eq:Ez}, but with different choices for $\mathcal{F}(z)$ and $\mathcal{R}(z)$:
\begin{itemizepoint}
	\item $\mathcal{F}(z)$ now represents a fidelity term penalizing the difference between the gradient of the recovered depth map $z$ and the datum $\mathbf{g}$:
	\begin{equation}
		\mathcal{F}(z) = \iint\displaylimits_{(u,v) \in \Omega} \mathrm{\Phi}\left( \left\| \nabla z(u,v) - \mathbf{g}(u,v) \right\|\right) \mathrm{d}u\,\mathrm{d}v
	\label{eq:3}
	\end{equation}
	
	\item The regularization term $\mathcal{R}(z)$ now represents prior knowledge of the depth\footnote{We consider only quadratic regularization terms: studying more robust ones (e.g., $L^1$ norm) is left as perspective.}:
	\begin{equation}
		\begin{array}{ll}
			& \mathcal{R}(z) = \displaystyle\iint\displaylimits_{(u,v) \in \Omega} \lambda(u,v) \left[z(u,v)-z^0(u,v)\right]^2
		\end{array}
	\label{eq:def_J}
	\end{equation}
	where $z^0$ is the prior, and $\lambda(u,v) \geq 0$ is a user-defined, spatially-varying, regularization weight. In this work, we consider for simplicity only the case where $\lambda$ does not depend on $z$. 
	\end{itemizepoint}

\subsection{Choosing $\lambda$ and $z^0$}
\label{sec:choiceLambda}

The {main} purpose of the regularization term $\mathcal{R}$ defined in~\eqref{eq:def_J} is to avoid numerical instabilities which may arise when considering solely the fidelity term~\eqref{eq:3}: this fidelity term depends only on $\nabla z$, and not on $z$ itself, hence the minimizer of~\eqref{eq:3} can be estimated only up to an additive ambiguity. 

Besides, one may also want to impose one or several control points on the surface~\cite{Horovitz:2004a,Kimmel:2003}. This can be achieved very simply within the proposed variational framework, by setting $\lambda(u,v) \equiv 0$ everywhere, except on the control points locations $(u,v)$ where a high value for $\lambda(u,v)$ must be set and the value $z^0(u,v)$ is fixed.

Another typical situation is when, given both a coarse depth estimate and an accurate normal estimate, one would like to ``merge" them in order to create a high-quality depth map. Such a problem arises, for instance, when refining the depth map of an RGB-D sensor (e.g.,  a Kinect) by means of shape-from-shading~\cite{Or-el:2015}, photometric stereo~\cite{Haque:2014a} or shape-from-polarization~\cite{Kadambi:2015a}. In such cases, we may set $z^0$ to the coarse depth map, and tune $\lambda$ so as to merge the coarse and fine estimates in the best way. Non-uniform weights may be used, in order to lower the influence of outliers in the coarse depth map~\cite{Haque:2014a}.

Eventually, in the absence of such priors, we will use the regularization term only to fix the integration constant: this is easily achieved by setting an arbitrary prior (e.g., $z^0(u,v) \equiv 0$), along with a small value for $\lambda$ (typically, $\lambda(u,v) \equiv \lambda = 10^{-6}$).

\section{Smooth Surfaces}
\label{sec:smooth}

We first tackle the problem of recovering a ``smooth" depth map $z$ from a noisy estimate $\mathbf{g}$ of $\nabla z$. To this end, we consider the quadratic variational problem:
\begin{align}
	& \underset{z}{\min}\iint\displaylimits_{(u,v) \in \Omega} \left\|\nabla z(u,v)-\mathbf{g}(u,v)\right\|^2 \nonumber \\[-1.5em]
	& \qquad\qquad\quad + \lambda(u,v) \left[z(u,v) - z^0(u,v)\right]^2 \mathrm{d}u\,\mathrm{d}v
\label{eq:continuous_LS}
\end{align}

{
When $\lambda \equiv 0$, Problem~\eqref{eq:continuous_LS} comes down to Horn and Brook's model~\cite{Horn:1986a}. In that particular case, an infinity of solutions $z \in W^{1,2}(\Omega)$ exist, and they differ by an additive constant\footnote{{Proof: by developing the terms inside the integral in~\eqref{eq:continuous_LS}, and integrating by parts, Theorem~6.2.5 in~\cite{Attouch:2014a} applies with $f:= -\nabla \cdot \mathbf{g}$ and $g:= \mathbf{g} \cdot \boldsymbol{\eta}$.}}. On the other hand, the regularization term allows us to guarantee uniqueness of the solution as soon as $\lambda$ is strictly positive almost everywhere\footnote{{Proof: by developing the terms inside the integral in~\eqref{eq:continuous_LS} and integrating by parts, Theorem~6.2.2-(ii) in~\cite{Attouch:2014a} applies with $f:= -\nabla \cdot \mathbf{g} + \lambda z^0$ and $g:= \mathbf{g} \cdot \boldsymbol{\eta}$.}}\footnote{{This condition makes the matrix of the associated discrete problem strictly diagonally dominant, see Section~\ref{sec:smooth_resolution}.}}.}

If the depth map $z$ is further assumed to be twice differentiable, the necessary optimality condition associated to the continuous optimization problem~\eqref{eq:continuous_LS} (Euler-Lagrange equation) {is written:}
\begin{align}
	& -\Delta z + \lambda z = -\nabla \cdot \mathbf{g} + \lambda z^0~~~\text{over}~\Omega \label{eq:ELCentre} \\
	& \left( \nabla z -\mathbf{g}\right)\cdot \boldsymbol{\eta} = 0 \qquad\qquad\quad~~ \text{over}~\partial \Omega \label{eq:ELBord}
\end{align}
with $\boldsymbol{\eta}$ a normal vector to the boundary $\partial \Omega$ of $\Omega$, $\Delta$ the Laplacian operator, and $\nabla \cdot$ the divergence operator. This condition is a linear PDE in $z$ {which can be discretized} using finite differences. 
Yet, providing a consistent discretization on the boundary of $\Omega$ is not straightforward~\cite{Harker:2015a}, {especially when dealing with non-rectangular domains $\Omega$ where many cases have to be considered~\cite{Breuss:2016a}}. Hence, we follow a different track, based on the discretization of the functional itself.


\subsection{Discretizing the Functional}
\label{sec:smooth_discretization}

Instead of a continuous gradient field $\mathbf{g}:\,\Omega \to \mathbb{R}^2$ over an open set $\Omega$, we are actually given a finite set of values $\{ \mathbf{g}_{u,v} = [p_{u,v},q_{u,v}]^\top,\,(u,v) \in \Omega \}$, where the $(u,v)$ represent the \emph{pixels} of a discrete subset $\Omega$ of a regular square 2D-grid\footnote{To ease the comparison between the variational and the discrete problems, we will use the same notation $\Omega$ for both the open set of $\mathbb{R}^2$ and the discrete subset of the grid.}. Solving the discrete integration problem requires estimating a finite set of values, i.e. the $|\Omega|$ unknown depth values $z_{u,v}$, $(u,v) \in \Omega$ ($|\cdot|$ denotes the cardinality), which are stacked columnwise in a vector $\mathbf{z} \in \mathbb{R}^{|\Omega|}$.

For now, let us use a Gaussian approximation for the noise contained in $\mathbf{g}$\footnote{{In 3D-recons\-truction applications such as photometric stereo~\cite{Woodham:1980a},  the assumption on the noise should rather be formulated on the images. This will be discussed in more details in Subsection~\ref{sec:PM}.}}, i.e., let us assume in the rest of this section that each datum $\mathbf{g}_{u,v},\, (u,v) \in \Omega$, is equal to the gradient $\nabla z(u,v)$ of the unknown depth map $z$, taken at point $(u,v)$, up to a zero-mean additive, homoskedastic (same variance at each location $(u,v)$), Gaussian noise:
\begin{equation}
	\mathbf{g}_{u,v} = \nabla z(u,v) + {\bm \epsilon}(u,v)
\label{eq:5}
\end{equation}
where ${\bm \epsilon}(u,v) \sim \mathcal{N}\left(\left[0,0\right]^\top,
	\begin{bmatrix}
		\sigma^2 & 0 \\ 0 & \sigma^2
	\end{bmatrix}\right)$ and $\sigma$ is unknown\footnote{The assumptions of equal variance $\sigma^2$ for both components and of a diagonal covariance matrix are introduced only for consistency with the least-squares problem~\eqref{eq:continuous_LS}. They are discussed with more care in Subsection~\ref{sec:PM}.}.
	Now, we need to give a discrete interpretation of the gradient operator in~\eqref{eq:5}, through finite differences.

In order to obtain a second-order accurate discretization, we combine forward and backward first-order finite differences, i.e. we consider that each measure of the gradient $\mathbf{g}_{u,v} = \left[p_{u,v},q_{u,v}\right]^\top$ provides us with up to four independent and identically distributed (i.i.d.) statistical observations, depending on the neighborhood of $(u,v)$. Indeed, its first component $p_{u,v}$ can be understood either in terms of both forward or backward finite differences (when both the bottom and the top\footnote{{The $u$-axis points ``downwards'', the $v$-axis points ``to the right'' and the $z$-axis points from the surface to the camera, see Figure~\ref{fig:troistrois}.}} neighbors are inside $\Omega$), by one of both these discretizations (only one neighbor inside $\Omega$), or by none of these finite differences (no neighbor inside $\Omega$). Formally, we model the $p$-observations in the following way:
\begin{align}
	& p_{u,v} = \overbrace{z_{u+1,v} - z_{u,v}}^{\partial_u^+ z_{u,v}} + \epsilon_u^+(u,v), \nonumber \\
	& \qquad\qquad\quad \forall (u,v) \in \underbrace{\left\{(u,v) \in \Omega \,|\, (u+1,v) \in \Omega \right\}}_{\Omega_u^+} \\[-0.5em]
	& p_{u,v} = \overbrace{z_{u,v} - z_{u-1,v}}^{\partial_u^- z_{u,v}} + \epsilon_u^-(u,v), \nonumber \\
	& \qquad\qquad\quad \forall (u,v) \in \underbrace{\left\{(u,v) \in \Omega \,|\, (u-1,v) \in \Omega \right\}}_{\Omega_u^-}
\end{align}
where $\epsilon^{+/-}_u \sim \mathcal{N}(0,\sigma^2)$. Hence, rather than considering that we are given $|\Omega|$ observations $p$, our discretization handles these data as $|\Omega_u^+|+|\Omega_u^-|$ observations, some of them being interpreted in terms of forward differences, some in terms of backward differences, some in terms of both forward and backward differences, the points without any neighbor in the $u$-direction being excluded. 


Symmetrically, the second component $q$ of $\mathbf{g}$ corresponds either to two, one or zero observations:
\begin{align}
	& q_{u,v} = \overbrace{z_{u,v+1} - z_{u,v}}^{\partial_v^+ z_{u,v}} + \epsilon_v^+(u,v), \nonumber \\
	& \qquad\qquad\quad \forall (u,v) \in \underbrace{\left\{(u,v) \in \Omega \,|\, (u,v+1) \in \Omega \right\}}_{\Omega_v^+} \\[-0.5em]
	& q_{u,v} = \overbrace{z_{u,v} - z_{u,v-1}}^{\partial_v^- z_{u,v}} + \epsilon_v^-(u,v),\nonumber \\
	& \qquad\qquad\quad \forall (u,v) \in \underbrace{\left\{(u,v) \in \Omega \,|\, (u,v-1) \in \Omega \right\}}_{\Omega_v^-} 
\end{align}
where $\epsilon^{+/-}_v \sim \mathcal{N}(0,\sigma^2)$. Given the Gaussianity of the noises $\epsilon^{+/-}_{u/v}$, their independence, and the fact that they all share the same standard deviation $\sigma$ and mean $0$, the joint likelihood of the observed gradients $\{\mathbf{g}_{u,v}\}_{(u,v)}$ is:
\begin{align}
	& L( \left\{\mathbf{g}_{u,v},\,(u,v) \in \Omega \right\}|\left\{z_{u,v},\,(u,v) \in \Omega \right\}) \nonumber \\
	& = \quad \prod_{(u,v) \in \Omega_u^+} \frac{1}{\sqrt{2\pi \sigma^2}} \exp\Bigg\{ - \frac{\left[ \partial_u^+z_{u,v} - p_{u,v} \right]^2}{2 \sigma^2} \Bigg\} \nonumber \\[-0.5em]
	& \quad \times \prod_{(u,v) \in \Omega_u^-} \frac{1}{\sqrt{2\pi \sigma^2}} \exp\Bigg\{ - \frac{\left[ \partial_u^-z_{u,v} - p_{u,v} \right]^2}{2 \sigma^2} \Bigg\} \nonumber \\[-0.5em]
	& \quad \times \prod_{(u,v) \in \Omega_v^+} \frac{1}{\sqrt{2\pi \sigma^2}} \exp\Bigg\{ - \frac{\left[ \partial_v^+z_{u,v} - q_{u,v} \right]^2}{2 \sigma^2} \Bigg\} \nonumber \\[-0.5em]
	& \quad \times \prod_{(u,v) \in \Omega_v^-} \frac{1}{\sqrt{2\pi \sigma^2}} \exp\Bigg\{ - \frac{\left[ \partial_v^-z_{u,v} - q_{u,v} \right]^2}{2 \sigma^2} \Bigg\}
\end{align}
and hence the maximum-likelihood estimate for the depth values is obtained by minimizing:
\begin{align}
	F_{L_2}(\mathbf{z}) = & \quad \frac12 \Bigg( \mathop{\sum\sum}_{(u,v) \in \Omega_u^+} \left[ \partial_u^+z_{u,v} - p_{u,v} \right]^2 \nonumber \\[-1em]
	& \quad~ + \mathop{\sum\sum}_{(u,v) \in \Omega_u^-} \left[ \partial_u^-z_{u,v} - p_{u,v} \right]^2 \Bigg) \nonumber \\[-0.25em]
	& + \frac12 \Bigg( \mathop{\sum\sum}_{(u,v) \in \Omega_v^+} \left[ \partial_v^+z_{u,v} - q_{u,v} \right]^2 \nonumber \\[-1em]
	& \quad~ + \mathop{\sum\sum}_{(u,v) \in \Omega_v^-} \left[ \partial_v^-z_{u,v} - q_{u,v} \right]^2 \Bigg)
\label{eq:11}
\end{align}
where the $\frac12$ coefficients are meant to ease the continuous interpretation: the integral of the fidelity term in~\eqref{eq:continuous_LS} is approximated by $F_{L_2}(\mathbf{z})$, expressed in~\eqref{eq:11} as the mean of the forward and the backward discretizations.

To obtain a more concise representation of this fidelity term, let us stack the data in two vectors $\mathbf{p} \in \mathbb{R}^{|\Omega|}$ and $\mathbf{q} \in \mathbb{R}^{|\Omega|}$. In addition, let us introduce four $|\Omega| \times |\Omega|$ differentiation matrices $\mathbf{D}^+_u$, $\mathbf{D}^-_u$, $\mathbf{D}^+_v$ and $\mathbf{D}^-_v$, associated with the finite differences operators $\partial_{u/v}^{+/-}$. For instance, the $i$-th line of $\mathbf{D}_u^+$ reads:
\begin{align}
	& \left(\mathbf{D}^+_u\right)_{i,\cdot} = \nonumber \\
	& ~~ \begin{cases}
		\Big[0,\dots,0,\underbrace{-1}_{\text{Position~}i},\underbrace{1}_{\text{Position~}i+1},0,\dots,0\Big]\quad\text{if~}m(i) \in \Omega^+_u \\
		\mathbf{0}^\top \quad \text{otherwise}
	\end{cases} 
\label{eq:DupDupDup}
\end{align}
where $m$ is the mapping associating linear indices $i$ with the pixel coordinates $(u,v)$:
\begin{equation}
	\begin{array}{lcl}
		m:\, & \{1,\dots,|\Omega|\} & \to \Omega \\
		~ & i & \mapsto m(i) = (u,v)
	\end{array}
\label{eq:defM}
\end{equation}

Once these matrices are defined,~\eqref{eq:11} is equal to:
\begin{align}
	F_{L_2}(\mathbf{z}) = & ~~~\, \frac12 \left( \left\|\mathbf{D}_u^+ \mathbf{z}-\mathbf{p}\right\|^2 + \left\| \mathbf{D}_u^- \mathbf{z}-\mathbf{p} \right\|^2 \right) \nonumber \\
	& \! + \frac12 \left( \left\| \mathbf{D}_v^+ \mathbf{z}-\mathbf{q} \right\|^2 + \left\| \mathbf{D}_v^- \mathbf{z}-\mathbf{q} \right\|^2 \right) \nonumber \\[-.5em]
	& \!- \frac12 \left( \mathop{\sum\sum}_{(u,v) \in \Omega \backslash \Omega_u^+} {p_{u,v}}^2 \!+ \!\!\!\!\!\! \mathop{\sum\sum}_{(u,v) \in \Omega \backslash \Omega_u^-} {p_{u,v}}^2 \right) \nonumber \\[-.25em]
	& \!- \frac12 \left( \mathop{\sum\sum}_{(u,v) \in \Omega \backslash \Omega_v^+} {q_{u,v}}^2 + \!\!\!\!\!\! \mathop{\sum\sum}_{(u,v) \in \Omega \backslash \Omega_v^-} {q_{u,v}}^2 \right)
\label{eq:13}
\end{align}
The terms in both the last rows of~\eqref{eq:13} being independent from the $z$-values, they do not influence the actual minimization and will thus be omitted from now on.

The regularization term~\eqref{eq:def_J} is discretized as:
\begin{align}
	& R(\mathbf{z}) \!=\! \mathop{\sum\sum}_{(u,v) \in \Omega} \lambda_{u,v} \left[z_{u,v}-z^0_{u,v} \right]^2 \! = \! \ \left\|{\bm \Lambda}\left(\mathbf{z}-\mathbf{z}^0\right)\right\|^2
\label{eq:14}
\end{align}
with ${\bm \Lambda}$ a $|\Omega|\times|\Omega|$ diagonal matrix containing the values $\sqrt{\lambda_{u,v}},\,(u,v) \in \Omega$.

Putting it altogether, our quadratic integration method 
reads as the minimization of the discrete functional:
\begin{align}
	& E_{L_2}(\mathbf{z}) = \frac12 \left( \left\|\mathbf{D}_u^+ \mathbf{z}-\mathbf{p}\right\|^2 + \left\| \mathbf{D}_u^- \mathbf{z}-\mathbf{p} \right\|^2 \right) \nonumber \\
	& + \frac12 \!\left( \left\| \mathbf{D}_v^+ \mathbf{z}-\mathbf{q} \right\|^2 \!\!\! +\!  \left\| \mathbf{D}_v^- \mathbf{z}-\mathbf{q} \right\|^2 \right) \!+\! \left\|{\bm \Lambda}\!\left(\mathbf{z}-\mathbf{z}^0\right)\right\|^2
\label{eq:discrete_L_2}
\end{align}

\subsection{Numerical Solution}
\label{sec:smooth_resolution}

The optimality condition associated with the discrete functional~\eqref{eq:discrete_L_2} is a linear equation in $\mathbf{z}$:
\begin{equation}
	\mathbf{A}\mathbf{z} = \mathbf{b}
\label{eq:15}
\end{equation}
where $\mathbf{A}$ is a $|\Omega|\times|\Omega|$ symmetric matrix\footnote{$\mathbf{A}$ and $\mathbf{b}$ are purposely divided by two in order to ease the continuous interpretation of {Subsection}~\ref{sec:smooth_continuous}.}:
\begin{align}
	& \mathbf{A} = \overbrace{ \frac12 \Big[ {\mathbf{D}^+_u}^\top \mathbf{D}^+_u + {\mathbf{D}^-_u}^\top \mathbf{D}^-_u + { \mathbf{D}^+_v }^\top \mathbf{D}^+_v + {\mathbf{D}^-_v }^\top \mathbf{D}^-_v \Big]}^{\mathbf{L}} \nonumber \\
	& \qquad + {\bm \Lambda}^2
\label{eq:def_A}
\end{align}
and $\mathbf{b}$ is a $|\Omega| \times 1$ vector:
\begin{align}
	& \mathbf{b} = \overbrace{\frac12\Big[ {\mathbf{D}^+_u}^\top + {\mathbf{D}^-_u}^\top \Big]}^{ \mathbf{D}_u} \mathbf{p} + \overbrace{ \frac12\Big[ {\mathbf{D}^+_v}^\top + {\mathbf{D}^-_v}^\top \Big]}^{ \mathbf{D}_v} \mathbf{q}\nonumber \\
	& \quad~~ + {\bm \Lambda}^2 \mathbf{z}^0
\label{eq:def_b}
\end{align}

The matrix $\mathbf{A}$ is {sparse}: it contains at most five non-zero entries per row. In addition, it is diagonal dominant: if $\left({\bm \Lambda}\right)_{i,i} = \mathbf{0}$, the value $\left({\bm A}\right)_{i,i}$ of a diagonal entry is equal to the opposite of the sum of the other entries $\left({\bm A}\right)_{i,j},\, i\neq j$, from the same row $i$. It becomes strictly superior as soon as $\left({\bm \Lambda}\right)_{i,i}$ is strictly positive. Let us also remark that, when $\Omega$ describes a rectangular domain and the regularization weights are uniform ($\lambda(u,v) \equiv \lambda$), $\mathbf{A}$ is a Toeplitz matrix. Yet, this structure is lost in the general case where it can only be said that $\mathbf{A}$ is a sparse, symmetric, diagonal dominant (SDD) matrix with at most $5 |\Omega|$ non-zero elements. It is positive semi-definite when ${\bm \Lambda} = \mathbf{0}$, and positive definite as soon as one of the $\lambda_{u,v}$ is non-zero.

System~\eqref{eq:15} can be solved by means of the conjugate gradient algorithm. Initialization will not influence the actual solution, but it may influence the number of iterations required to reach convergence. In our experiments, we used $z^0$ as initial guess, yet more elaborate initialization strategies may yield faster convergence~\cite{Breuss:2016a}. To ensure $\mathcal{P}_\text{Fast}$, we used the multigrid preconditioning technique~\cite{Koutis:2011}, {which has a negligible cost of computation and still} bounds the computational complexity required to reach a $\epsilon$ relative accuracy\footnote{In our experiments, the threshold of the stopping criterion is set to $\epsilon = 10^{-4}$.} by:
\begin{equation}
	O\left(5n \, \log(n) \, \log(1/\epsilon) \right)
	\label{eq:complexity}
\end{equation}
where $n = |\Omega|$\footnote{{In~\eqref{eq:complexity}, the factor $5n$ is nothing else than the number of non-zero elements in $\mathbf{A}$. Therefore, exploiting sparsity is not as ``fruitless'' as argued in~\cite{Harker:2015a} when it comes to solving large linear systems faster than using Gaussian elimination (complexity $O(n^3)$).}}. This complexity is inbetween the complexities of the approaches based on Sylvester equations~\cite{Harker:2015a} ($O(n^{1.5})$) and on DCT~\cite{Simchony:1990a} ($O(n \log(n))$). {Besides}, these competing methods explicitly require that $\Omega$ is rectangular, while ours does not.

By construction, {the} integration method consisting in minimizing~\eqref{eq:discrete_L_2} satisfies the $\mathcal{P}_\text{Robust}$ property (it is the maximum-likelihood estimate in the presence of zero-mean Gaussian noise). The discretization we introduced does not assume any particular shape for $\Omega$, neither treats the boundary in a specific manner, hence $\mathcal{P}_{\text{FreeB}}$ and $\mathcal{P}_{\text{NoRect}}$ are also satisfied. We also showed that $\mathcal{P}_\text{Fast}$ could be satisfied, using a solving method based on the preconditioned conjugate gradient algorithm. Eventually, let us recall that tuning $\lambda$ and$/$or manually fixing the values of the prior $z_0$ is necessary only to introduce a prior, but not in general. Hence, $\mathcal{P}_\text{NoPar}$ is also enforced. In conclusion, all the desired properties are satisfied, except $\mathcal{P}_\text{Disc}$. Let us now provide additional remarks on the connections between the proposed discrete approach and a fully variational one.

\subsection{Continuous Interpretation}
\label{sec:smooth_continuous}

System~\eqref{eq:15} is nothing else than a discrete analogue of the {continuous optimality conditions~\eqref{eq:ELCentre} and~\eqref{eq:ELBord}:}
\begin{equation}
	\underbrace{ \mathbf{L} \mathbf{z}}_{\approx - \Delta z} + \underbrace{{\bm \Lambda}^2 \mathbf{z}}_{\approx \lambda z} = \underbrace{\mathbf{D}_u \mathbf{p} + \mathbf{D}_v \mathbf{q}}_{\approx - \nabla \cdot \mathbf{g}} + \underbrace{{\bm \Lambda}^2 \mathbf{z}^0}_{\approx \lambda z^0}
\label{eq:28}
\end{equation}
where the matrix-vector products are easily interpreted in terms of the differential operators in the continuous formula~\eqref{eq:ELCentre}. One major advantage when reasoning from the beginning in the discrete setting is that one does not need to find out how to discretize the \emph{natural}\footnote{As stated in~\cite{Harker:2015a}, homogeneous Neumann {boundary} conditions of the type $\nabla z \cdot {\bm \eta} = 0$, used e.g. in~\cite{Agrawal:2006a}, should be avoided. } boundary condition~\eqref{eq:ELBord}, which was already emphasized in~\cite{Durou:2007a,Harker:2015a}. Yet, the identifications in~\eqref{eq:28} show that both the discrete and continuous approaches are equivalent, provided that an appropriate discretization of the continuous optimality condition is used. It is thus possible to derive $O(5n \, \log(n)\, \log(1/\epsilon))$ algorithms based on the discretization of the Euler-Lagrange equation, contrarily to what is stated in~\cite{Harker:2015a}. The real drawback of such approaches does not lie in complexity, but in the difficult discretization of the boundary condition. This is further explored in the next subsection.

\subsection{Example}
\label{sec:smooth_example}

\begin{figure}[!ht]
\begin{center}
	\def\svgwidth{0.75\linewidth}
		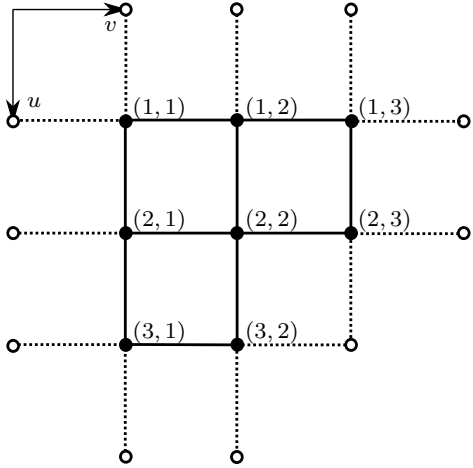
\end{center}
\caption{Example of non-rectangular domain $\Omega$ (solid dots) inside a $3 \times 3$ grid. When invoking the continuous optimality condition, the discrete approximations of the Laplacian and {of} the divergence near the boundary involve several points inside $\partial \Omega$ (circles) for which no data is available. First-order approximation of the \emph{natural} boundary condition~\eqref{eq:ELBord} is thus required. Relying only on discrete optimization simplifies a lot the boundary handling.}
\label{fig:troistrois}
\end{figure}

To clarify the proposed discretization of the integration problem, let us consider a non-rectangular domain $\Omega$ inside a $3 \times 3$ grid, like the one depicted in Figure~\ref{fig:troistrois}.

The vectorized unknown depth $\mathbf{z}$ and the vectorized components $\mathbf{p}$ and $\mathbf{q}$ of the gradient write in this case:
\begin{equation}
	\mathbf{z} = \begin{bmatrix}
		z_{1,1} \\
		z_{2,1} \\
		z_{3,1} \\
		z_{1,2} \\
		z_{2,2} \\
		z_{3,2} \\
		z_{1,3} \\
		z_{2,3}
	\end{bmatrix}
	\quad
	\mathbf{p} = \begin{bmatrix}
		p_{1,1} \\
		p_{2,1} \\
		p_{3,1} \\
		p_{1,2} \\
		p_{2,2} \\
		p_{3,2} \\
		p_{1,3} \\
		p_{2,3}
	\end{bmatrix}
	\quad
	\mathbf{q} = \begin{bmatrix}
		q_{1,1} \\
		q_{2,1} \\
		q_{3,1} \\
		q_{1,2} \\
		q_{2,2} \\
		q_{3,2} \\
		q_{1,3} \\
		q_{2,3}
	\end{bmatrix}
\end{equation}

The sets $\Omega_{u/v}^{+/-}$ all contain five pixels:
\begin{align}
	& \Omega_u^+ = \left\{
		\left(1,1\right),\left(2,1\right),\left(1,2\right),\left(2,2\right),\left(1,3\right)
	\right\} \\
	& \Omega_u^- = \left\{
		\left(2,1\right),\left(3,1\right),\left(2,2\right),\left(3,2\right),\left(2,3\right)
	\right\} \\
	& \Omega_v^+ = \left\{
		\left(1,1\right),\left(2,1\right),\left(3,1\right),\left(1,2\right),\left(2,2\right)
	\right\} \\
	& \Omega_v^- = \left\{
		\left(1,2\right),\left(2,2\right),\left(3,2\right),\left(1,3\right),\left(2,3\right)
	\right\}
\end{align}
so that the differentiation matrices $\mathbf{D}_{u/v}^{+/-}$ have five non-zero rows {according to their definition~\eqref{eq:DupDupDup}}. For instance, the matrix associated with the forward finite differences operator $\partial_u^+$ reads:
\begin{align}
	& \mathbf{D}_u^+ =
	\begin{bmatrix}
		-1 & 1 & 0 & 0 & 0 & 0 & 0 & 0 \\
		0 & -1 & 1 & 0 & 0 & 0 & 0 & 0 \\
		0 & 0 & 0 & 0 & 0 & 0 & 0 & 0 \\
		0 & 0 & 0 & -1 & 1 & 0 & 0 & 0 \\
		0 & 0 & 0 & 0 & -1 & 1 & 0 & 0 \\
		0 & 0 & 0 & 0 & 0 & 0 & 0 & 0 \\
		0 & 0 & 0 & 0 & 0 & 0 & -1 & 1 \\
		0 & 0 & 0 & 0 & 0 & 0 & 0 & 0 \\
	\end{bmatrix}
\end{align}

The {negative} Laplacian matrix~$\mathbf{L}$ defined in~\eqref{eq:def_A} is worth:
\begin{equation}
	\mathbf{L} =
	\begin{bmatrix}
		2 & -1 & 0 & -1 & 0 & 0 & 0 & 0 \\
		-1 & 3 & -1 & 0 & -1 & 0 & 0 & 0 \\
		0 & -1 & 2 & 0 & 0 & -1 & 0 & 0 \\
		-1 & 0 & 0 & 3 & -1 & 0 & -1 & 0 \\
		0 & -1 & 0 & -1 & 4 & -1 & 0 & -1 \\
		0 & 0 & -1 & 0 & -1 & 2 & 0 & 0 \\
		0 & 0 & 0 & -1 & 0 & 0 & 2 & -1 \\
		0 & 0 & 0 & 0 & -1 & 0 & -1 & 2 \\
	\end{bmatrix}
\end{equation}
One can observe that this matrix describes the connectivity of the graph representing the discrete domain $\Omega$: the diagonal elements $\left(\mathbf{L}\right)_{i,i}$ are the numbers of neighbors connected to the $i$-th point, and the off-diagonals elements $\left(\mathbf{L}\right)_{i,j}$ are worth $-1$ if the $i$-th and $j$-th points are connected, $0$ otherwise.

Eventually, the matrices $\mathbf{D}_u$ and $\mathbf{D}_v$ defined in~\eqref{eq:def_b} are equal to:
\begin{align}
	& \mathbf{D}_u = \frac12
	\begin{bmatrix}
		-1 & -1 & 0 & 0 & 0 & 0 & 0 & 0 \\
		1 & 0 & -1 & 0 & 0 & 0 & 0 & 0 \\
		0 & 1 & 1 & 0 & 0 & 0 & 0 & 0 \\
		0 & 0 & 0 & -1 & -1 & 0 & 0 & 0 \\
		0 & 0 & 0 & 1 & 0 & -1 & 0 & 0 \\
		0 & 0 & 0 & 0 & 1 & 1 & 0 & 0 \\
		0 & 0 & 0 & 0 & 0 & 0 & -1 & -1 \\
		0 & 0 & 0 & 0 & 0 & 0 & 1 & 1
	\end{bmatrix} \\
	& \mathbf{D}_v = \frac12
	\begin{bmatrix}
		-1 & 0 & 0 & -1 & 0 & 0 & 0 & 0 \\
		0 & -1 & 0 & 0 & -1 & 0 & 0 & 0 \\
		0 & 0 & -1 & 0 & 0 & -1 & 0 & 0 \\
		1 & 0 & 0 & 0 & 0 & 0 & -1 & 0 \\
		0 & 1 & 0 & 0 & 0 & 0 & 0 & -1 \\
		0 & 0 & 1 & 0 & 0 & 1 & 0 & 0 \\
		0 & 0 & 0 & 1 & 0 & 0 & 1 & 0 \\
		0 & 0 & 0 & 0 & 1 & 0 & 1 & 1
	\end{bmatrix}
\end{align}

Let us now show how these matrices relate to the discretization of the continuous optimality condition~\eqref{eq:ELCentre}. Using second-order central finite differences approximations of the Laplacian ($\Delta z_{u,v} \approx z_{u,v-1}+z_{u-1,v}+z_{u+1,v}+z_{u,v+1}-4z_{u,v}$) and of the divergence operator ($\nabla \cdot \mathbf{g}_{u,v} \approx \frac12\left(p_{u+1,v}-p_{u-1,v}\right)+\frac12 \left(q_{u,v+1}-q_{u,v-1}\right)$), we obtain:
\begin{align}
	& \left[4z_{u,v} \!-\! z_{u,v-1} \!-\! z_{u-1,v} \!-\! z_{u+1,v} \!-\! z_{u,v+1} \right] \!+\! \lambda_{u,v} z_{u,v} = \nonumber \\
	& \frac12\left[p_{u-1,v}-p_{u+1,v}\right]\!+\frac12\!\left[q_{u,v-1}-q_{u,v+1}\right] \!+\! \lambda_{u,v} z^0_{u,v}
\label{eq:Da26}
\end{align}
The pixel $(u,v) = (2,2)$ is the only one whose four neighbors are inside $\Omega$. In that case,~\eqref{eq:Da26} becomes:
\begin{align}
	& \underbrace{\left[4z_{2,2} - z_{2,1}-z_{1,2}-z_{3,2}-z_{2,3} \right]}_{= \left(\mathbf{L}\right)_{5,\cdot}\mathbf{z}} + \underbrace{\lambda_{2,2} z_{2,2}}_{= \left({\bm \Lambda}^2\right)_{5,\cdot}\mathbf{z}} \nonumber \\
	& \qquad = \underbrace{\frac12\left[p_{1,2}-p_{3,2}\right]}_{= \left(\mathbf{D}_u\right)_{5,\cdot}\mathbf{p}}+\underbrace{\frac12\left[q_{2,1}-q_{2,3}\right]}_{= \left(\mathbf{D}_v\right)_{5,\cdot}\mathbf{q}} + \underbrace{\lambda_{2,2} z^0_{2,2}}_{= \left({\bm \Lambda}^2\right)_{5,\cdot}\mathbf{z}^0}
\end{align}
where we recognize the fifth equation of the discrete optimality condition~\eqref{eq:28}. This shows that, for pixels having all four neighbors inside $\Omega$, both the continuous and the discrete variational formulations yield the same discretizations.

Now, let us consider a pixel near the boundary, for instance pixel $(1,1)$. Using the same second-order differences, \eqref{eq:Da26} reads:
\begin{align}
	& \left[4z_{1,1} - z_{1,0}-z_{0,1}-z_{2,1}-z_{1,2} \right] + \lambda_{1,1} z_{1,1} \nonumber \\
	& \qquad = \frac12\left[p_{0,1}-p_{2,1}\right]+\frac12\left[q_{1,0}-q_{1,2}\right] + \lambda_{1,1} z^0_{1,1}
\label{eq:discrete_BC}
\end{align}
which involves the values $z_{1,0}$ and $z_{0,1}$ of the depth map, which we are not willing to estimate, and the values $p_{0,1}$ and $q_{1,0}$ of the gradient field, which are not provided as data. To eliminate these four values, we need to resort to boundary conditions on $z$, $p$ and $q$. The discretizations, using first order forward finite differences, of the natural boundary condition~\eqref{eq:ELBord}, at locations $(1,0)$ and $(0,1)$, read:
\begin{align}
	& z_{1,1} - z_{1,0} = q_{1,0} \\
	& z_{1,1} - z_{0,1} = p_{0,1}
\end{align}
hence the unknown depth values $z_{1,0}$ and $z_{0,1}$ can be eliminated from {Equation}~\eqref{eq:discrete_BC}:
\begin{align}
	& \left[2z_{1,1} - z_{2,1}-z_{1,2} \right] + \lambda_{1,1} z_{1,1} \nonumber \\
	& \quad~ = \frac12\left[-p_{0,1}-p_{2,1}\right]+\frac12\left[-q_{1,0}-q_{1,2}\right] + \lambda_{1,1} z^0_{1,1}
\label{eq:discrete_BC2}
\end{align}
Eventually, the unknown values $p_{0,1}$ and $q_{1,0}$ need to be approximated. Since we have no information at all about the values of $\mathbf{g}$ outside $\Omega$, we use homogeneous Neumann boundary conditions\footnote{This assumption is weaker than the homogeneous Neumann boundary condition $\nabla z \cdot \boldsymbol{\eta} = 0$ used {by Agrawal et al.} in~\cite{Agrawal:2006a}.}:
\begin{align}
	\nabla p \cdot \boldsymbol{\eta} = 0 \quad\text{~over~} \partial\Omega \\
	\nabla q \cdot \boldsymbol{\eta} = 0 \quad\text{~over~} \partial\Omega
\end{align}
Discretizing these boundary conditions using first order forward finite differences, we obtain:
\begin{align}
	& p_{0,1} = p_{1,1} \\
	& q_{1,0} = q_{1,1}
\end{align}
{Using these identifications}, the discretized optimality condition~\eqref{eq:discrete_BC2} is given by:
\begin{align}
	& \underbrace{\left[2z_{1,1} - z_{2,1}-z_{1,2} \right]}_{= \left(\mathbf{L}\right)_{1,\cdot}\mathbf{z}} + \underbrace{\lambda_{1,1} z_{1,1}}_{= \left({\bm \Lambda}^2\right)_{1,\cdot}\mathbf{z}} \nonumber \\
	& ~ = \underbrace{\frac12\left[-p_{1,1}-p_{2,1}\right]}_{= \left(\mathbf{D}_u\right)_{1,\cdot}\mathbf{p}}+\underbrace{\frac12\left[-q_{1,1}-q_{1,2}\right]}_{= \left(\mathbf{D}_v\right)_{1,\cdot}\mathbf{q}} + \underbrace{\lambda_{1,1} z^0_{1,1}}_{= \left({\bm \Lambda}^2\right)_{1,\cdot}\mathbf{z}^0}
\label{eq:discrete_BC3}
\end{align}
which is exactly the first equation of the discrete optimality condition~\eqref{eq:28}.

\begin{figure*}[!ht]
\begin{center}
	\begin{tabular}{cc}
		\includegraphics[width = 0.45\linewidth]{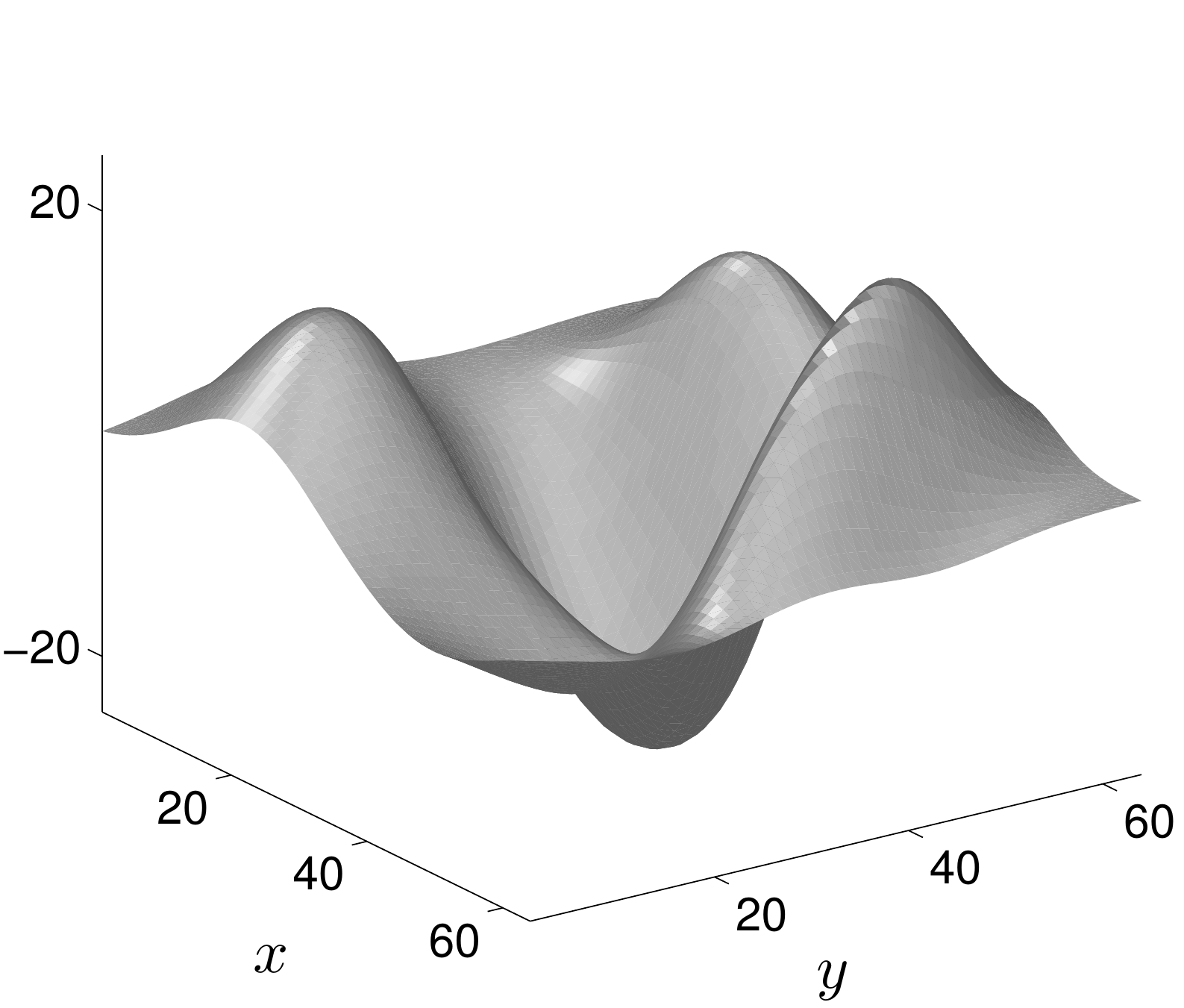} &
		\includegraphics[width = 0.45\linewidth]{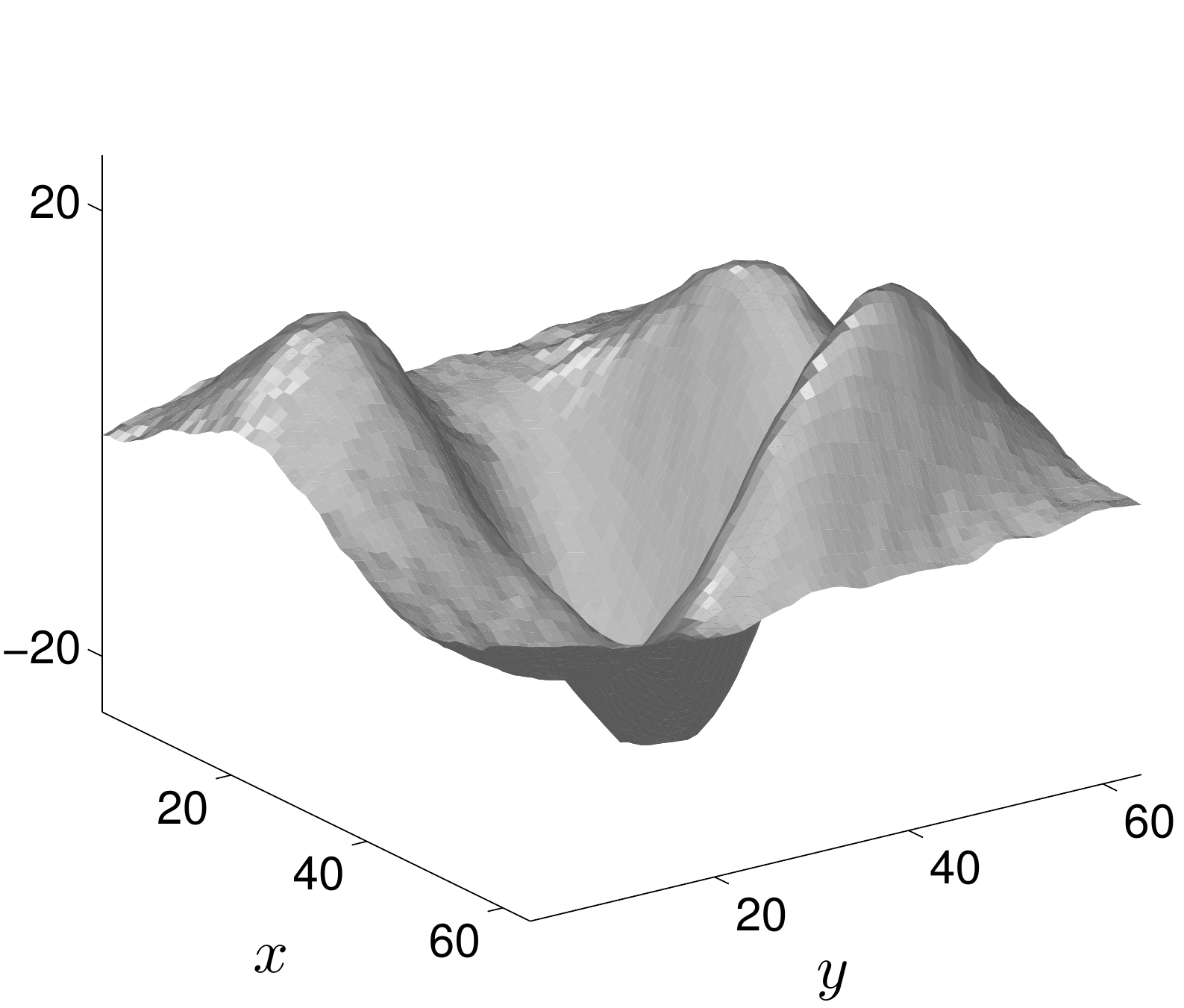} \\
		Ground-truth & Simchony et al.~\cite{Simchony:1990a} \\
		\includegraphics[width = 0.45\linewidth]{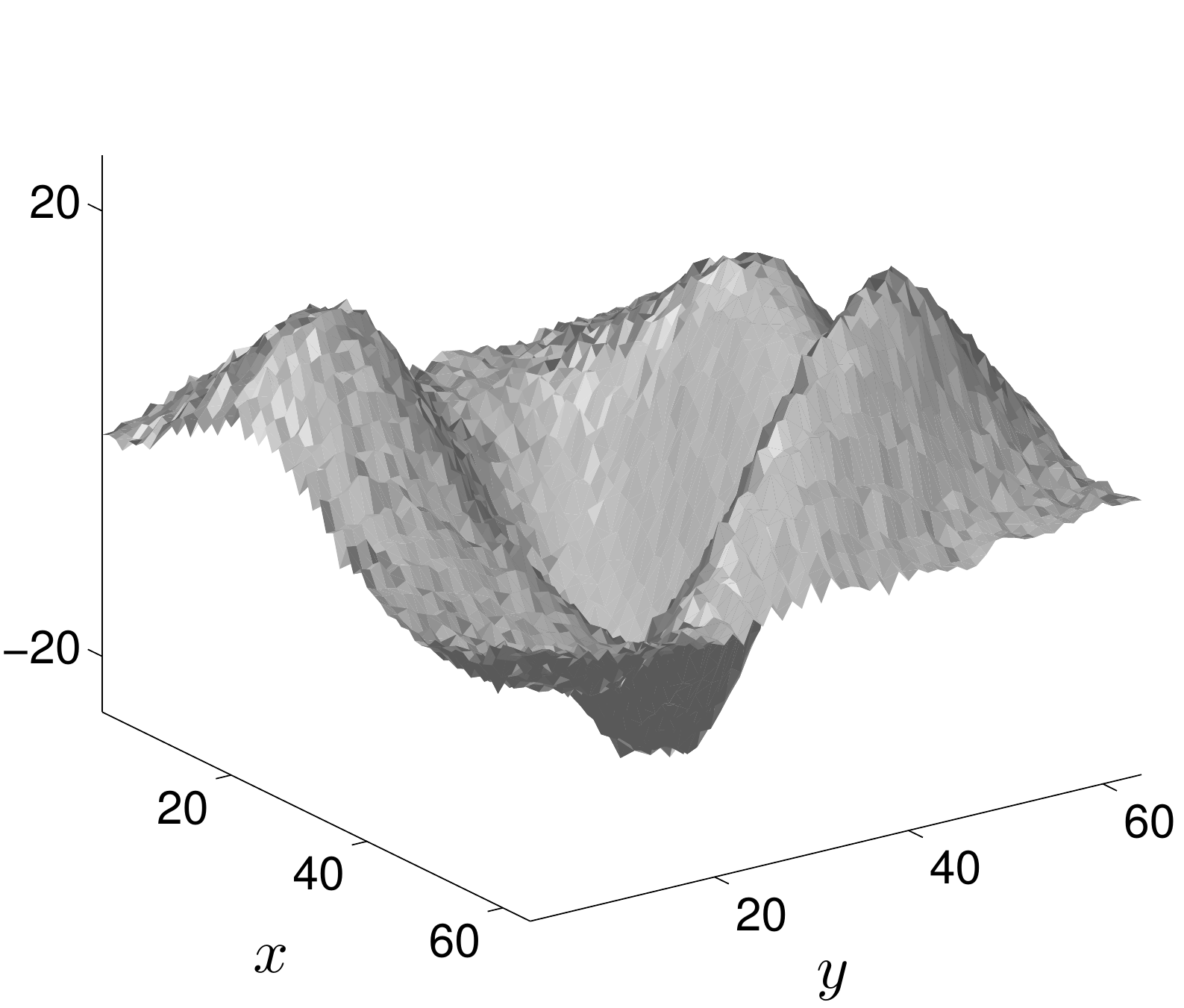} &
		\includegraphics[width = 0.45\linewidth]{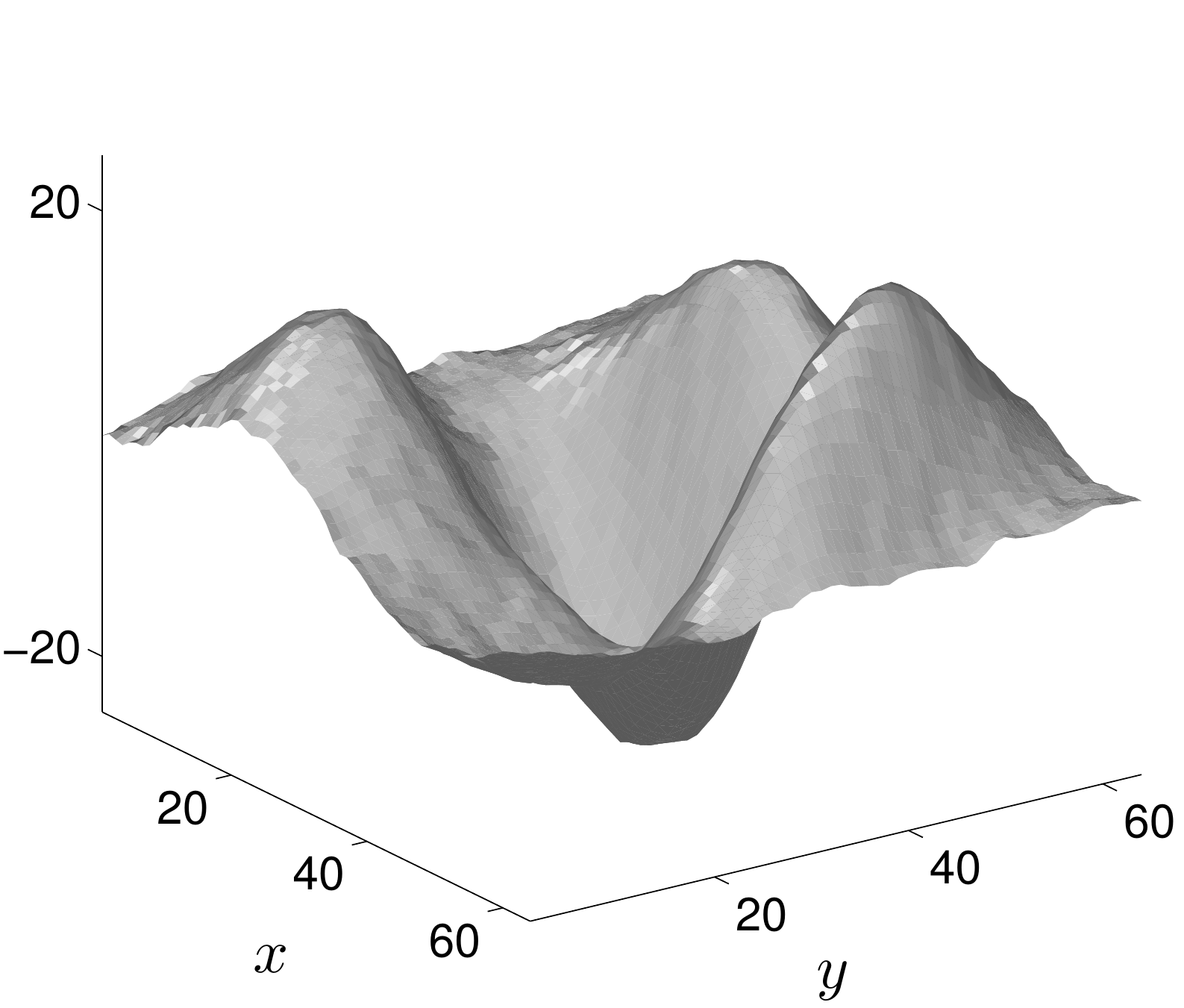} \\
		Harker and O'Leary~\cite{Harker:2015a} & Proposed
	\end{tabular}
\end{center}
\caption{Qualitative evaluation of the $\mathcal{P}_\text{Robust}$ property. An additive, zero-mean, Gaussian noise with standard deviation $0.1 \|\mathbf{g}\|_\infty$ was added to the (analytically known) gradient of the ground-truth surface, before integrating this gradient by three least-squares methods. Ours qualitatively provides better results than the Sylvester equations method from Harker and O'Leary~\cite{Harker:2015a}. It seems to provide similar robustness as the DCT solution from Simchony et al.~\cite{Simchony:1990a}, but the quantitative evaluation from {Figure}~\ref{fig:courbes_smooth} shows that our method is actually more accurate.}
\label{fig:peaks_smooth}
\end{figure*}

Using a similar rationale, we obtain equivalence of both formulations for the eight points inside $\Omega$. Yet, let us emphasize that discretizing the continuous optimality condition requires treating, on this example with a rather ``simple" shape for $\Omega$, not less than seven different cases (only pixels $(3,2)$ and $(2,3)$ are similar). More general shapes bring out to play even more particular cases (points having only one neighbor inside $\Omega$). Furthermore, boundary conditions must be invoked in order to approximate the depth values and the data outside $\Omega$. On the other hand, the discrete functional provides exactly the same optimality condition, but without these drawbacks. The boundary conditions can be viewed as \emph{implicitly} enforced, hence $\mathcal{P}_{\text{FreeB}}$ is satisfied.

\subsection{Empirical Evaluation}

\begin{figure}[!ht]
\begin{center}
	\begin{tabular}{c}
		~~~\includegraphics[width = 0.95\linewidth]{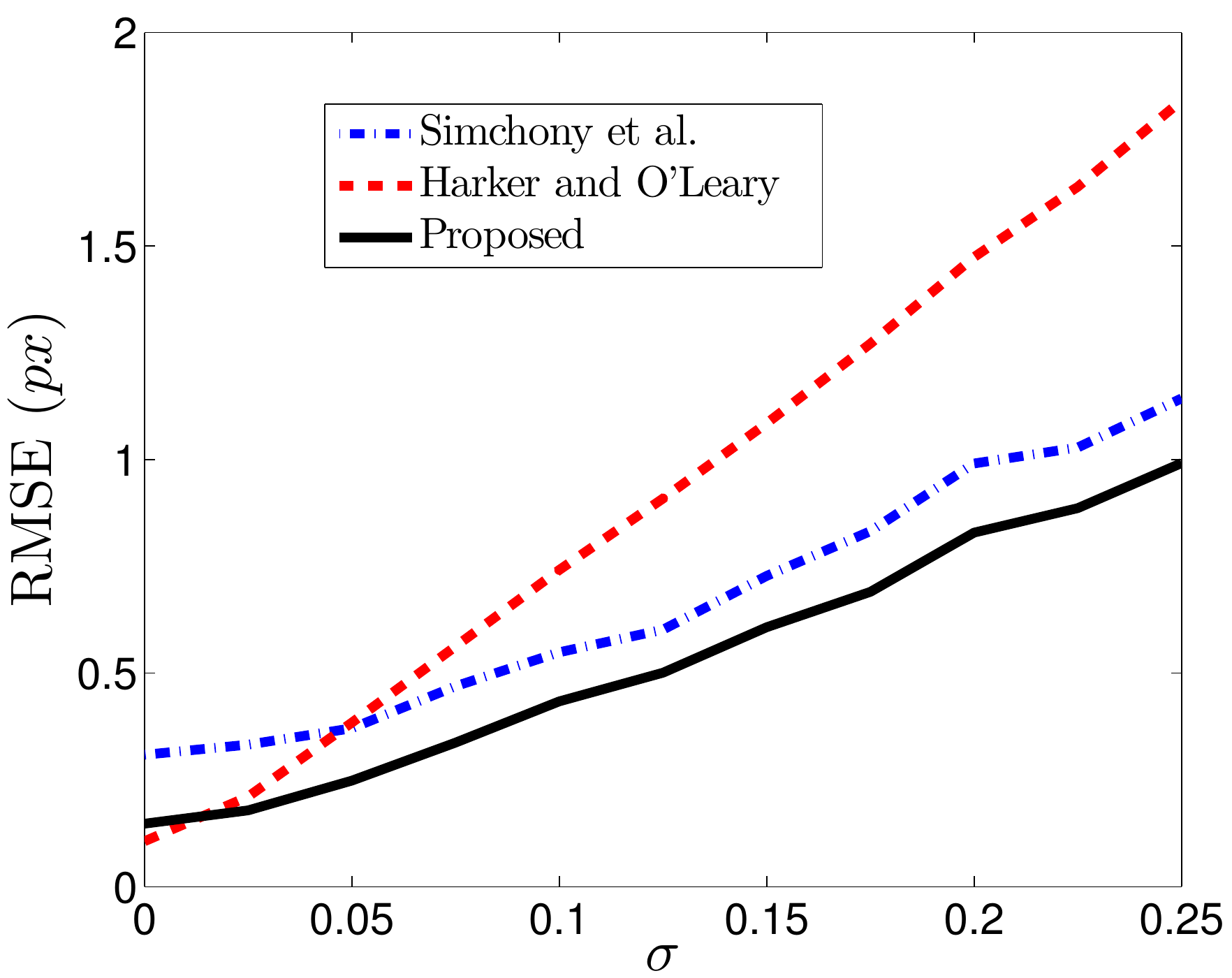} \\
		\includegraphics[width = 0.99\linewidth]{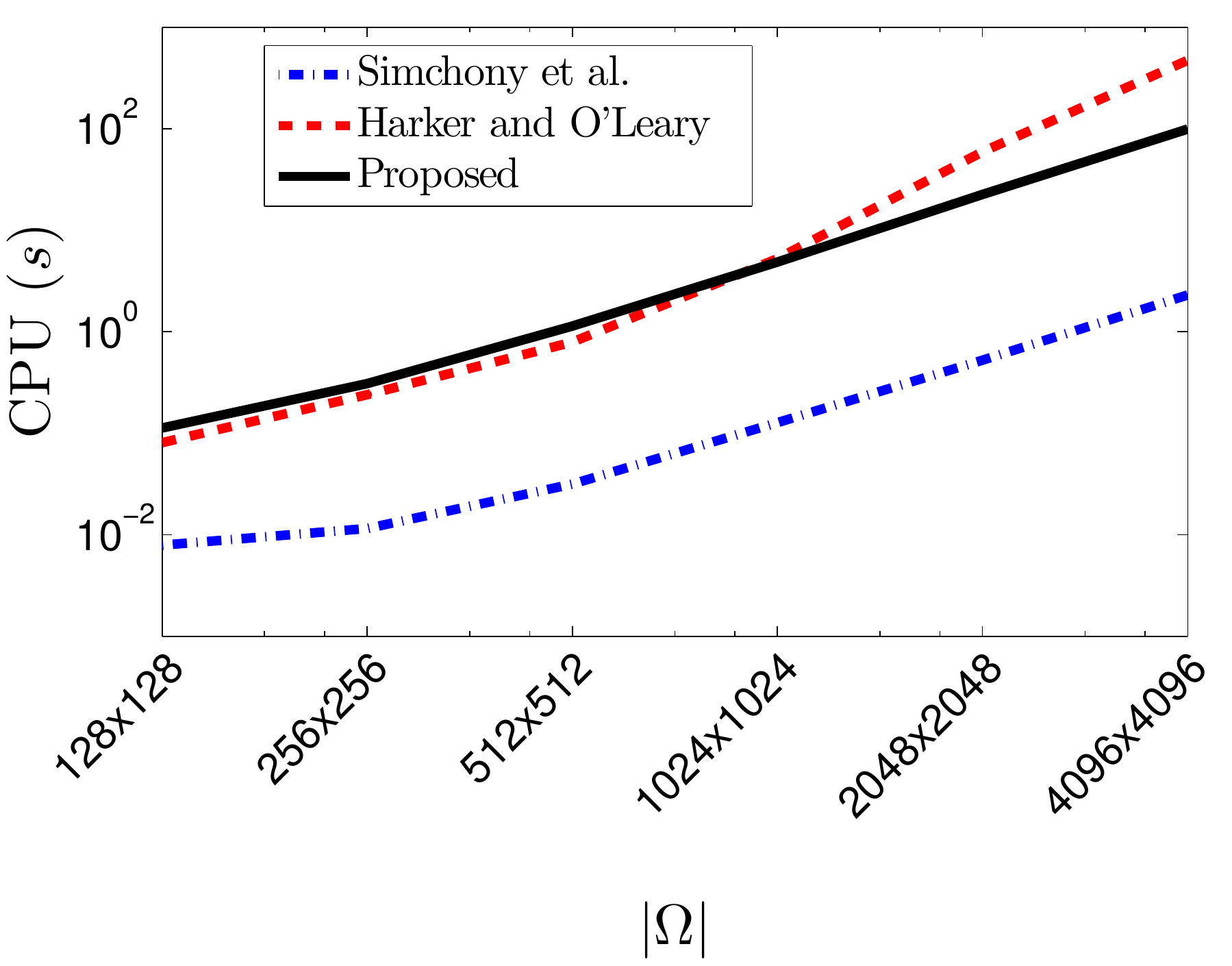}
	\end{tabular}
\end{center}
\caption{Quantitative evaluation of the $\mathcal{P}_\text{Robust}$ (top) and $\mathcal{P}_\text{Fast}$ (bottom) properties. Top: RMSE between the depth ground-truth and the ones reconstructed from noisy gradients (adding a zero-mean Gaussian noise with standard deviation $\sigma \|\mathbf{g}\|_\infty$, {for several values of $\sigma$}). Bottom: {Computation time as a function of the size $|\Omega|$ of the reconstruction domain $\Omega$. The method we put forward has a complexity which is inbetween those of the methods of Simchony et al.~\cite{Simchony:1990a} (based on DCT) and of Harker and O'Leary~\cite{Harker:2015a} (based on Sylvester equations), while being slightly more accurate than both of them.}}
\label{fig:courbes_smooth}
\end{figure}

We first consider the smooth surface from {Figure}~\ref{fig:peaks_smooth}, whose normals are analytically known~\cite{Harker:2015a}, and compare three discrete least-squares methods which all satisfy $\mathcal{P}_{\text{Fast}}$, $\mathcal{P}_{\text{Robust}}$ and $\mathcal{P}_{\text{FreeB}}$: the DCT solution~\cite{Simchony:1990a}, the Sylvester equations method~\cite{Harker:2015a}, and the proposed one. 
As shown in {Figures}~\ref{fig:peaks_smooth} and~\ref{fig:courbes_smooth}, our solution is slightly more accurate. 
Indeed, the bias near the boundary induced by the DCT method is corrected. On the other hand, we believe the reason why our method is more accurate than that from~\cite{Harker:2015a} is because we use {a combination of forward and backward} finite differences, while \cite{Harker:2015a} relies on central differences. {Indeed, when using central differences to discretize the gradient, the second-order operator (Laplacian) {appearing} in the Sylvester equations from~\cite{Harker:2015a} involves none of the direct neighbors, which may be non-robust for noisy data (see, for instance, Appendix 3 in~\cite{AubertKornprobst}). For instance, let us consider a 1D domain $\Omega$ with 7 pixels. Then, the following differentiation matrix is advocated in~\cite{Harker:2015a}:
\begin{equation}
\mathbf{D}_u = \frac12 \begin{bmatrix}
-3 & 4 & -1 & 0 & 0 & 0 & 0 \\
-1 & 0 &  1 & 0 & 0 & 0 & 0 \\
0 & -1 & 0 & 1 & 0 & 0 & 0 \\
0 & 0 & -1 & 0 & 1 & 0 & 0 \\
0 & 0 & 0 & -1 & 0 & 1 & 0 \\
0 & 0 & 0 & 0 & -1 & 0 & 1 \\
0 & 0 & 0 & 0 & 1  & -4 & 3 \\
\end{bmatrix}
\end{equation} 
The optimality condition (Sylvester equation) in~\cite{Harker:2015a} involves the following second-order operator ${\mathbf{D}_u}^{\top} \mathbf{D}_u$:
\begin{equation}
{\mathbf{D}_u}^\top \mathbf{D}_u = \frac14 \begin{bmatrix}
10 & -12 & 2 & 0 & 0 & 0 & 0 \\
-12 & 17 & -4 & -1 & 0 & 0 & 0 \\
2 & -4 & 3 & 0 & -1 & 0 & 0 \\
\textbf{0} & \textbf{-1} & \textbf{0} & \textbf{2} & \textbf{0} & \textbf{-1} & \textbf{0} \\
0 & 0 & -1 & 0 & 3 & -4 & 2 \\
0 & 0 & 0 & -1  & -4 & 17 & -12 \\
0 & 0 & 0 & 0 & 2  & -12 & 10 
\end{bmatrix}
\end{equation} 
The bolded values of this matrix indicate that computation of the second-order derivatives for the fourth pixel does not involve the third and fifth pixels. On the other hand, with the proposed operator defined in Equation~\eqref{eq:def_b}, the second-order operator always involves the ``correct'' neighborhood: 
\begin{equation}
{\mathbf{D}_u}^\top \mathbf{D}_u = \begin{bmatrix}
1 & -1 & 0 & 0 & 0 & 0 & 0 \\
-1 & 2 & -1 & 0 & 0 & 0 & 0 \\
0 & -1 & 2 & -1 & 0 & 0 & 0 \\
\textbf{0} & \textbf{0} & \textbf{-1} & \textbf{2} & \textbf{-1} & \textbf{0} & \textbf{0} \\
0 & 0 & 0 & -1 & 2 & -1 & 0 \\
0 & 0 & 0 & 0 & -1 & 2 & -1 \\
0 & 0 & 0 & 0 & 0  & -1 & 1 
\end{bmatrix}
\end{equation}
}

In addition, as predicted by the complexity analysis in {Subsection}~\ref{sec:smooth_resolution}, our solution relying on preconditioned conjugate gradient iterations has an asymptotic complexity ($O(5n \, \log(n)\, \log(1/\epsilon)$)) {which is inbetween that of the Sylvester equations approach~\cite{Harker:2015a} ($O(n^{1.5})$) and of DCT~\cite{Simchony:1990a} ($O(n\, \log(n))$)}. The CPU times of our method and of the DCT solution, measured using Matlab codes running on a recent i7 processor, actually seem proportional: according to {this} complexity analysis, we guess the proportionality {factor} is around $5 \log(1/\epsilon)$. Indeed, with $\epsilon = 10^{-4}$, which is the value we used in our experiments, $5 \log(1/\epsilon) \approx 46$, which is consistent with the second graph in {Figure}~\ref{fig:courbes_smooth}.

\begin{figure}[!ht]
\begin{center}
	\begin{tabular}{c}
		\includegraphics[width = 0.98 \linewidth,trim={0 0 0 2cm},clip]{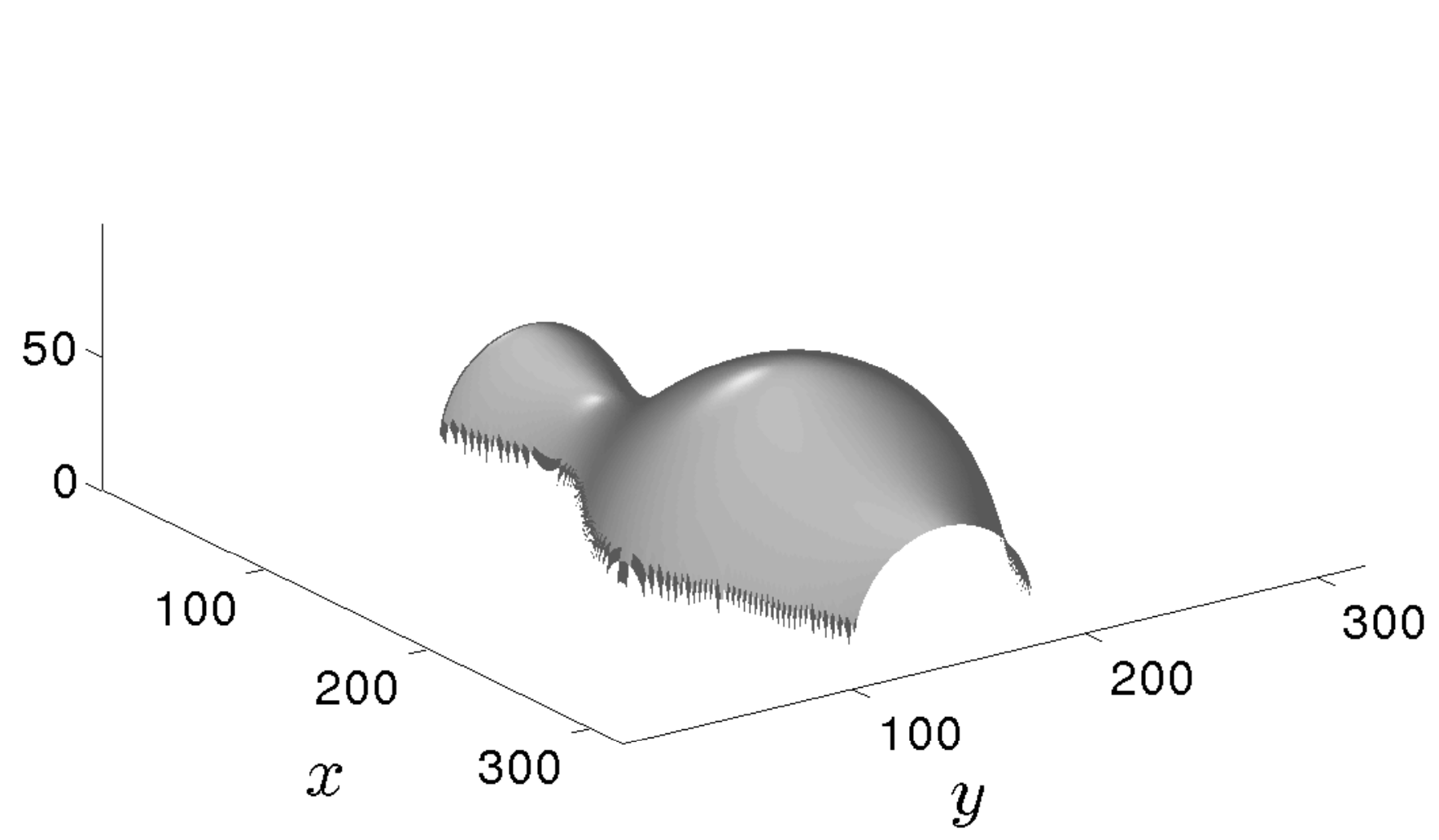} \\
		RMSE $= 0.11$ \\
		\includegraphics[width = 0.98 \linewidth,trim={0 0 0 2cm},clip]{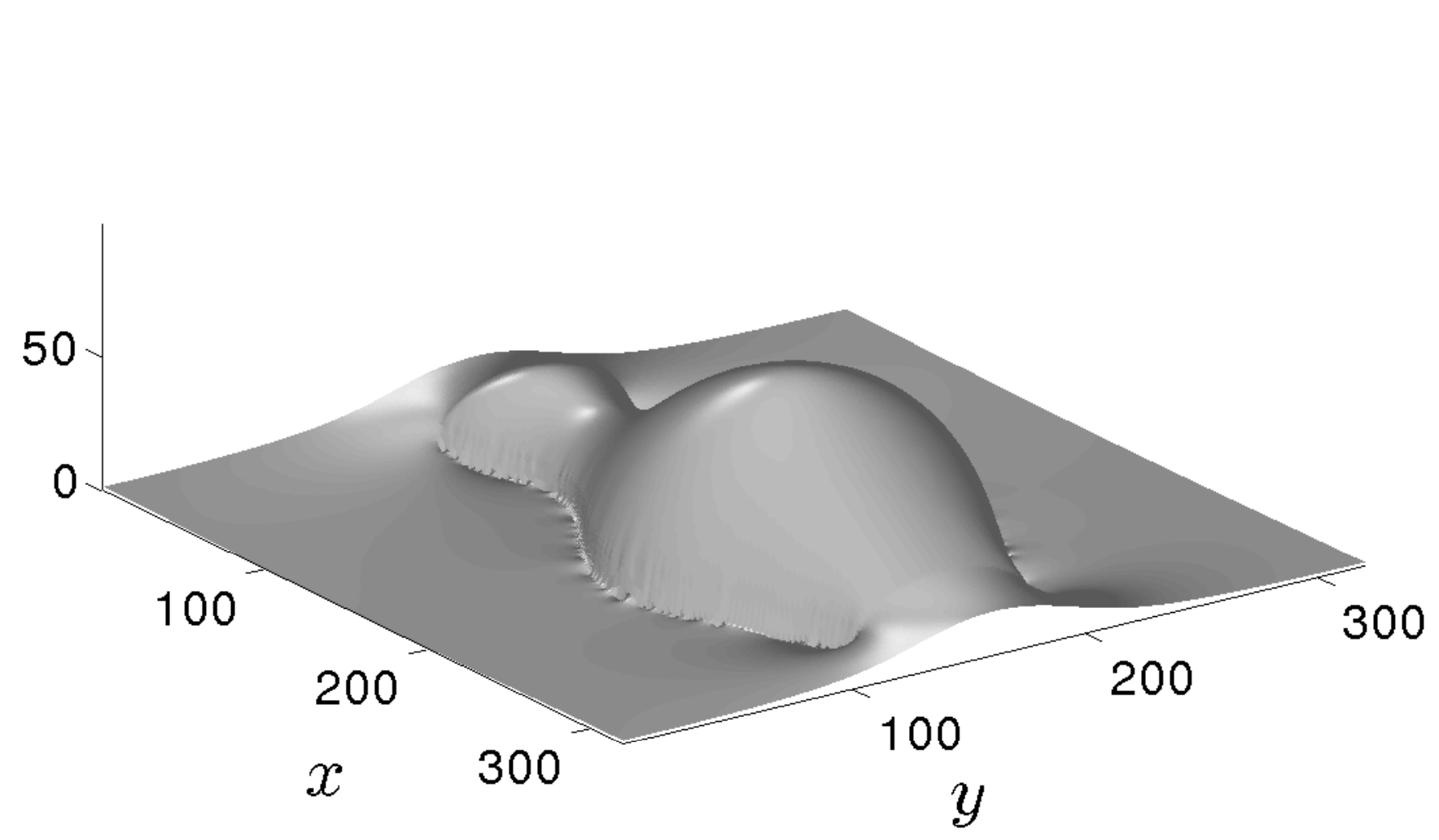} \\
		RMSE $= 4.66$
	\end{tabular}
\end{center}
\caption{3D-reconstruction of surface $\mathcal{S}_\text{vase}$ (see Figure~3 in~\cite{Durou:2016a}) from its (analytically known) normals, using the proposed discrete least-squares method. Top: when $\Omega$ is restricted to the image of the vase. Bottom: when $\Omega$ is the whole rectangular grid. Quadratic integration smooths the depth discontinuities and produces Gibbs phenomena near the kinks.}
\label{fig:vase_smooth}
\end{figure}

Besides its improved accuracy, the major advantage of our method over~\cite{Harker:2015a,Simchony:1990a} is its ability to handle non-rectangular domains ($\mathcal{P}_{\text{NoRect}}$). This makes possible the 3D-reconstruction of piecewise-smooth surfaces, provided that a user segments the domain into pieces where $z$ is smooth beforehand (see {Figure}~\ref{fig:vase_smooth}). Yet, if the segmentation is not {performed \emph{a priori}}, artifacts are visible near the discontinuities, which get smoothed, and Gibbs phenomena appear near the continuous, yet non-differentiable kinks. We will discuss in the next section several strategies for removing such artifacts.

\section{Piecewise Smooth Surfaces}
\label{sec:nonsmooth}

We now tackle the problem of recovering a surface which is smooth only \emph{almost} everywhere, i.e. everywhere except on a ``small" set where discontinuities and kinks are allowed. Since all the methods discussed hereafter rely on the same discretization as in {Section}~\ref{sec:smooth}, they inherit its $\mathcal{P}_{\text{FreeB}}$ and $\mathcal{P}_{\text{NoRect}}$ properties, which will not be discussed in this section. Instead, we focus on the $\mathcal{P}_{\text{Fast}}$, $\mathcal{P}_{\text{Robust}}$, $\mathcal{P}_{\text{NoPar}}$, and of course $\mathcal{P}_{\text{Disc}}$ properties.

\subsection{Recovering Discontinuities and Kinks}
\label{sec:disc}

\begin{figure}[!ht]
\begin{center}
	\def\svgwidth{0.95 \linewidth}
		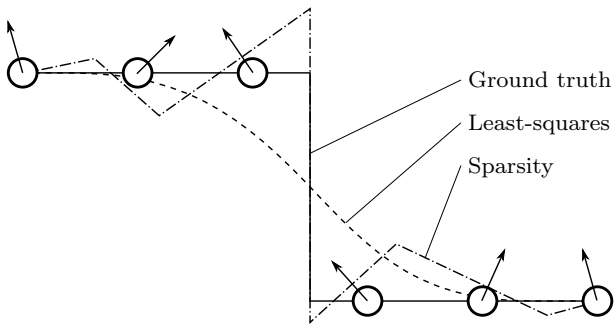
\end{center}
\caption{1D-illustration of integration of a noisy normal field (arrows) over a regular grid (circles), in the presence of discontinuities. The least-squares approach is robust to noise, but smooths the discontinuities. The sparsity approach preserves the discontinuities, but is not robust to noise. An ideal integration method would inherit robustness from least-squares, and the ability to preserve discontinuities from sparsity.}
\label{fig:schemaDisc}
\end{figure}

In order to clarify which variational formulations may provide robustness to discontinuities, let us first consider the 1D-example of {Figure}~\ref{fig:schemaDisc}, with Dirichlet boundary conditions. As illustrated in this example, least-squares integration of a noisy normal field will provide a smooth surface. 
Replacing the least-squares estimator $\mathrm{\Phi}_{L_2}(s) = s^2$ by the sparsity one $\mathrm{\Phi}_{L_0}(s) = 1-\delta(s)$ will minimize the cardinality of the difference between $\mathbf{g}$ and $\nabla z$, which provides a surface whose gradient is almost everywhere equal to $\mathbf{g}$. As a consequence, robustness to noise is lost, yet discontinuities may be preserved.

These estimators can be interpreted {as follows:} least-squares assume that \emph{all} residuals defined by $\| \nabla z(u,v) - \mathbf{g}(u,v) \|$ are ``low", while sparsity assumes that \emph{most of} them are ``zero". The former is commonly used for ``noise", and the latter for ``outliers". In the case of normal integration, outliers may occur when: 1) $\nabla z(u,v)$ exists but its estimate $\mathbf{g}(u,v)$ is not reliable; 2) $\nabla z(u,v)$ is not defined because $(u,v)$ lies within the vicinity of a discontinuity or a kink. Considering that situation 1) should rather be handled by robust estimation of the gradient~\cite{Ikehata:2014a}, we deal only with the second one, {and use the terminology ``discontinuity" instead of ``outlier", although this also covers the concept of ``kink".}

\begin{figure}[!ht]
\begin{center}
	\includegraphics[width = 0.95\linewidth]{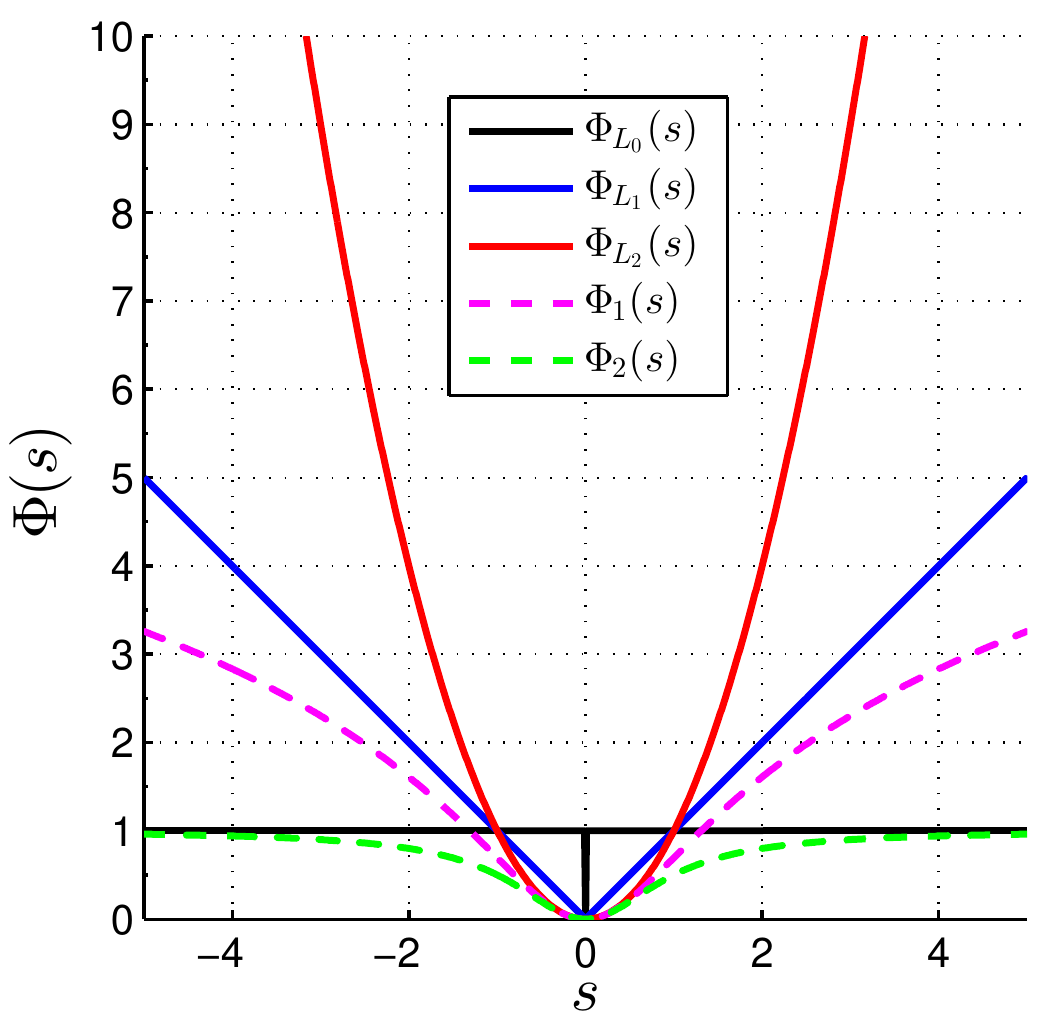}
\end{center}
\caption{Graph of some robust estimators. The ability of $\mathrm{\Phi}_{L_2}$ to handle noise (small residuals) comes from its over-linear behavior around zero, while that of $\mathrm{\Phi}_{L_0}$ to preserve discontinuities (large residuals) is induced by its sub-linear behavior in $+ \infty$. An estimator holding both these properties is necessarily non-convex (e.g., $\mathrm{\Phi}_1$ and $\mathrm{\Phi}_2$, whose graphs are shown with $\beta = \gamma = 1$), although $\mathrm{\Phi}_{L_1}$ may be an acceptable convex compromise.}
\label{fig:courbesDisc}
\end{figure}

We are looking for an estimator which combines the robustness of least-squares to noise, and that of sparsity to discontinuities. These abilities are actually due to their asymptotic behaviors. Robustness of least-squares to noise comes from the quadratic behavior around $0$, which ensures that ``low" residuals are considered as ``good" estimates, while this quadratic behavior becomes problematic in $\pm \infty$: discontinuities yield ``high" residuals, which are over-penalized. The sparsity estimator has the opposite behavior: treating the high residuals (discontinuities) exactly as the low ones ensures that discontinuities are not over-penalized, yet low residuals (noise) are. A good estimator would thus be quadratic around zero, but sub-linear around $\pm \infty$. Obviously, only non-convex estimators hold both these properties. We will discuss several choices ``inbetween" the quadratic estimator $\mathrm{\Phi}_{L_2}$ and the sparsity one $\mathrm{\Phi}_{L_0}$ (see {Figure}~\ref{fig:courbesDisc}): the convex compromise $\mathrm{\Phi}_{L_1}(s) = |s|$ is studied in {Subsection}~\ref{sec:TV}, and the non-convex estimators $\mathrm{\Phi}_1(s) = \log(s^2+\beta^2) $ and $\mathrm{\Phi}_2(s) = \frac{s^2}{s^2+\gamma^2}$, where $\beta$ and $\gamma$ are hyper-parameters, in {Subsection}~\ref{sec:nonconvex}.

Another strategy consists in keeping least-squares as basis, but using it in a non-uniform manner. The simplest way would be to remove the discontinuity points from the integration domain $\Omega$, and then to apply our quadratic method from the previous section, since it is able to manage non-rectangular domains. Yet, this would require {detecting the discontinuities beforehand}, which might be tedious. It is actually more convenient to introduce weights in the least-squares functionals, {which are} inversely proportional to the probability of lying on a discontinuity~\cite{Queau:2015b,Saracchini:2012a}. We discuss this weighted least-squares approach in {Subsection}~\ref{sec:PM}, where a statistical interpretation of the Perona and Malik's anisotropic diffusion model~\cite{Perona_Malik} is also exhibited. Eventually, an extreme case of weighted least-squares consists in using binary weights, where the weights indicate the presence of discontinuities. This is closely related to Mumford and Shah's segmentation method~\cite{Mumford_Shah}, which simultaneously estimates the discontinuity set and the surface. We show in {Subsection}~\ref{sec:MS} that this approach is the one which is actually the most adapted to the problem of integrating a noisy normal field in the presence of discontinuities.

\subsection{Total Variation-like Integration}
\label{sec:TV}

The problem of handling outliers in a noisy normal field has been tackled by Du, Robles-Kelly and Lu, who compare in~\cite{Du:2007a} the performances of several M-estimators. They conclude that regularizers based on the $L_1$ norm are the most effective ones. We provide in this subsection several numerical considerations regarding the discretization of the $L_1$ fidelity term:
\begin{align}
	\mathcal{F}_{L_1}(z) & = \iint\displaylimits_{(u,v) \in \Omega} \|\nabla z(u,v) - \mathbf{g}(u,v)\|_1 \mathrm{d}u\,\mathrm{d}v \nonumber \\
	& = \iint\displaylimits_{(u,v) \in \Omega} \Big\{ |\partial_u z(u,v) - p(u,v)| \nonumber \\
	& \,\,\,\,\,\,\,\,\,\,\,\,\,\,\,\,\,\,\,\,\, + |\partial_v z(u,v) - q(u,v)| \Big\} \mathrm{d}u\,\mathrm{d}v
\label{eq:anisTV}
\end{align}

When $p(u,v) \equiv 0$ and $q(u,v) \equiv 0$, \eqref{eq:anisTV} is the so-called ``anisotropic total variation" (anisotropic TV) regularizer, which tends to favor piecewise-constant solutions while allowing discontinuity jumps. Considering the discontinuities and kinks as the equivalent of edges in image restoration, it seems natural to believe that the fidelity term~\eqref{eq:anisTV} may be useful for discontinuity-preserving integration.

This fidelity term is not only convex, but also decouples the two directions $u$ and $v$, which allows fast ADMM-based (Bregman iterations) numerical schemes involving shrinkages~\cite{Goldstein:2009a,Queau:2015b}. On the other hand, it is not so natural to use such a decoupling: if the value of $p$ is not reliable at some point $(u,v)$, usually that of $q$ is not reliable either. Hence, it may be wortwhile to use instead a regularizer adapted from the ``isotropic TV". This leads us to adapt the {well-known} model from Rudin, Osher and Fatemi~\cite{Rudin:1992} to the integration problem:
\begin{align}
	& \mathcal{E}_{\text{TV}}(z) = \iint\displaylimits_{(u,v) \in \Omega} \|\nabla z(u,v) - \mathbf{g}(u,v) \| \nonumber \\
	& \qquad\qquad\qquad + \lambda(u,v)\left[z(u,v)-z^0(u,v)\right]^2 \! \mathrm{d}u\,\mathrm{d}v
\end{align}

\paragraph{Discretization.} Since the term $\|\nabla z(u,v) - \mathbf{g}(u,v) \|$ can be interpreted in different manners, depending on the neighborhood of $(u,v)$, we need to discretize it appropriately. Let us consider all four possible first-order discretizations of the gradient $\nabla z$, associated to the four following sets of pixels:
\begin{align}
	\Omega^{UV} = \Omega_u^U \cap \Omega_v^V,~(U,V) \in \{+,-\}^2
\end{align}
The discrete functional to minimize is thus given by:
\begin{align}
	& E_{\text{TV}}(\mathbf{z}) \!=\! \frac14\Bigg(\!\!\mathop{\sum\sum}_{(u,v) \in \Omega^{++}} \!\! \sqrt{ \left[ \partial_u^+ z_{u,v} \!-\! p_{u,v} \right]^2 \!\!\!+ \! \left[ \partial_v^+ z_{u,v} \!-\! q_{u,v} \right]^2} \nonumber \\
	& \qquad\quad + \mathop{\sum\sum}_{(u,v) \in \Omega^{+-}} \!\! \sqrt{ \left[ \partial_u^+ z_{u,v} \!-\! p_{u,v} \right]^2 \!\!\!+ \! \left[ \partial_v^- z_{u,v} \!-\! q_{u,v} \right]^2} \nonumber \\
	& \qquad\quad+ \mathop{\sum\sum}_{(u,v) \in \Omega^{-+}} \!\! \sqrt{ \left[ \partial_u^- z_{u,v} \!-\! p_{u,v} \right]^2 \!\!\!+ \! \left[ \partial_v^+ z_{u,v} \!-\! q_{u,v} \right]^2} \nonumber \\
	& \qquad\quad+ \mathop{\sum\sum}_{(u,v) \in \Omega^{--}} \!\! \sqrt{ \left[ \partial_u^- z_{u,v} \!-\! p_{u,v} \right]^2 \!\!\!+ \! \left[ \partial_v^- z_{u,v} \!-\! q_{u,v} \right]^2} \Bigg)\nonumber \\
	& \qquad\quad+ \mathop{\sum\sum}_{(u,v) \in \Omega} \lambda_{u,v} \left[z_{u,v}-z^0_{u,v} \right]^2
\label{eq:minimini}
\end{align}

Minimizing~\eqref{eq:minimini} comes down to solving the following constrained optimization problem:
\begin{align}
	& \underset{\mathbf{z}, \{\mathbf{r}^{UV} \} }{\min}\quad \frac14\!\!\mathop{\sum\sum}_{(U,V) \in \{+,-\}^2} \mathop{\sum\sum}_{(u,v) \in \Omega^{UV}} \|\mathbf{r}_{u,v}^{UV}\| \nonumber \\
	& \qquad\qquad + \mathop{\sum\sum}_{(u,v) \in \Omega} \lambda_{u,v} \left[z_{u,v}-z^0_{u,v} \right]^2 \nonumber \\
	& \text{s.t.~} \quad \mathbf{r}_{u,v}^{UV} = \nabla^{UV} z_{u,v} - \mathbf{g}_{u,v}
\label{eq:Const}
\end{align}
where we denote $\nabla^{UV} = [\partial_u^U,\partial_v^V]^\top,\,(U,V) \in \{+,-\}^2$, the discrete approximation of the gradient corresponding to domain $\Omega^{UV}$.

\paragraph{Numerical Solution.} We solve the constrained optimization problem~\eqref{eq:Const} by the augmented Lagrangian method, through an ADMM algorithm~\cite{Gabay:1976} (see~\cite{Boyd:2011a} for a recent overview of such algorithms). This algorithm reads:
\begin{align}
	& \mathbf{z}^{(k+1)} = \underset{\mathbf{z} \in \mathbb{R}^{|\Omega|}}{\operatorname{argmin}}~ \frac{\alpha}{8} \!\!\! \mathop{\sum\sum}_{(U,V) \in \{+,-\}^2} \mathop{\sum\sum}_{(u,v) \in \Omega^{UV}} \!\! \Big\|\nabla^{UV}\!\!z_{u,v} \nonumber \\
	& \qquad\qquad\qquad\qquad\qquad\qquad \!-\!\left( \mathbf{g}_{u,v} \!+\! {\mathbf{r}^{UV}_{u,v}}^{(k)} \!\!\!\!-\! {\mathbf{b}^{UV}_{u,v}}^{(k)} \right) \Big\|^2 \nonumber \\
	& \qquad\qquad\qquad+ \mathop{\sum\sum}_{(u,v) \in \Omega} \lambda_{u,v} \left[z_{u,v}-z^0_{u,v} \right]^2 \label{eq:ADMM1} \\
	& { \mathbf{r}_{u,v}^{UV} }^{(k+1)} \!\!= \underset{\mathbf{r} \in \mathbb{R}^{2}}{\operatorname{argmin}} \frac{\alpha}{8} \left\|\mathbf{r} \!-\! \left( \nabla^{UV}\!\!z_{u,v}^{(k+1)} \!-\! \mathbf{g}_{u,v} +\! {\mathbf{b}^{UV}_{u,v}}^{(k)}\right)\!\right\|^2 \nonumber \\
	& \qquad\qquad\qquad\qquad + \|\mathbf{r}\| \label{eq:ADMM2} \\
	& {\mathbf{b}^{UV}_{u,v}}^{(k+1)} = {\mathbf{b}^{UV}_{u,v}}^{(k)} + \nabla^{UV} z_{u,v}^{(k+1)} - \mathbf{g}_{u,v} - {\mathbf{r}_{u,v}^{UV}}^{(k+1)}
\label{eq:ADMM3}
\end{align}
where the $\mathbf{b}^{UV}$ are the scaled dual variables, and $\alpha >0$ corresponds to a descent stepsize, which is supposed to be fixed beforehand. Note that the choice of this parameter influences only the convergence rate, not the actual minimizer. In our experiments, we used $\alpha =1$.

\begin{figure*}[!ht]
\begin{center}
	\begin{tabular}{ccc}
		\includegraphics[width = 0.33\linewidth,trim={0 0 0 2cm},clip]{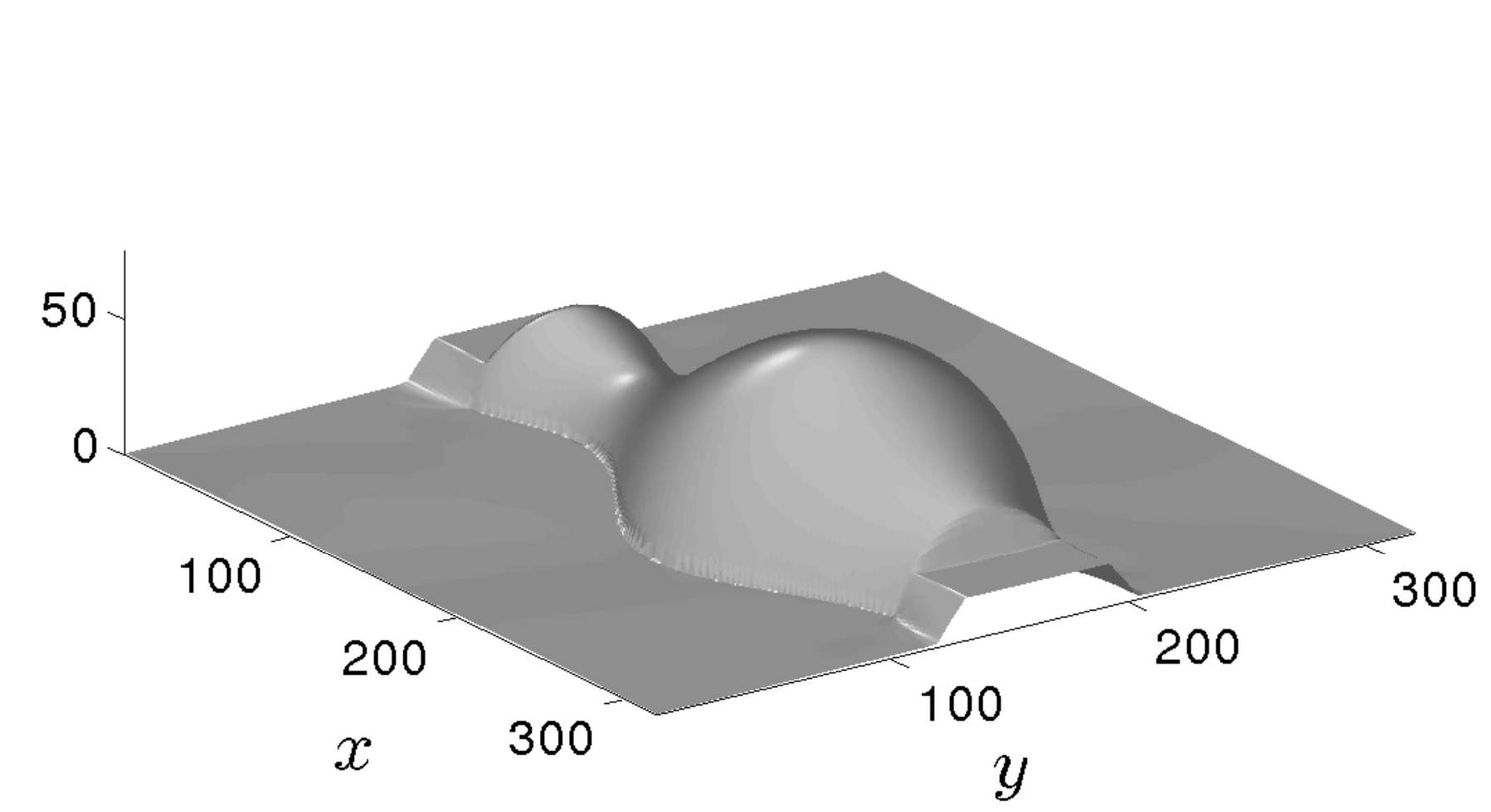} \!\!\!\!\!\! & \!\!\!\!\!\! 
		\includegraphics[width = 0.33\linewidth,trim={0 0 0 2cm},clip]{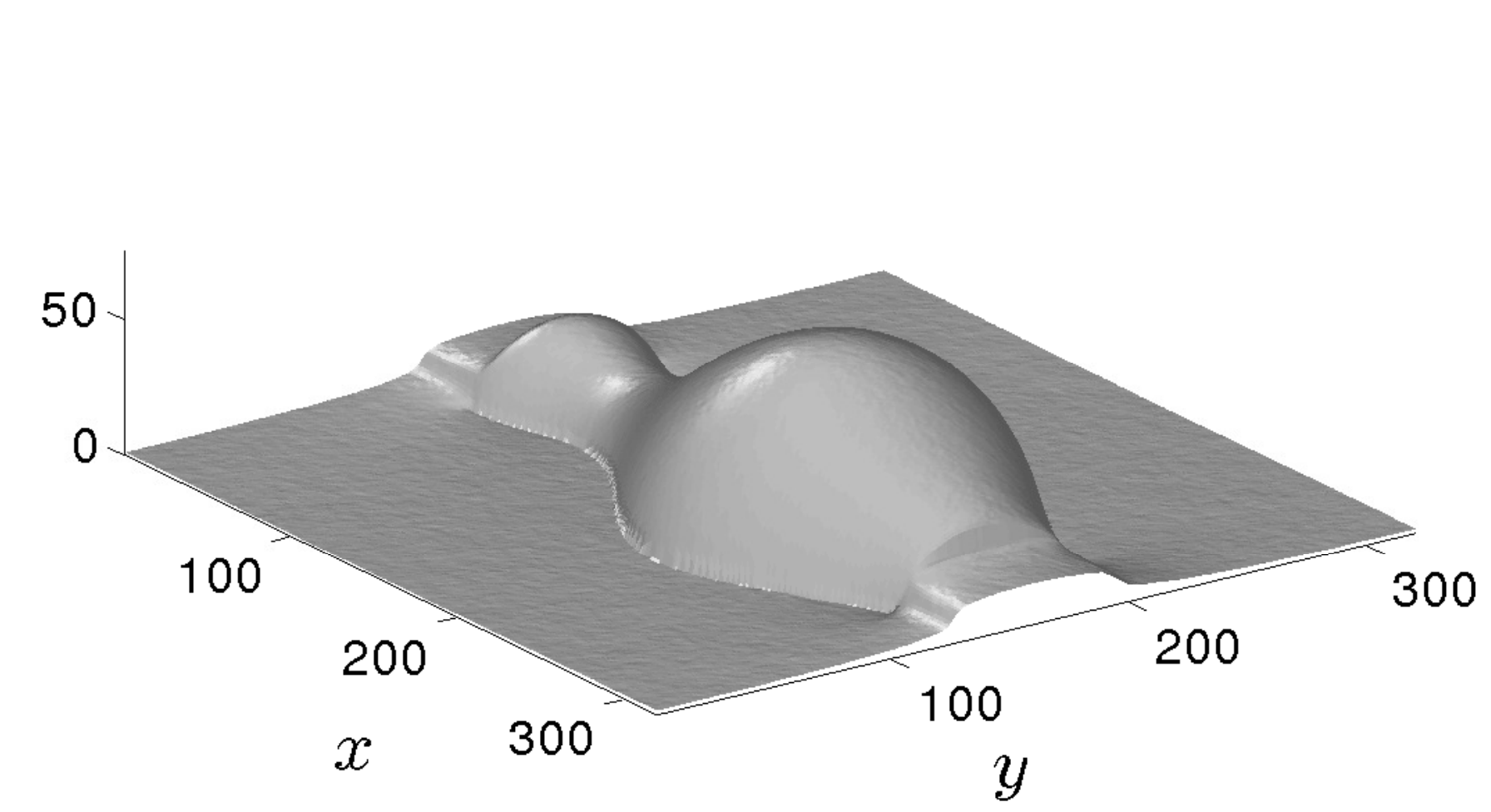} \!\!\!\!\!\!&\!\!\!\!\!\!
		\includegraphics[width = 0.33\linewidth,trim={0 0 0 2cm},clip]{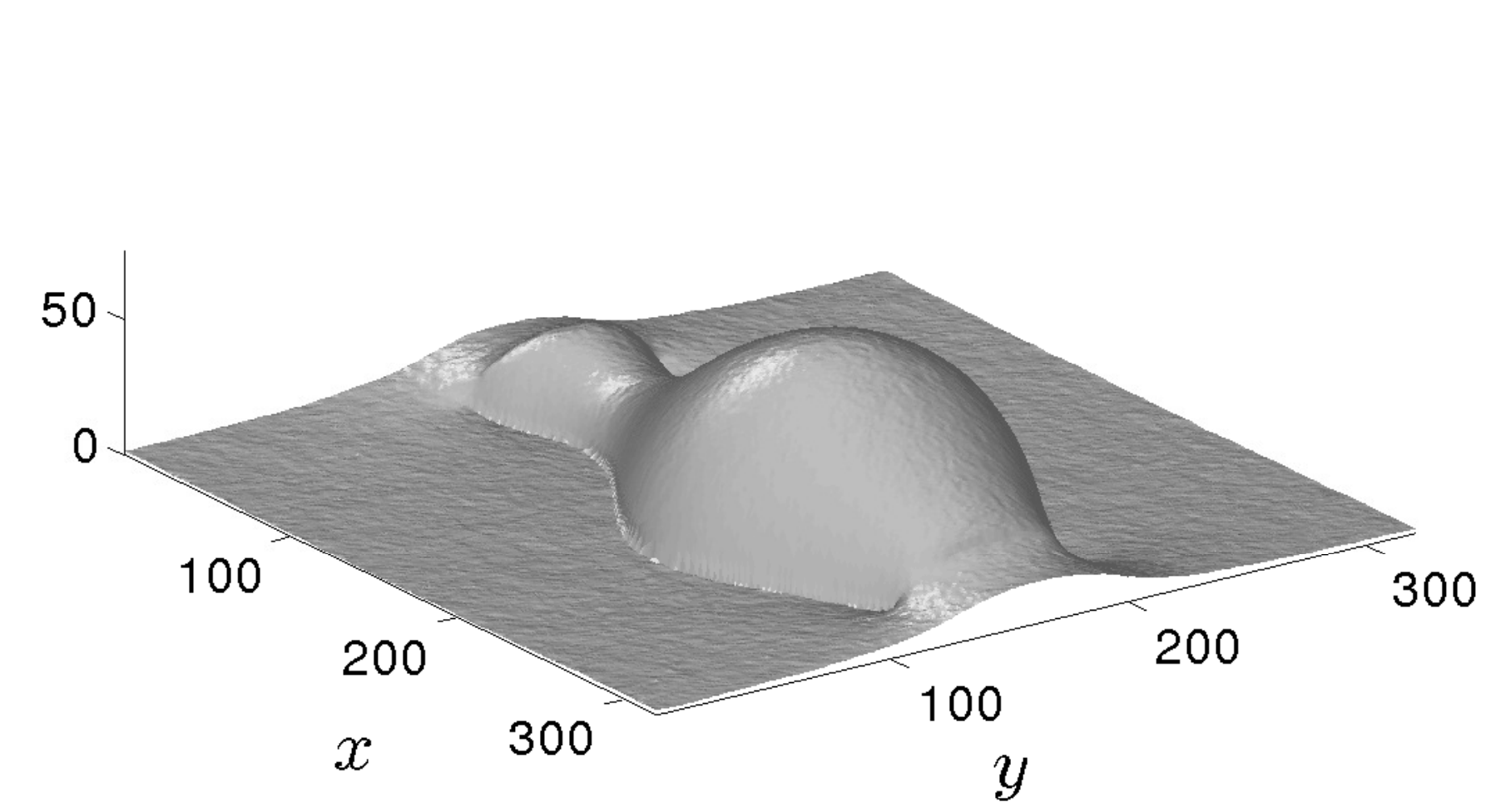} \\
		$\sigma = 0\%$ - RMSE $= 4.52$ \!\!\!\!\!\!&\!\!\!\!\!\!
		$\sigma = 0.5\%$ - RMSE $= 4.62$ \!\!\!\!\!\!&\!\!\!\!\!\!
		$\sigma = 1\%$ - RMSE $= 4.79$ 
	\end{tabular}
\end{center}
\caption{Depth estimated after $1000$ iterations of the TV-like approach, in the presence of additive, zero-mean, Gaussian noise with standard deviation equal to $\sigma \|\mathbf{g}\|_\infty$. The indicated RMSE is computed on the whole domain. In the absence of noise, both discontinuities and kinks are restored, although staircasing artifacts appear. In the presence of noise, the discontinuities are smoothed. Yet, the 3D-reconstruction near the kinks is still more satisfactory than the least-squares one: Gibbs phenomena are not visible, unlike in the second row of {Figure}~\ref{fig:vase_smooth}. }
\label{fig:TV_par}
\end{figure*}

The $z$-update~\eqref{eq:ADMM1} is a linear least-squares problem simimilar to the one which was tackled in {Section}~\ref{sec:smooth}. Its solution $\mathbf{z}^{(k+1)}$ is the solution of the following SDD linear system:
\begin{equation}
	\mathbf{A}_{\text{TV}} \mathbf{z}^{(k+1)} = \mathbf{b}_{\text{TV}}^{(k)}
\label{eq:LinSolveTV}
\end{equation}
with :
\begin{align}
	& \mathbf{A}_{\text{TV}} \!=\! \frac{\alpha}{8} \!\!\! \mathop{\sum\sum}_{(U,V) \in \{+,-\}^2} \!\!\!\!\!\!\! \Big[ { \mathbf{D}_u^U }^\top \mathbf{D}_u^U + { \mathbf{D}_v^V }^\top \mathbf{D}_v^V \Big] \!\!+\! {\bm \Lambda}^2 \\
	& \mathbf{b}_{\text{TV}}^{(k)} \!=\!\! \frac{\alpha}{8}\!\!\!\!\!\!\! \mathop{\sum\sum}_{(U,V) \in \{+,-\}^2}\!\!\!\!\!\!\! \Big[ {\mathbf{D}_u^U }^\top \! {\mathbf{p}^{UV}}^{(k)} \!\!\!+ { \mathbf{D}_v^V }^\top \! {\mathbf{q}^{UV}}^{(k)} \Big] \!\!+\!\! {\bm \Lambda}^2 \mathbf{z}^0
\end{align}
where the $\mathbf{D}_{u/v}^{U/V}$ matrices are defined as in~\eqref{eq:DupDupDup}, the ${\bm \Lambda}$ matrix as in~\eqref{eq:14}, and where we denote ${\mathbf{p}^{UV}}^{(k)}$ and ${\mathbf{q}^{UV}}^{(k)}$ the components of $\mathbf{g} + {\mathbf{r}^{UV}}^{(k)} - {\mathbf{b}^{UV}}^{(k)} $.

The solution of System~\eqref{eq:LinSolveTV} can be approximated by conjugate gradient iterations, {choosing at each iteration the previous estimate $\mathbf{z}^{(k)}$ as initial guess} (setting $\mathbf{z}^{(0)}$, for instance, as the least-squares solution from {Section}~\ref{sec:smooth}). {In addition,}
the matrix $\mathbf{A}_{\text{TV}}$ is always the same: this allows computing the preconditioner only once.

Eventually, the $\mathbf{r}$-updates~\eqref{eq:ADMM2}, $(u,v) \in \Omega$, are basis pursuit problems~\cite{Chen1998}, which admit the following closed-form solution (generalized shrinkage):
\begin{equation}
	{ \mathbf{r}_{u,v}^{UV} }^{(k+1)} \!\!=\! \max\left\{\|{\mathbf{s}^{UV}_{u,v}}^{(k+1)}\|-\frac{4}{\alpha},0\right\} \frac{ {\mathbf{s}^{UV}_{u,v}}^{(k+1)}}{\| {\mathbf{s}^{UV}_{u,v}}^{(k+1)}\|}
\end{equation}
with:
\begin{equation}
	{\mathbf{s}^{UV}_{u,v}}^{(k+1)} = \nabla^{UV}z_{u,v}^{(k+1)} - \mathbf{g}_{u,v} + {\mathbf{b}^{UV}_{u,v}}^{(k)}
\end{equation}

\paragraph{Discussion.} This TV-like approach has two main advantages: apart from the stepsize $\alpha$ which controls the speed of convergence, it does not depend on the choice of a parameter, and it is convex. The initialization has influence only on the speed of convergence, and not on the actual minimizer: convergence towards the global minimum is guaranteed~\cite{Shefi2014}. It can be shown that the convergence rate of this scheme is ergodic, and this rate can be improved rather simply~\cite{Goldstein:2014a}. We cannot consider that $\mathcal{P}_{\text{Fast}}$ is satisfied since, in comparison with the quadratic method from {Section}~\ref{sec:smooth}, yet the TV approach is ``reasonably" fast. {Possibly faster algorithms could be employed, as for instance}
the FISTA algorithm from Beck and Teboulle~\cite{Beck2009}, {or} primal-dual algorithms~\cite{Chambolle:2010a}, 
but we leave such {improvements} as future work.

On the other hand, according to the results from {Figure}~\ref{fig:TV_par}, discontinuities are recovered in the absence of noise, although staircasing artifacts appear (such artifacts are partly due to the non-differentiability of TV in zero~\cite{Nikolova}). Yet, the recovery of discontinuities is deceiving when the noise level increases. On noisy datasets, the only advantage of this approach over least-squares is thus that it removes the Gibbs phenomena around the kinks i.e., where the surface is continuous, but non-differentiable (e.g., the sides of the vase).


Because of the staircasing artifacts and of the lack of robustness to noise, we cannot find this first approach satisfactory. Yet, since turning the quadratic functional into a non-quadratic one seems to have positive influence on discontinuities recovery, we believe that exploring non-quadratic models is a promising route. Staircasing artifacts could probably be reduced by replacing total variation by total generalized variation~\cite{Bredies2015}, but we rather consider now non-convex models.

\subsection{Non-convex Regularization}
\label{sec:nonconvex}

Let us now consider non-convex estimators $\mathrm{\Phi}$ in the fidelity term \eqref{eq:3}, which are often referred to as ``$\mathrm{\Phi}$-functions''~\cite{AubertKornprobst}. As discussed in {Subsection}~\ref{sec:disc}, the choice of a specific $\mathrm{\Phi}$-function should be made according to several principles:
\begin{itemizepoint}
	\item $\mathrm{\Phi}$ should have a quadratic behavior around zero, in order to ensure that the integration is guided by the ``good" data. The typical choice ensuring this property is $\mathrm{\Phi}_{L_2}(s) = s^2$, which was discussed in {Section}~\ref{sec:smooth};
	\item $\mathrm{\Phi}$ should have a sublinear behavior at infinity, so that outliers do not have a predominant influence, and also to preserve discontinuities and kinks. The typical choice is the sparsity estimator $\mathrm{\Phi}_{L_0}(s) = 0$ if $s=0$ and $\mathrm{\Phi}_{L_0}(s) = 1$ otherwise;
	\item $\mathrm{\Phi}$ should ideally be a convex function.
\end{itemizepoint}

Obviously, it is not possible to simultaneously satisfy these three properties. The TV-like fidelity term introduced in {Subsection}~\ref{sec:TV} is a sort of ``compromise": it is the only convex function being (over-) linear in $0$ and (sub-) linear in $\pm \infty$. Although it does not depend on the choice of any hyper-parameter, we saw that it has the drawback of yielding the so-called ``staircase effect", and that discontinuities were not recovered so well in the presence of noise. If we accept to lose the {convexity} of $\mathrm{\Phi}$, we can actually design estimators which better fit both other properties. Although there may then be several minimizers, such non-convex estimators were recently shown to be very effective for image restoration~\cite{Lanza:2016}.

We will consider two classical $\mathrm{\Phi}$-functions, whose graphs are plotted in {Figure}~\ref{fig:courbesDisc}:
\begin{equation}
	\begin{cases}
		\mathrm{\Phi}_1(s) = \log(s^2+\beta^2) \\
		\mathrm{\Phi}_2(s) = \displaystyle\frac{s^2}{s^2+\gamma^2}
	\end{cases}
	\!\!\!\!\!\!\!\!\!\!\Rightarrow
	\begin{cases}
		\mathrm{\Phi}_1'(s)= \displaystyle\frac{2\,s}{s^2+\beta^2} \\
		\mathrm{\Phi}_2'(s)= \displaystyle\frac{2 \, \gamma^2 \, s}{(s^2+ \gamma^2)^2}
	\end{cases}
\label{eq:35}
\end{equation}

Let us remark that these estimators were initially introduced in~\cite{Durou:2009a} in this context, and that other non-convex estimators can be considered, based for instance on $L^p$ norms, with $0<p<1$~\cite{Badri2014}.

Let us now show how to numerically minimize the resulting functionals:
\begin{align}
	& \mathcal{E}_{\mathrm{\Phi}}(z) = 	\iint\displaylimits_{(u,v) \in \Omega} \mathrm{\Phi}\left(\| \nabla z(u,v)-\mathbf{g}(u,v) \|\right) \nonumber \\
	& \qquad\qquad\quad + \lambda(u,v) \left[ z(u,v)-z^0(u,v)\right]^2 \, \mathrm{d}u\,\mathrm{d}v
\end{align}

\paragraph{Discretization.} We consider the same discretization strategy as in {Subsection}~\ref{sec:TV}, aiming at minimizing the discrete functional:
\begin{align}
	& E_{\mathrm{\Phi}}(\mathbf{z}) = \frac14 \!\!\! \mathop{\sum\sum}_{(U,V) \in \{+,-\}^2} \mathop{\sum\sum}_{(u,v) \in \Omega^{UV}} \!\! \mathrm{\Phi}\Big(\left\|\nabla^{UV}\!\!z_{u,v}-\mathbf{g}_{u,v}\right\|\Big) \nonumber \\
	& \qquad\qquad + \mathop{\sum\sum}_{(u,v) \in \Omega} \lambda_{u,v} \left[z_{u,v}-z^0_{u,v} \right]^2
\label{eq:EPhi}
\end{align}
which resembles the TV functional defined in~\eqref{eq:minimini}, and where $\nabla^{UV}$ represents the finite differences approximation of the gradient used over the domain $\Omega^{UV}$, with $\{U,V\} \in \{+,-\}^2$.

Introducing the notations:
\begin{align}
	& f(\mathbf{z}) \!\! = \!\! \frac14 \!\!\!\!\mathop{\sum\sum}_{(U,V) \in \{+,-\}^2} \mathop{\sum\sum}_{(u,v) \in \Omega^{UV}}\!\! \mathrm{\Phi}\left( \|\nabla^{UV}\!\!z_{u,v} - \mathbf{g}_{u,v}\|\right) \label{eq:defF} \\
	& g(\mathbf{z}) = \|{\bm \Lambda} (\mathbf{z} - \mathbf{z}^0)\|^2 \label{eq:defG}
\end{align}
the discrete functional~\eqref{eq:EPhi} is rewritten:
\begin{equation}
	E_{\mathrm{\Phi}}(\mathbf{z}) = f(\mathbf{z}) + g(\mathbf{z})
\label{eq:EPhi2}
\end{equation}
where $f$ is smooth, but \emph{non-convex}, and $g$ is \emph{convex} (and smooth, although non-smooth functions $g$ could be handled).

\begin{figure*}[!ht]
\begin{center}
	\begin{tabular}{ccc}
		\includegraphics[width = 0.31\linewidth,trim={0 0 0 2cm},clip]{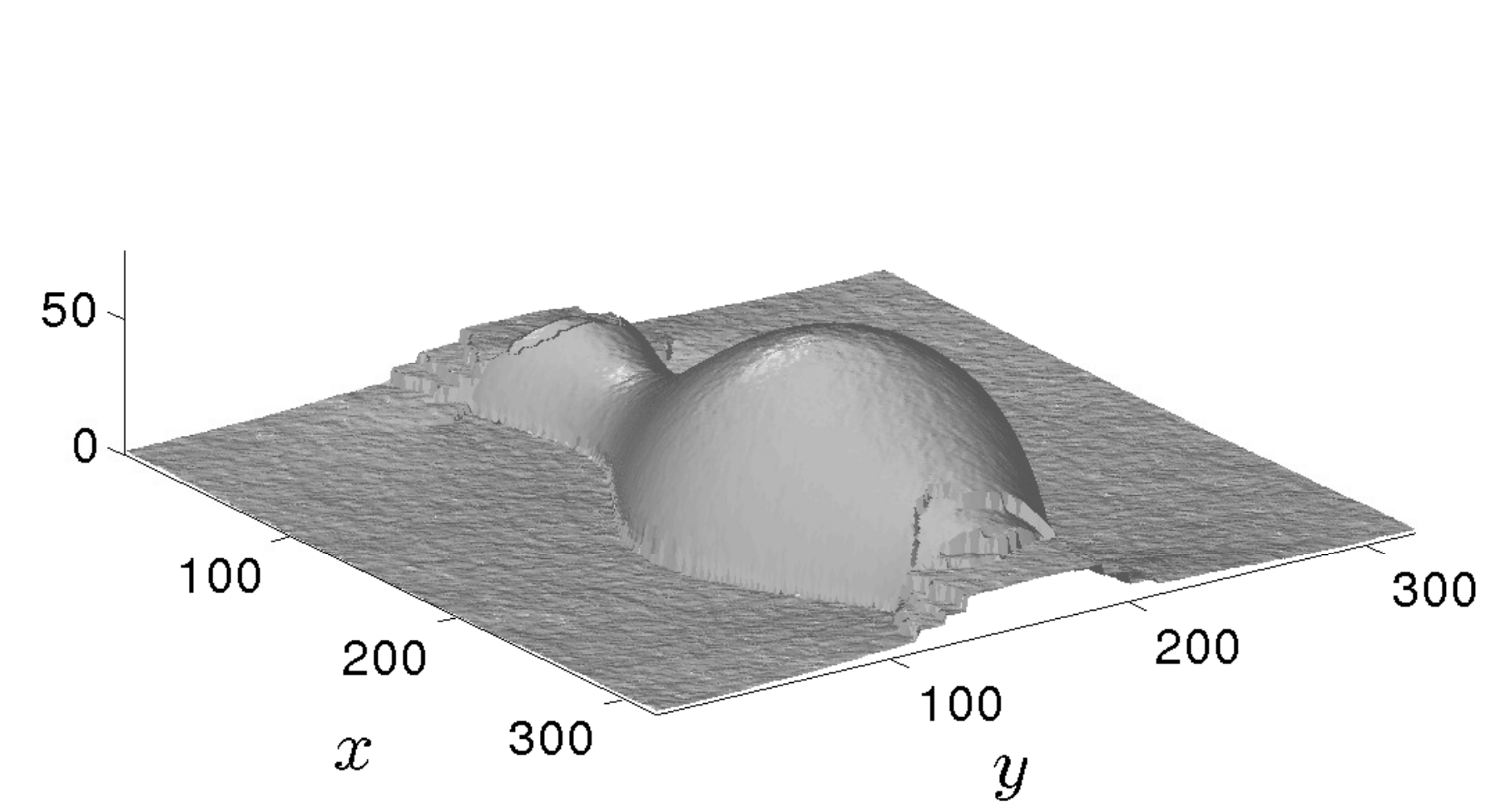} &
		\includegraphics[width = 0.31\linewidth,trim={0 0 0 2cm},clip]{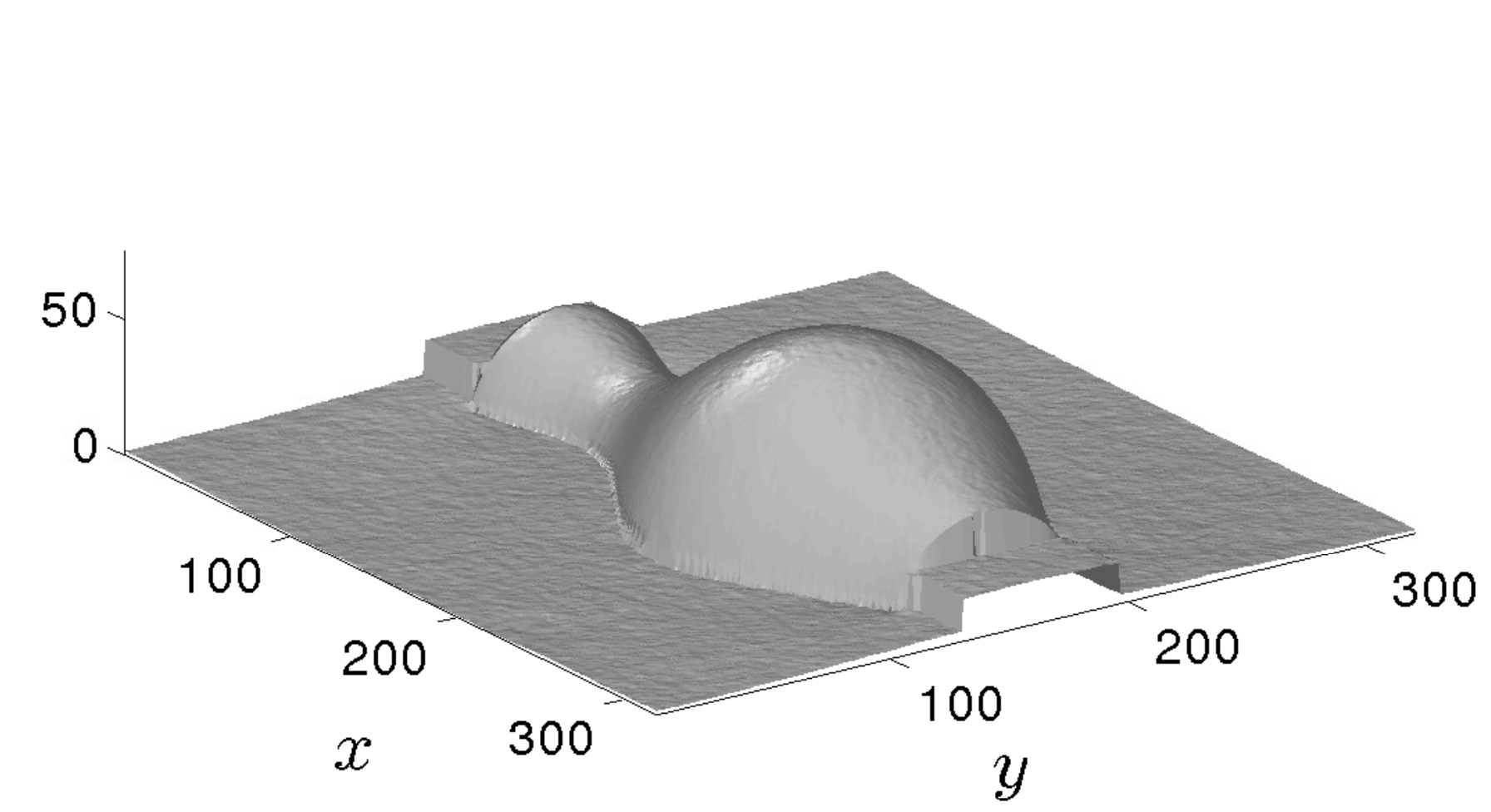} &
		\includegraphics[width = 0.31\linewidth,trim={0 0 0 2cm},clip]{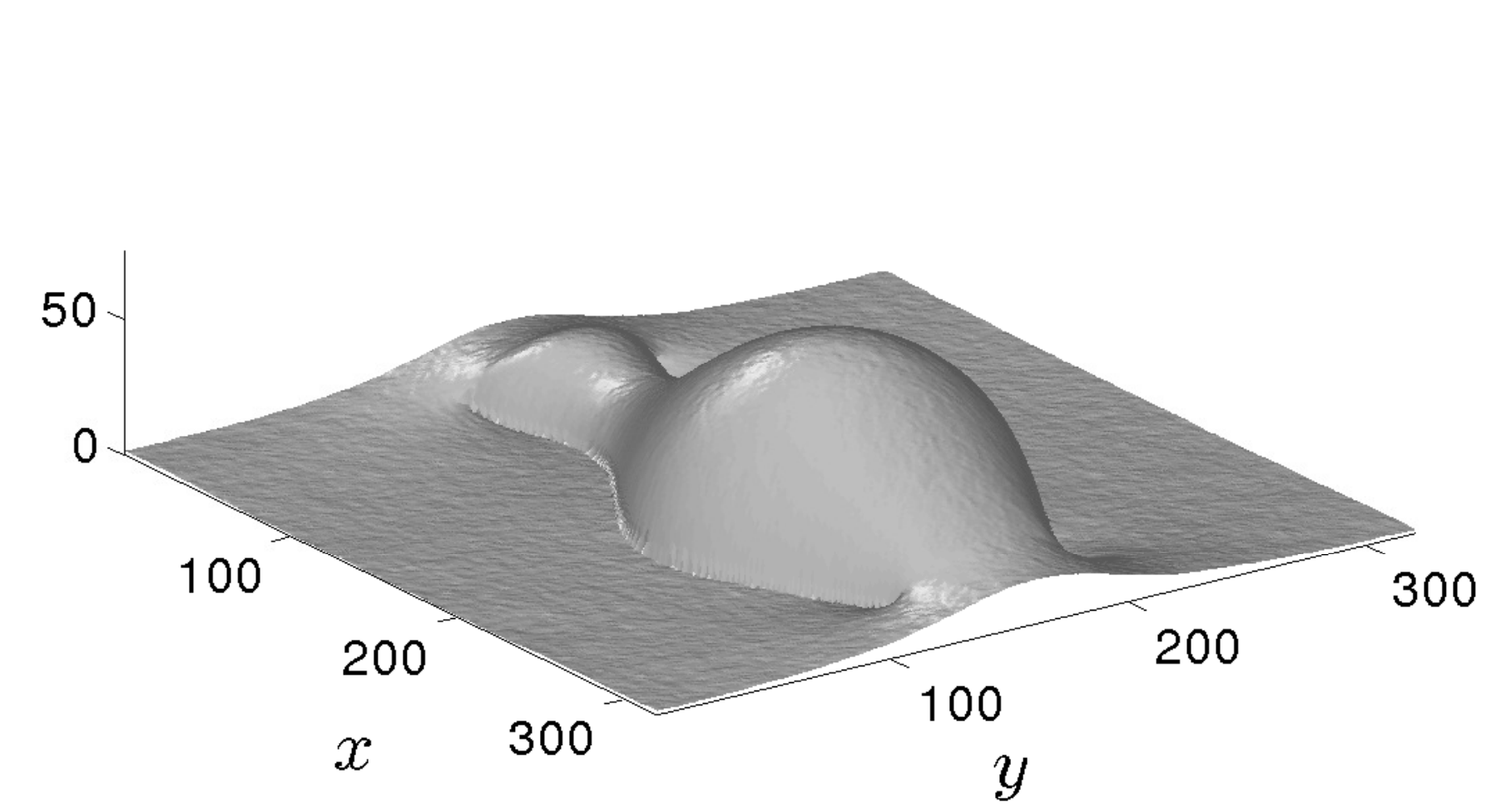} \\
		$\beta = 0.1$ - RMSE $=4.60$ &
		$\beta = 0.5$ - RMSE $=4.42$ &
		$\beta = 1$ - RMSE $=5.08$ \\
		\includegraphics[width = 0.31\linewidth,trim={0 0 0 2cm},clip]{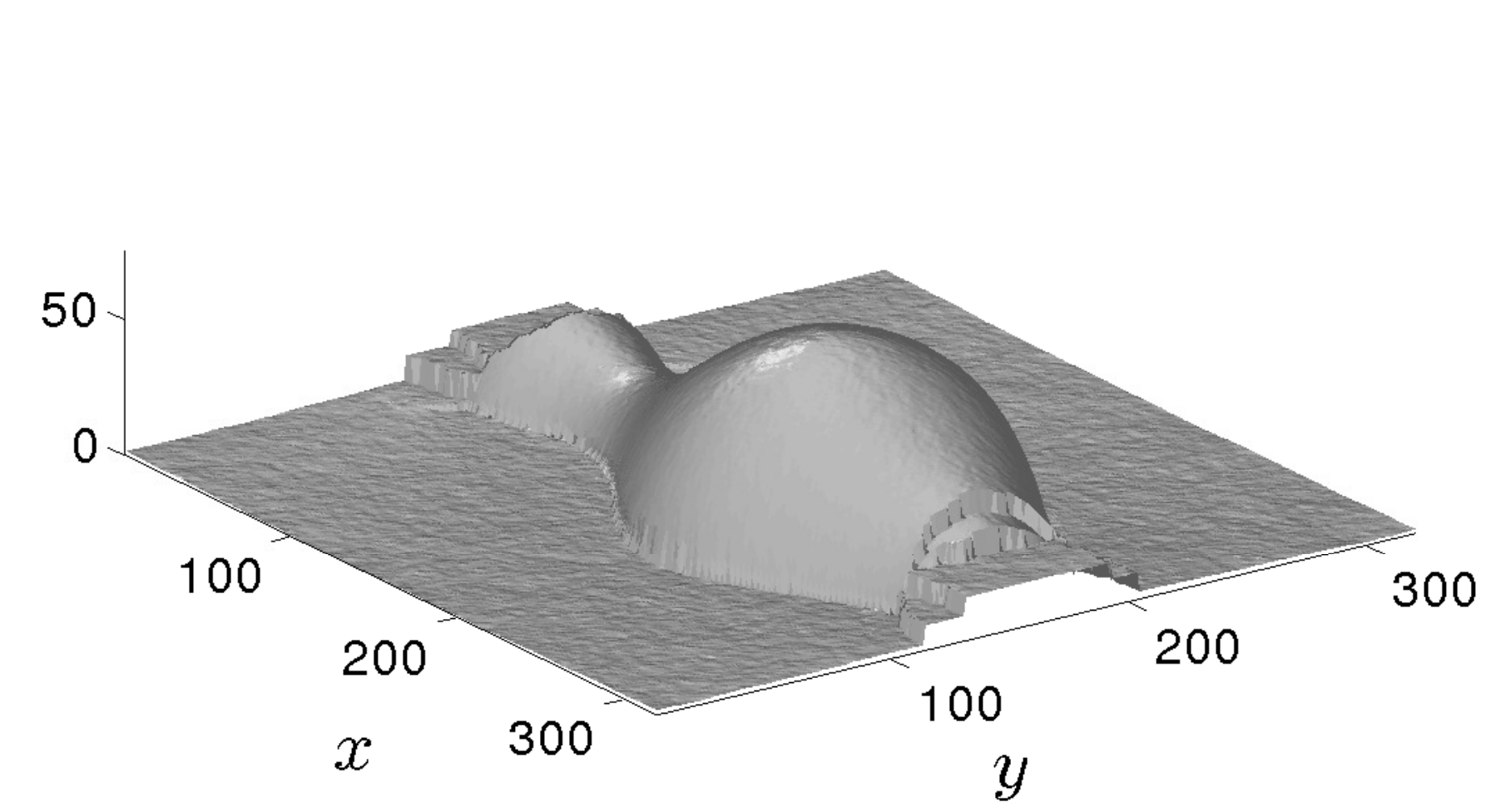} &
		\includegraphics[width = 0.31\linewidth,trim={0 0 0 2cm},clip]{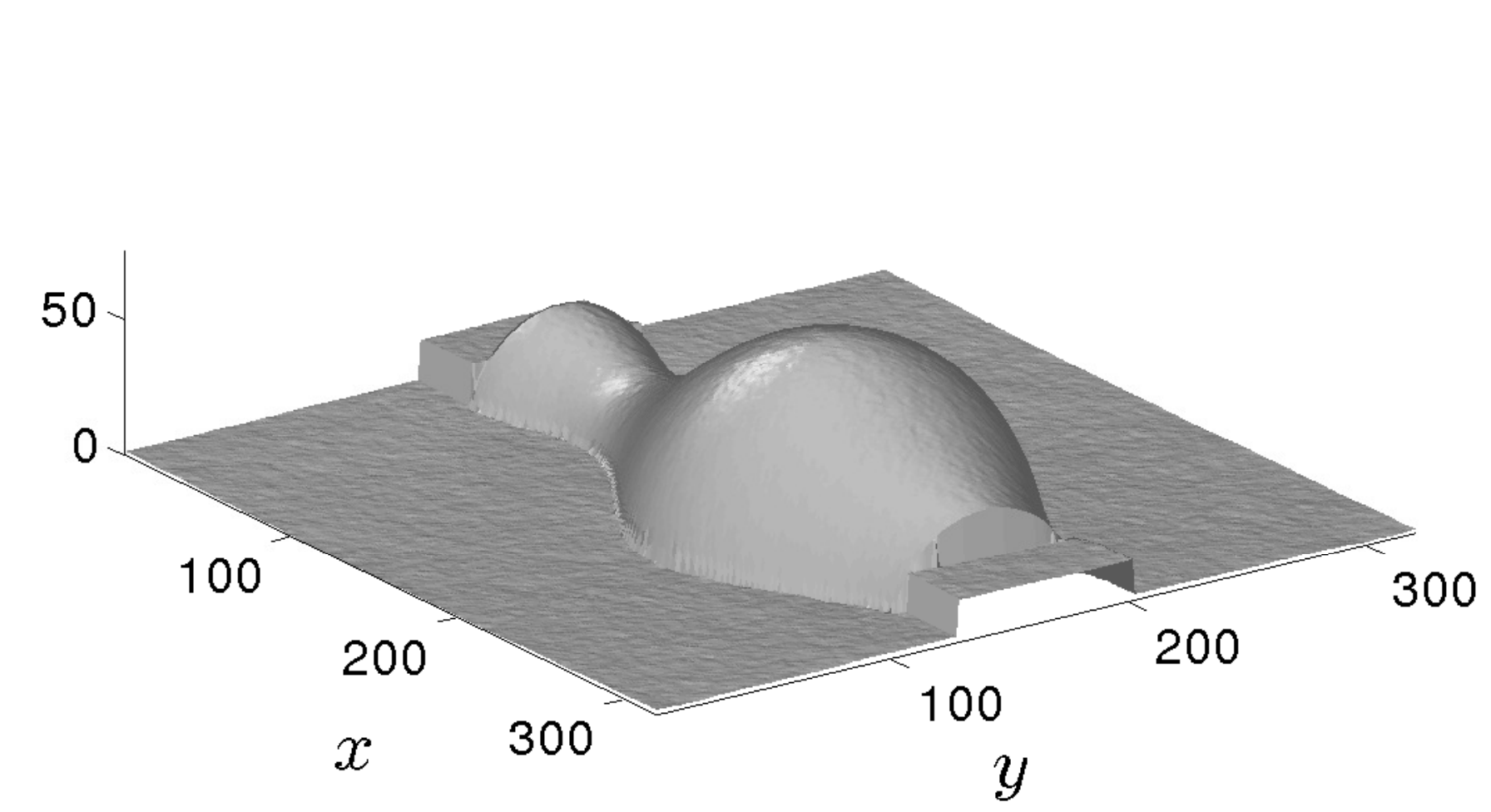} &
		\includegraphics[width = 0.31\linewidth,trim={0 0 0 2cm},clip]{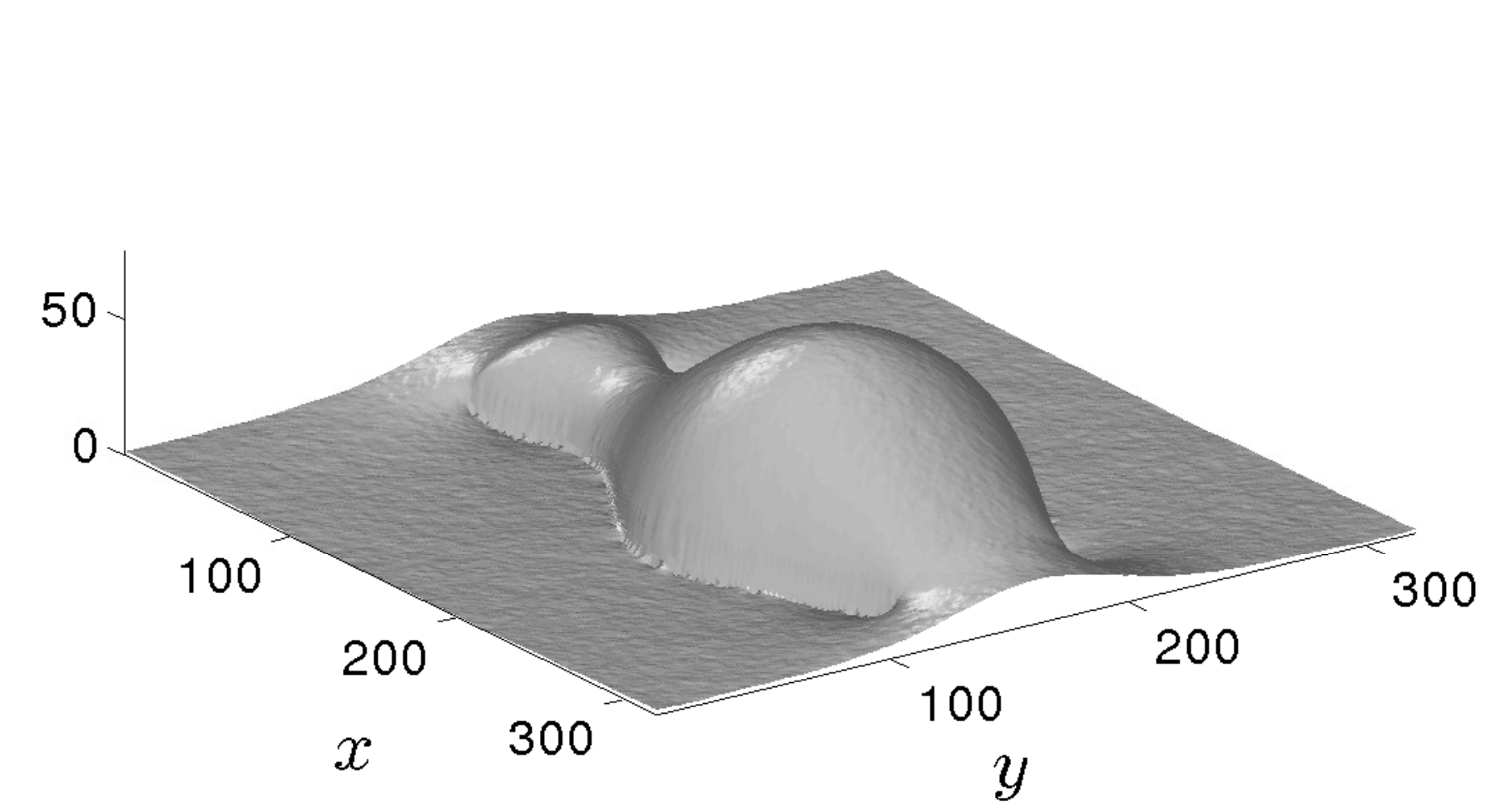} \\
		$\gamma = 0.5$ - RMSE $=4.51$ &
		$\gamma = 1$ - RMSE $=4.44$ &
		$\gamma = 5$ - RMSE $=4.67$ \\
	\end{tabular}
\end{center}
\caption{Non-convex 3D-reconstructions of surface $\mathcal{S}_\text{vase}$, using $\mathrm{\Phi}_1$ (top) or $\mathrm{\Phi}_2$ (bottom). An additive, zero-mean, Gaussian noise with standard deviation $\sigma \|\mathbf{g}\|_\infty$, $\sigma = 1\%$, was added to the gradient field. The non-convex approaches depend on the tuning of a parameter ($\beta$ or $\gamma$), but they are able to reconstruct the discontinuities in the presence of noise, unlike the TV approach. Staircasing artifacts indicate the presence of local minima (we used as initial guess $z^{(0)}$ the least-squares solution).}
\label{fig:nonconvexVase}
\end{figure*}

\paragraph{Numerical Solution.} The problem of minimizing a discrete energy like~\eqref{eq:EPhi2}, yielded by the sum of a convex term $g$ and a non-convex, yet smooth term $f$, can be handled by forward-backward splitting. We use the ``iPiano" iterative algorithm by Ochs et al.~\cite{Ochs:2014a}, which reads:
\begin{equation}
	\mathbf{z}^{(k+1)} \!\!= \! \left(\mathbf{I}\!+\!\alpha_1 \partial g\right)^{-1} \!\!\left( \!\mathbf{z}^{(k)} \!\!-\! \alpha_1 \!\nabla\! f(\mathbf{z}^{(k)}) \!+\! \alpha_2 \!\left(\!\mathbf{z}^{(k)} \!\!-\! \mathbf{z}^{(k-1)} \!\right) \!\right)
\end{equation}
where $\alpha_1$ and $\alpha_2$ are suitable descent stepsizes (in our implementation, $\alpha_2$ is fixed to 0.8, and $\alpha_1$ is chosen by the ``lazy backtracking" procedure described in~\cite{Ochs:2014a}), $ \left(\mathbf{I}+\alpha_1 \partial g\right)^{-1}$ is the \emph{proximal operator} of $g$, and $\nabla f(\mathbf{z}^{(k)})$ is the gradient of $f$ evaluated at current estimate $\mathbf{z}^{(k)}$. We detail hereafter how to evaluate the proximal operator of $g$ and the gradient of $f$.

The proximal operator of $g$ writes, using~\eqref{eq:defG}:
\begin{align}
	\left(\mathbf{I}+\alpha_1 \partial g\right)^{-1} \left(\widehat{\mathbf{x}} \right) & = \underset{\mathbf{x} \in \mathbb{R}^{|\Omega|}}{\operatorname{argmin}}~ \frac{\|\mathbf{x} - \widehat{\mathbf{x}}\|}{2} + \alpha_1 g(\mathbf{x}) \\
	~ & = \left( \mathbf{I} + 2 \alpha_1 {\bm \Lambda}^2 \right)^{-1} \left( \widehat{\mathbf{x}} + 2 \alpha_1 {\bm \Lambda} \mathbf{z}^0 \right)
\end{align}
where the inversion is easy to compute, since the matrix involved is diagonal.

In order to obtain a closed-form expression of the gradient of $f$ defined in~\eqref{eq:defF}, let us rewrite this function in the following manner:
\begin{align}
	& f(\mathbf{z}) \!=\! \frac14 \!\!\!\mathop{\sum\sum}_{(U,V) \in \{+,-\}^2} \mathop{\sum\sum}_{(u,v) \in \Omega^{UV}} \mathrm{\Phi}\left( \|\mathbf{D}_{u,v}^{UV} \mathbf{z} - \mathbf{g}_{u,v}\|\right)
\end{align}
where $\mathbf{D}_{u,v}^{UV}$ is a $2 \times |\Omega|$ finite differences matrix used for approximating the gradient at location $(u,v)$, using the finite differences operator $\nabla^{UV}$, $\{U,V\} \in \{+,-\}^2$:
\begin{equation}
	\mathbf{D}_{u,v}^{UV} =
	\begin{bmatrix}
		\left(\mathbf{D}_u^U\right)_{m^{-1}(u,v),\cdotp} \\
		\left(\mathbf{D}_v^V\right)_{m^{-1}(u,v),\cdotp}
	\end{bmatrix}
\end{equation}
where we recall that the mapping $m$ associates linear indices with pixel coordinates (see {Equation}~\eqref{eq:defM}).

The gradient of $f$ is thus given by:
\begin{align}
	& \nabla f(\mathbf{z}) = \frac14 \!\!\!\! \mathop{\sum\sum}_{(U,V) \in \{+,-\}^2} \mathop{\sum\sum}_{(u,v) \in \Omega^{UV}} \Bigg\{ {\mathbf{D}_{u,v}^{UV}}^\top \left(\mathbf{D}_{u,v}^{UV} \mathbf{z} - \mathbf{g}_{u,v} \right) \nonumber \\
	& \qquad\qquad\qquad\qquad\qquad \times \frac{\mathrm{\Phi}'\left( \|\mathbf{D}_{u,v}^{UV} \mathbf{z} - \mathbf{g}_{u,v}\|\right)}{\|\mathbf{D}_{u,v}^{UV} \mathbf{z} - \mathbf{g}_{u,v}\|} \Bigg\}
\end{align}
Given the choices~\eqref{eq:35} for the $\mathrm{\Phi}$-functions, this can be further simplified:
\begin{align}
	& \nabla f_1(\mathbf{z}) \!=\! \frac12 \!\!\! \mathop{\sum\sum}_{(U,V) \in \{+,-\}^2} \mathop{\sum\sum}_{(u,v) \in \Omega^{UV}} \frac{ {\mathbf{D}_{u,v}^{UV}}^\top \left(\mathbf{D}_{u,v}^{UV} \mathbf{z} - \mathbf{g}_{u,v} \right)}{ \|\mathbf{D}_{u,v}^{UV} \mathbf{z} - \mathbf{g}_{u,v}\|^2 + \beta^2} \\
	& \nabla f_2(\mathbf{z}) \!=\! \frac12 \!\!\! \mathop{\sum\sum}_{(U,V) \in \{+,-\}^2} \mathop{\sum\sum}_{(u,v) \in \Omega^{UV}} \frac{ \gamma^2 {\mathbf{D}_{u,v}^{UV}}^\top \left(\mathbf{D}_{u,v}^{UV} \mathbf{z} - \mathbf{g}_{u,v} \right)}{\left(\|\mathbf{D}_{u,v}^{UV} \mathbf{z} - \mathbf{g}_{u,v}\|^2 +\gamma^2 \right)^2}
\end{align}

\paragraph{Discussion.} Contrarily to the TV-like approach ({see Subsection}~\ref{sec:TV}), the non-convex estimators require setting one hyper-parameter ($\beta$ or $\gamma$). As shown in {Figure}~\ref{fig:nonconvexVase}, the choice of this parameter is crucial: when it is too high, discontinuities are smoothed, while setting a too low value leads to strong staircasing artifacts. Inbetween, the values $\beta = 0.5$ and $\gamma = 1$ seem to preserve discontinuities, even in the presence of noise (which was not the case using the TV-like approach).

\begin{figure}[!ht]
\begin{center}
	\begin{tabular}{c}
		\includegraphics[width = 0.85\linewidth,trim={0 0 0 1.8cm},clip]{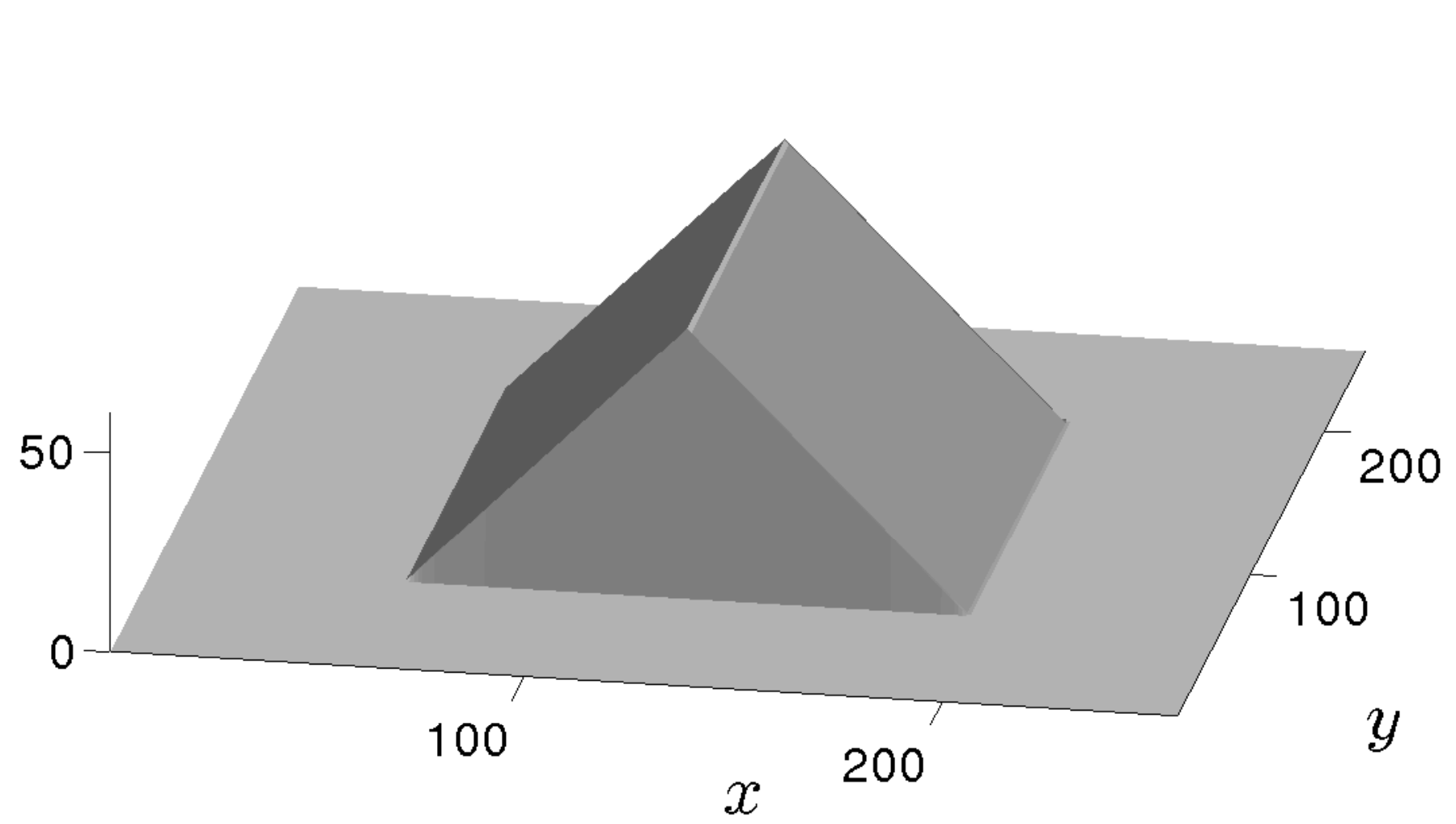} \\
		Ground-truth \\
		\includegraphics[width = 0.85\linewidth,trim={0 0 0 1.2cm},clip]{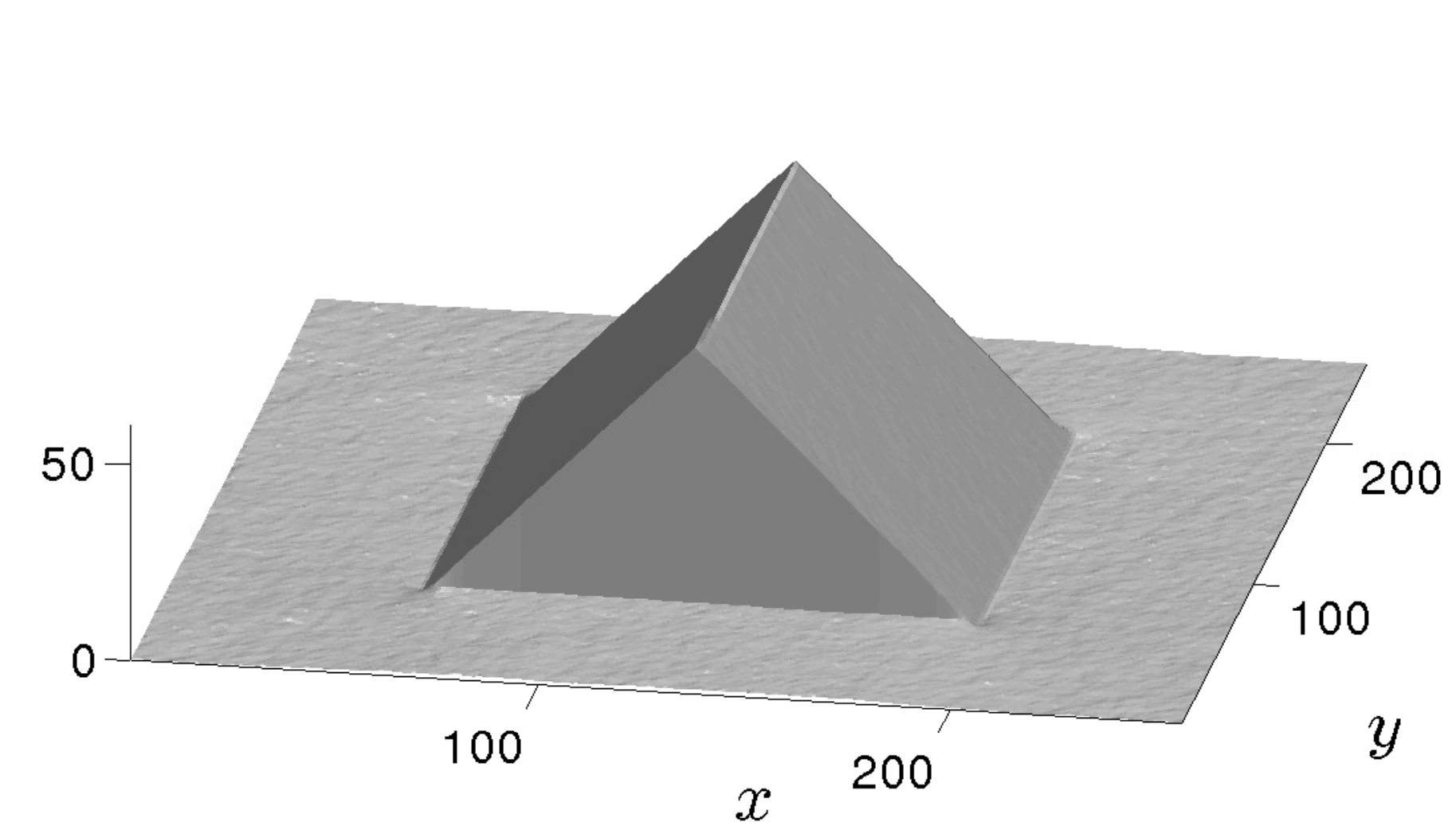} \\
		$z^{(0)} = $ least-squares solution - RMSE $= 0.78$ \\
		\includegraphics[width = 0.85\linewidth,trim={0 0 0 1.2cm},clip]{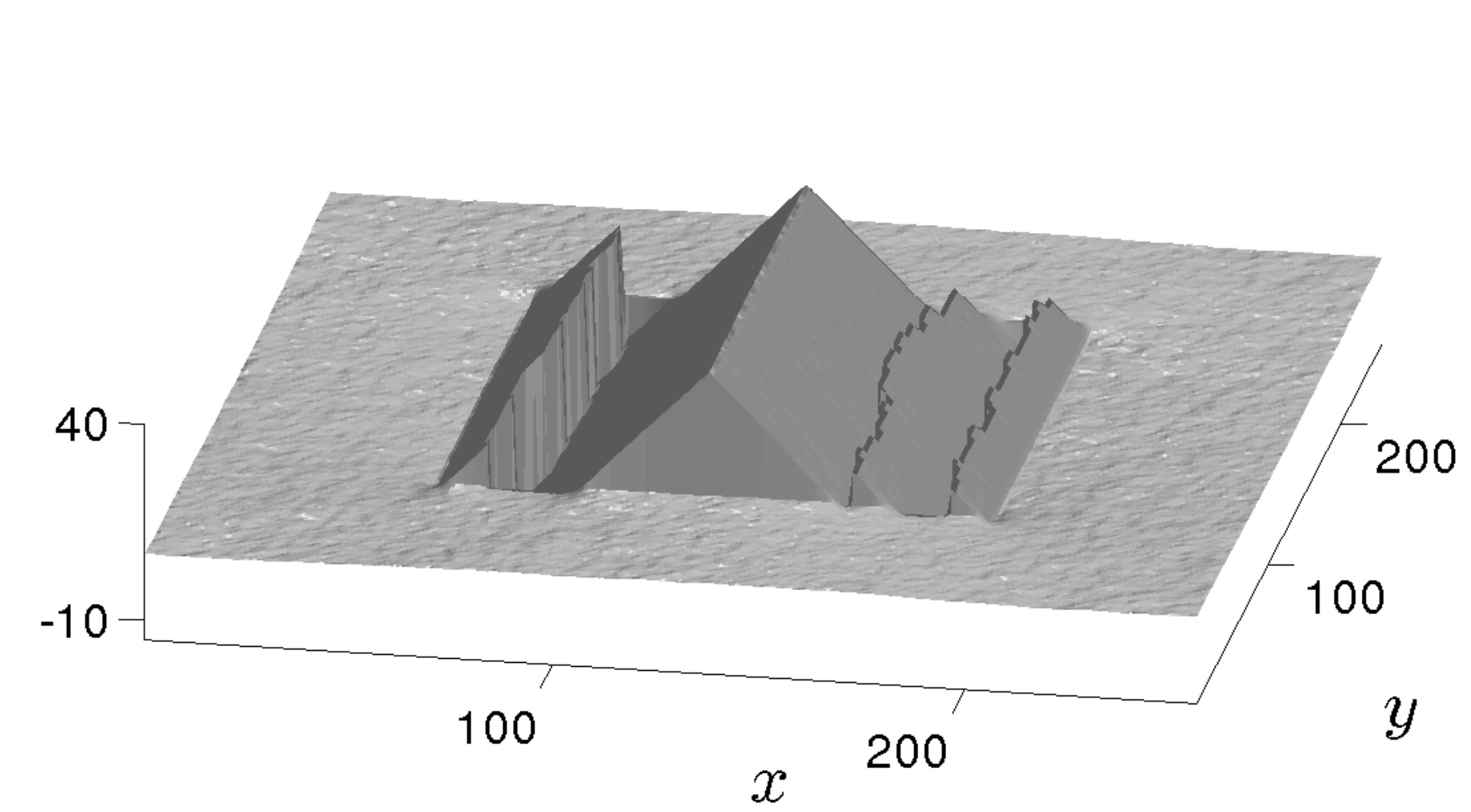} \\
		$z^{(0)} \equiv 0$ - RMSE $= 13.16$
	\end{tabular}
\end{center}
\caption{3D-reconstruction of a ``Canadian tent"-like surface from its noisy gradient ($\sigma = 1\%$), by the non-convex integrator $\mathrm{\Phi}_1$ ($\beta = 0.5$, $12000$ iterations), using two different initializations. The objective function being non-convex, the iterative scheme may converge towards a local minimum.}
\label{fig:nonconvexCT}
\end{figure}

Yet, staircasing artifacts are still present. Despite their non-convexity, the new estimators $\mathrm{\Phi}_1$ and $\mathrm{\Phi}_2$ are differentiable, hence these artifacts do not come from a lack of differentiability, as this was the case for TV. They rather indicate the presence of local minima. This is illustrated in {Figure}~\ref{fig:nonconvexCT}, where the 3D-reconstruction of a ``Canadian tent"-like surface, with additive, zero-mean, Gaussian noise ($\sigma = 10\%$), is presented. When using the least-squares solution as initial guess $z^{(0)}$, the 3D-reconstruction is very close to the genuine surface. Yet, when using the trivial initialization $z^{(0)} \equiv 0$, we obtain a surface whose slopes are ``almost everywhere" equal to the real ones, but unexpected discontinuity jumps appear. Since only the initialization differs in these experiments, this clearly shows that the artifacts indicate the presence of local minima.

Although local minima can sometimes be avoided by using the least-squares solution as initial guess (e.g., {Figure}~\ref{fig:nonconvexCT}), this is not always the case (e.g., {Figure}~\ref{fig:nonconvexVase}). Hence, the non-convex estimators perform overall better than the TV-like approach, but they are still not optimal. We now follow other routes, which use least-squares as basis estimator, yet in a non-uniform manner, in order to allow discontinuities.

\subsection{Integration by Anisotropic Diffusion}
\label{sec:PM}

Both previous methods (total variation and non-convex estimators) replace the least-squares estimator by another one, assumed to be robust to discontinuities. Yet, it is possible to proceed differently: the 1D-graph in {Figure}~\ref{fig:schemaDisc} shows that most of data are corrupted only by noise, and that the discontinuity set is ``small". Hence, applying least-squares everywhere except on this set should provide an optimal 3D-reconstruction. To achieve this, a first possibility is to consider weighted least-squares:
\begin{align}
	& \underset{z}{\min} \iint\displaylimits_{(u,v) \in \Omega} \left\|\mathbf{W}(u,v) \left[\nabla z(u,v) - \mathbf{g}(u,v)\right] \right\|^2 \nonumber \\
	& \qquad\qquad\quad + \lambda(u,v)\left[z(u,v)-z^0(u,v)\right]^2 \! \mathrm{d}u\,\mathrm{d}v
\label{eq:optiPM}
\end{align}
where $\mathbf{W}$ is a $\Omega \to \mathbb{R}^{2 \times 2}$ tensor field, acting as a weight map designed to reduce the influence of discontinuity points. The weights can be computed {beforehand} according to the integrability of $\mathbf{g}$~\cite{Queau:2015b}, or by convolution of the components of $\mathbf{g}$ by a Gaussian kernel~\cite{Agrawal:2006a}. Yet, such approaches are of limited interest when $\mathbf{g}$ contains noise. In this case, the weights should rather be set as a function inversely proportional to $\|\nabla z(u,v)\|$, e.g.:
\begin{equation}
	\mathbf{W}(u,v) = \frac{1}{\sqrt{\left(\frac{\|\nabla z(u,v)\|}{\mu}\right)^2+1}} \, \mathbf{I}_2
\label{eq:PMDiffTens}
\end{equation}
with $\mu$ a user-defined hyper-parameter. The latter tensor is the one proposed by Perona and Malik in~\cite{Perona_Malik}: the continuous optimality condition associated to~\eqref{eq:optiPM} is related to their ``anisotropic diffusion model"~\footnote{Although \eqref{eq:PMDiffTens} actually yields an isotropic diffusion model, since it ``utilizes a scalar-valued diffusivity and not a 
diffusion tensor"~\cite{Weickert1998}.}. Such tensor fields $\mathbf{W}:\,\Omega \to \mathbb{R}^{2 \times 2}$ are called ``diffusion tensors": we refer the reader to~\cite{Weickert1998} for a complete overview.

The use of diffusion tensors for the integration problem is not new~\cite{Queau:2015b}, but we provide hereafter additional comments on the statistical interpretation of such tensors. Interestingly, the diffusion tensor~\eqref{eq:PMDiffTens} also appears when making different assumptions on the noise model than those we considered so far. Up to now, we assumed that the input gradient field $\mathbf{g}$ was equal to the gradient $\nabla z$ of the depth map $z$, up to an additive, zero-mean, Gaussian noise: $\mathbf{g} = \nabla z + {\bm \epsilon}$, ${\bm \epsilon} \sim \mathcal{N}\left( [0,0]^\top,\begin{bmatrix}
	\sigma^2 & 0 \\0 & \sigma^2
\end{bmatrix}\right)$. This hypothesis may not always be realistic. For instance, in 3D-reconstruction scenarii such as photometric stereo~\cite{Woodham:1980a}, one estimates the normal field $\mathbf{n}:\Omega \to \mathbb{R}^3$ pixelwise, rather than the gradient $\mathbf{g}:\,\Omega \to \mathbb{R}^2$, from a set of images. Hence, the Gaussian assumption should rather be made on these images. In this case, and provided that a maximum-likelihood for the normals is used, it may be assumed that the estimated normal field is the genuine one, up to an additive Gaussian noise. Yet, this does not imply that the noise in the gradient field $\mathbf{g}$ is Gaussian-distributed. Let us clarify this point.

Assuming orthographic projection, the relationship between $\mathbf{n} = \left[n_1,n_2,n_3\right]^\top$ and $\nabla z$ is written, {in} every point $(u,v)$ where the depth map $z$ is differentiable:
\begin{equation}
	\mathbf{n}(u,v) = \frac{1}{\sqrt{\|\nabla z(u,v)\|^2+1}} \left[-\nabla z(u,v)^\top,\,1 \right]^\top
\label{eq:normale_gradient}
\end{equation}
which implies that $[-\frac{{n}_1}{{n}_3},-\frac{{n}_2}{{n}_3}]^\top = [\partial_u z,\partial_v z]^\top = \nabla z$. If we denote $\overline{\mathbf{n}} = \left[\overline{{n}}_1,\overline{{n}}_2,\overline{n}_3 \right]^\top$ the estimated normal field, it follows from \eqref{eq:normale_gradient} that $[-\frac{\overline{n}_1}{\overline{n}_3},-\frac{\overline{n}_2}{\overline{n}_3}]^\top = [p,q]^\top = \mathbf{g}$.

Let us assume that $\overline{\mathbf{n}}$ and $\mathbf{n}$ differ according to an additive, zero-mean, Gaussian noise:
\begin{equation}
	\overline{\mathbf{n}}(u,v) = \mathbf{n}(u,v) + {\bm \epsilon}(u,v)
\end{equation}
where :
\begin{equation}
	{\bm \epsilon}(u,v) \sim \mathcal{N}
	\left(
		[0,0,0]^\top,
		\begin{bmatrix}
			\sigma^2 & 0 & 0 \\
			0 & \sigma^2 & 0 \\
			0 & 0 & \sigma^2
		\end{bmatrix}
	\right)
\end{equation}

Since $\overline{n}_3$ is unlikely to take negative values (this would mean that the estimated surface is not oriented towards the camera), the following Geary-Hinkley transforms:
\begin{align}
	& t_1 = \frac{n_3 \left( \frac{\overline{n}_1}{\overline{n}_3} \right) - n_1 }{\sqrt{\sigma^2 \left( \left( \frac{\overline{n}_1}{\overline{n}_3} \right)^2 + 1 \right)}} \\
	& t_2 = \frac{n_3 \left( \frac{\overline{n}_2}{\overline{n}_3} \right) - n_2 }{\sqrt{\sigma^2 \left( \left( \frac{\overline{n}_2}{\overline{n}_3} \right)^2 + 1 \right)}}
\end{align}
both follow standard Gaussian distribution $\mathcal{N}(0,1)$~\cite{Hayya:1975a}. After some algebra, this can be rewritten as:
\begin{align}
	\frac{1}{\sigma \sqrt{1+p^2} \, \sqrt{\|\nabla z\|^2+1}} \left[ \partial_u z - p \right] \sim \mathcal{N}\left(0,1\right) \\
	\frac{1}{\sigma \sqrt{1+q^2} \, \sqrt{\|\nabla z\|^2+1}} \left[ \partial_v z - q \right] \sim \mathcal{N}\left(0,1\right)
\end{align}

This rationale suggests the use of the following fidelity term:
\begin{align}
	\mathcal{F}_{\text{PM}}(z) = \!\!\iint\displaylimits_{(u,v) \in \Omega}
	\!\!\left\|
	\mathbf{W}(u,v)
	\left[ \nabla z(u,v) - \mathbf{g}(u,v)\right]
	\right\|^2
	\mathrm{d}u\,\mathrm{d}v
\label{eq:FPM}
\end{align}
where $\mathbf{W}(u,v)$ is the following $2\times 2$ anisotropic diffusion tensor field:
\begin{equation}
	\mathbf{W}(u,v) \!= \!
	\small{ \frac{1}{\sqrt{\!\|\nabla z(\!u,v\!)\|^2\!+\!1}}}
	\begin{bmatrix}
		\frac{1}{\sqrt{1\!+\!p(\!u,v\!)^2}} & 0 \\
		0 & \frac{1}{\sqrt{1\!+\!q(\!u,v\!)^2}}
	\end{bmatrix}
\label{eq:anis_diff}
\end{equation}

Unfortunately, we experimentally found {with the choice}~\eqref{eq:anis_diff} for the diffusion tensor field, discontinuities were not always recovered. Instead, following the pioneering ideas from Perona and Malik~\cite{Perona_Malik}, we introduce two parameters $\mu$ and $\nu$ to control the respective influences of the terms {depending on the gradient of the unknown $\|\nabla z\|$ and on the input gradient $(p,q)$}. {The new tensor field is then given by}:
\begin{equation}
	\mathbf{W}(u,v) \! = \!\!
	\small{ \frac{1}{\sqrt{\left(\!\frac{\|\nabla z(\!u,v\!)\|}{\mu}\!\right)^2\!\!\!+\!\!1}}}
	\begin{bmatrix}
		\frac{1}{\sqrt{1\!+\!\left(\!\frac{p(\!u,v\!)}{\nu}\!\right)^2}} & 0 \\
		0 & \frac{1}{\sqrt{1\!+\!\left(\!\frac{q(\!u,v\!)}{\nu}\!\right)^2}}
	\end{bmatrix}
\label{eq:anis_diff2}
\end{equation}
Replacing the matrix in~\eqref{eq:anis_diff2} by $\mathbf{I}_2$ yields exactly
 the Perona-Malik diffusion tensor~\eqref{eq:PMDiffTens}, which reduces the influence of the fidelity term on locations $(u,v)$ where $\|\nabla z(u,v)\|$ increases, which are likely to indicate discontinuities. Yet, our diffusion tensor~\eqref{eq:anis_diff2} also reduces the influence of points where $p$ or $q$ is high, which are also likely to correspond to discontinuities. In our experiments, we found that $\nu = 10$ could always be used, yet the choice of $\mu$ has more influence on the actual results.

\paragraph{Discretization.} Using the same discretization strategy as in Subsections~\ref{sec:TV} and~\ref{sec:nonconvex} leads us to the following discrete functional:
\begin{align}
	& E_{\text{PM}}(\mathbf{z}) = \frac14 \mathop{\sum\sum}_{(U,V) \in \{+,-\}^2} \!\! \Bigg\{ \left\| \mathbf{A}^{UV}(\mathbf{z}) \left( \mathbf{D}_{u}^U \mathbf{z} \!-\! \mathbf{p} \right) \right\|^2 \! \nonumber \\
	& \qquad\qquad\qquad\qquad\quad+ \! \left\| \mathbf{B}^{UV}(\mathbf{z}) \left( \mathbf{D}_{v}^V \mathbf{z} \!-\! \mathbf{q} \right) \right\|^2 \Bigg\} \nonumber \\
	& \qquad\qquad\qquad\qquad\quad + \left\|{\bm \Lambda} \left( \mathbf{z}-\mathbf{z}^0 \right)\right\|^2
\label{eq:EPM}
\end{align}
where the $\mathbf{A}^{UV}(\mathbf{z})$ and $\mathbf{B}^{UV}(\mathbf{z})$ are $|\Omega|\times|\Omega|$ diagonal matrices containing the following values:
\begin{align}
	a^{UV}_{u,v} = \frac{1}{\sqrt{1+\left(\frac{p_{u,v}}{\nu}\right)^2} \, \sqrt{\frac{\left(\partial_u^Uz_{u,v}\right)^2+\left(\partial_v^Vz_{u,v}\right)^2}{\mu^2}+1}} \\
	b^{UV}_{u,v} = \frac{1}{\sqrt{1+\left(\frac{q_{u,v}}{\nu}\right)^2} \, \sqrt{\frac{\left(\partial_u^Uz_{u,v}\right)^2+\left(\partial_v^Vz_{u,v}\right)^2}{\mu^2}+1}}
\end{align}
with $(U,V) \in \{+,-\}^2$.


\paragraph{Numerical Solution.} Since the coefficients $a^{UV}_{u,v}$ and $b^{UV}_{u,v}$ depend in a nonlinear way on the unknown values $z_{u,v}$, it is difficult to derive a closed-form expression for the minimizer of~\eqref{eq:EPM}. To deal with this issue, we use the following fixed point scheme, which iteratively updates the anisotropic diffusion tensors and the $z$-values:
\begin{align}
	& \mathbf{z}^{(k+1)} = \underset{\mathbf{z} \in \mathbb{R}^{|\Omega|}}{\operatorname{argmin}} \frac14\!\! \mathop{\sum\sum}_{(U,V) \in \{+,-\}^2} \!\!\Bigg\{ \left\| \mathbf{A}^{UV}(\mathbf{z}^{(k)}) \left( \mathbf{D}_{u}^U \mathbf{z} \!-\! \mathbf{p} \right) \right\|^2 \! \nonumber \\
	& \qquad\qquad\qquad\qquad\qquad\qquad+ \! \left\| \mathbf{B}^{UV}(\mathbf{z}^{(k)}) \left( \mathbf{D}_{v}^V \mathbf{z} \!-\! \mathbf{q} \right) \right\|^2 \Bigg\} \nonumber \\
	& \qquad\qquad\qquad\qquad\qquad\qquad + \left\|{\bm \Lambda} \left( \mathbf{z}-\mathbf{z}^0 \right)\right\|^2
\label{eq:LinPM}
\end{align}
Now that the diffusion tensor coefficients are fixed, each optimization problem~\eqref{eq:LinPM} is reduced to a simple linear least-squares problem. In our implementation, we solve the corresponding optimality condition using Cholesky factorization, which we experimentally found to provide more stable results than conjugate gradient iterations.

\paragraph{Discussion.} We first experimentally verify that the proposed anisotropic diffusion approach is indeed a statistically meaningful approach in the context of photometric stereo. As stated in~\cite{Noakes:2003a}, ``in previous work on photometric stereo, noise is [wrongly] added to the gradient of the height function rather than camera images". Hence, we consider the images from the ``Cat" dataset presented in~\cite{Shi:2016a}, and add a zero-mean, Gaussian noise with standard deviation $\sigma \|I\|_\infty$, $\sigma = 5\%$, to the images, where $\|I\|_\infty$ is the maximum graylevel value. The normals were computed by photometric stereo~\cite{Woodham:1980a} over the part representing the cat. Then, since only the normals ground-truth is provided in~\cite{Shi:2016a}, and not the depth ground-truth, we a posteriori computed the final normal maps by central finite differences. This allows us to calculate the angular error, in degrees, between the real surface and the reconstructed one. The mean angular error (MAE) can eventually be computed over the set of pixels for which central finite differences make sense (boundary and background points are excluded).

\begin{figure}[!ht]
\begin{center}
		\begin{tabular}{ccc}
			\includegraphics[width = 0.28\linewidth]{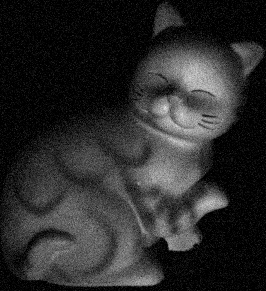} &
			\includegraphics[width = 0.28\linewidth]{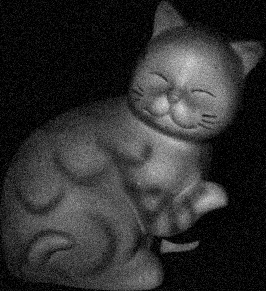} &
			\includegraphics[width = 0.28\linewidth]{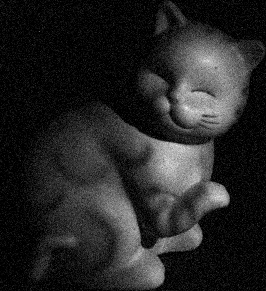}
		\end{tabular}
		\begin{tabular}{cc}
			\includegraphics[width = 0.47\linewidth,trim={0 0 0 4.2cm},clip]{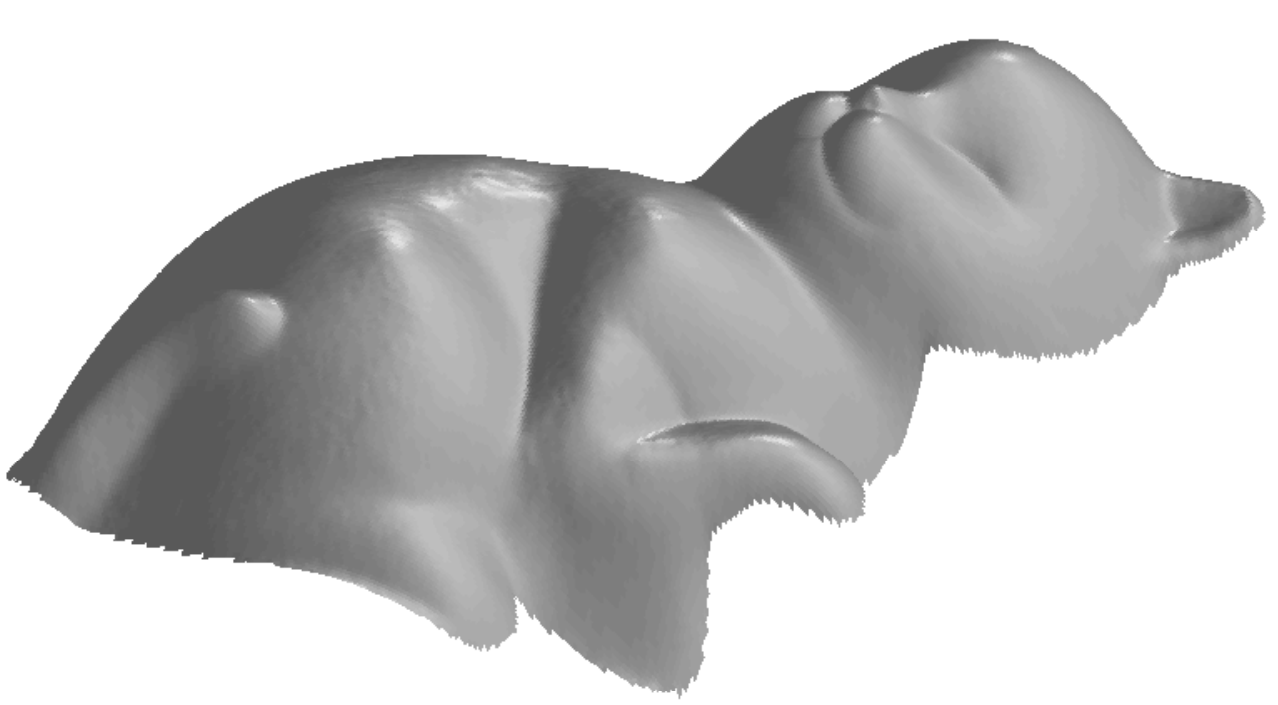} &
			\includegraphics[width = 0.47\linewidth,trim={0 0 0 4.2cm},clip]{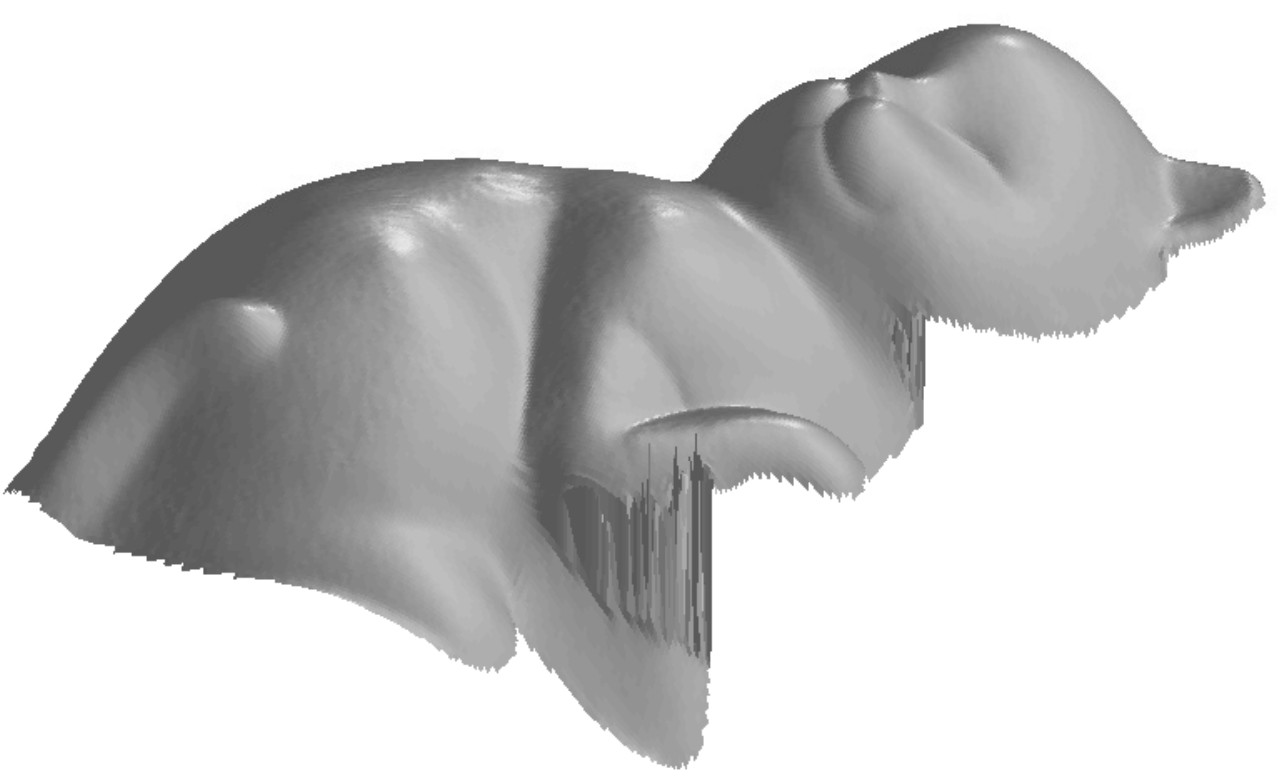} \\
			Least-squares & Anisotropic diffusion \\
			~ \\
			\includegraphics[width = 0.45\linewidth]{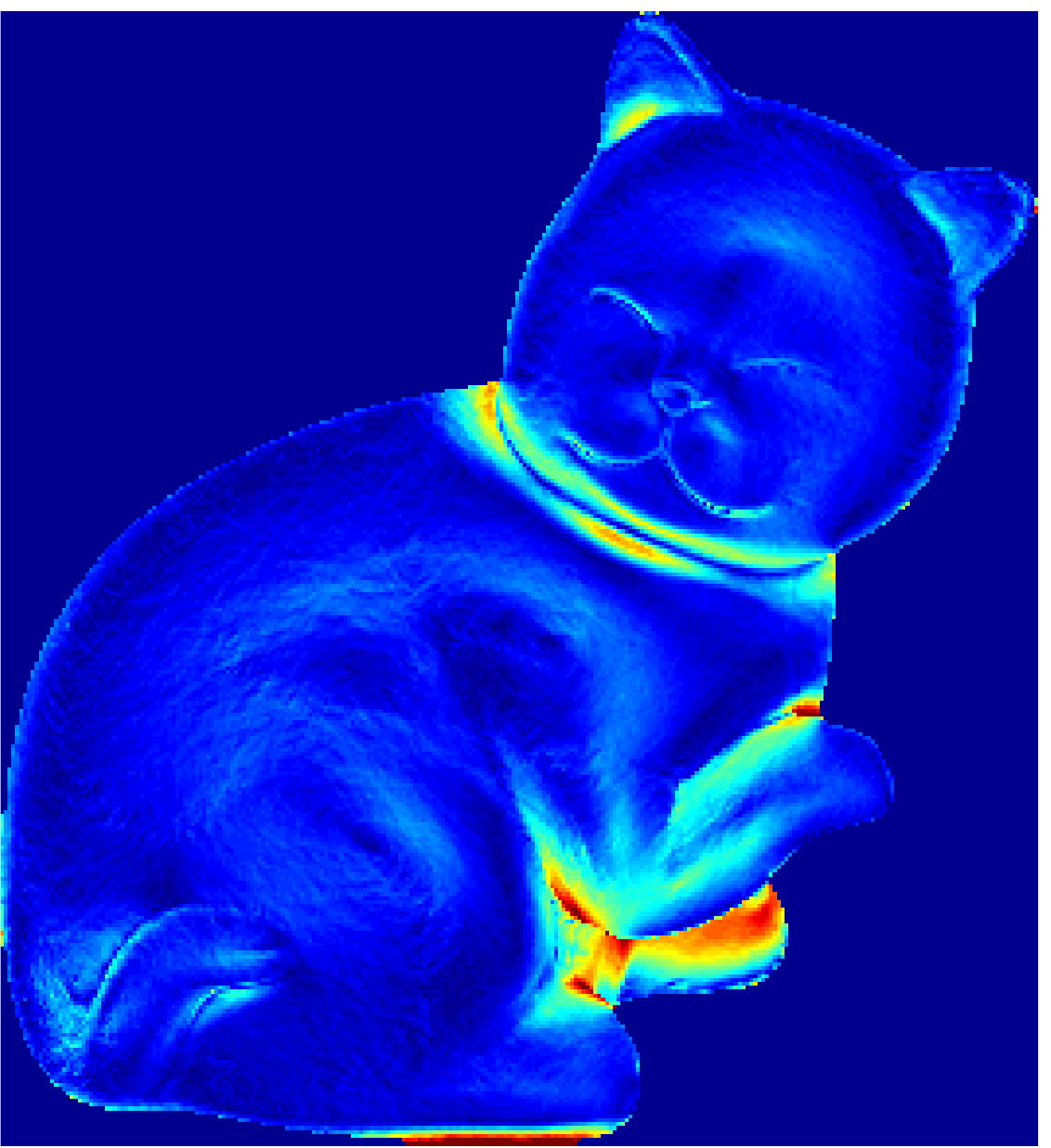} &
			\includegraphics[width = 0.45\linewidth]{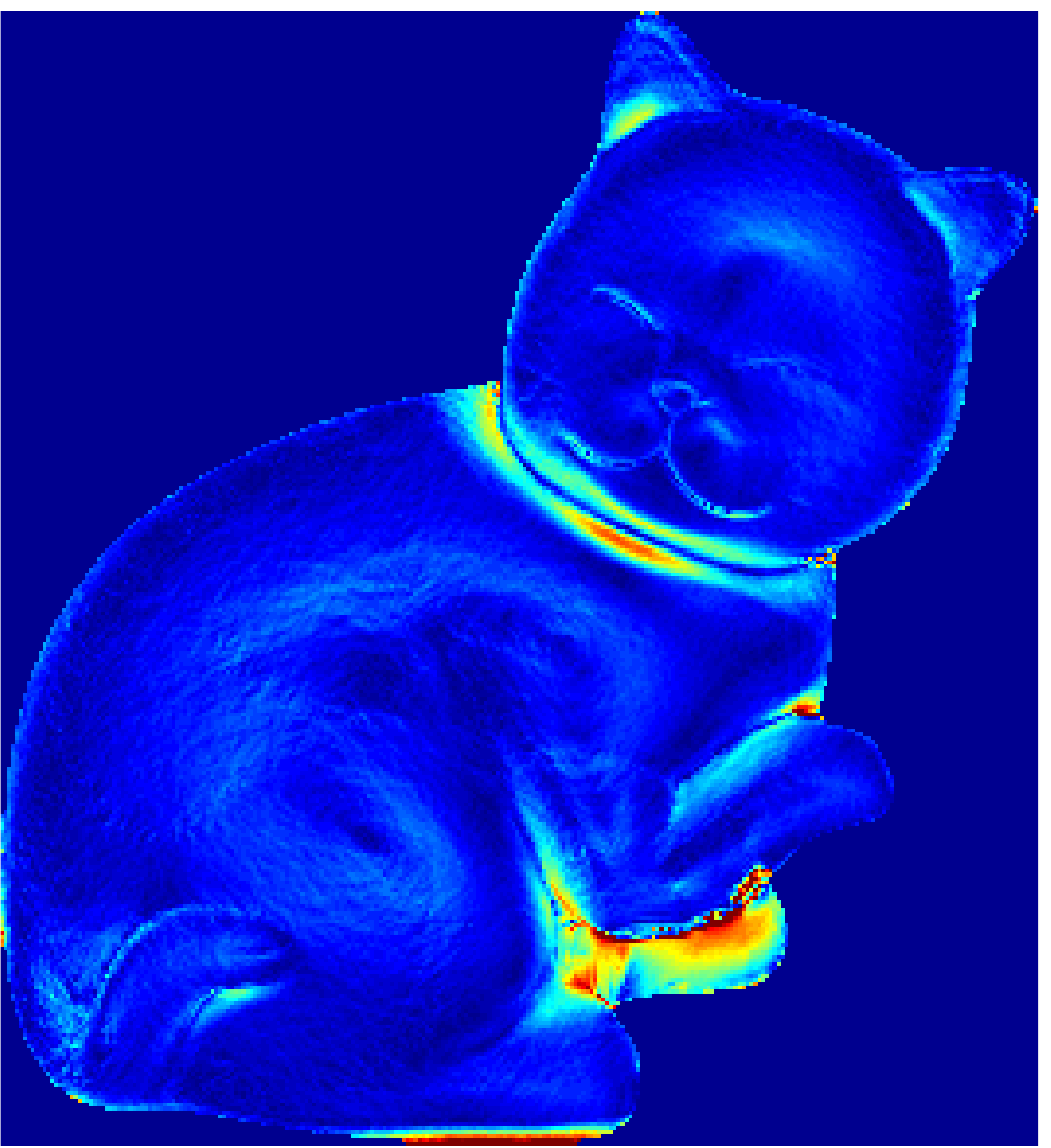} \\
			(MAE $= 9.29$ degrees) & (MAE $ = 8.43$ degrees)
		\end{tabular}
\end{center}
\caption{Top row: three out of the 96 input images used for estimating the normals by photometric stereo~\cite{Woodham:1980a}. Middle row{, left}: 3D-reconstruction by least-squares integration of the normals ({see Section}~\ref{sec:smooth}). {Bottom row, left:} angular error map (blue is $0$ degree, red is $60$ degrees). The estimation is biased around the occluded areas. {Middle and bottom rows, right}: same, using anisotropic diffusion integration with the tensor field defined in~\eqref{eq:anis_diff}. The errors remain confined in the occluded parts, and do not propagate over the discontinuities.}
\label{fig:Cat}
\end{figure}

\begin{figure}[!ht]
\begin{center}
	\includegraphics[width = 0.65\linewidth]{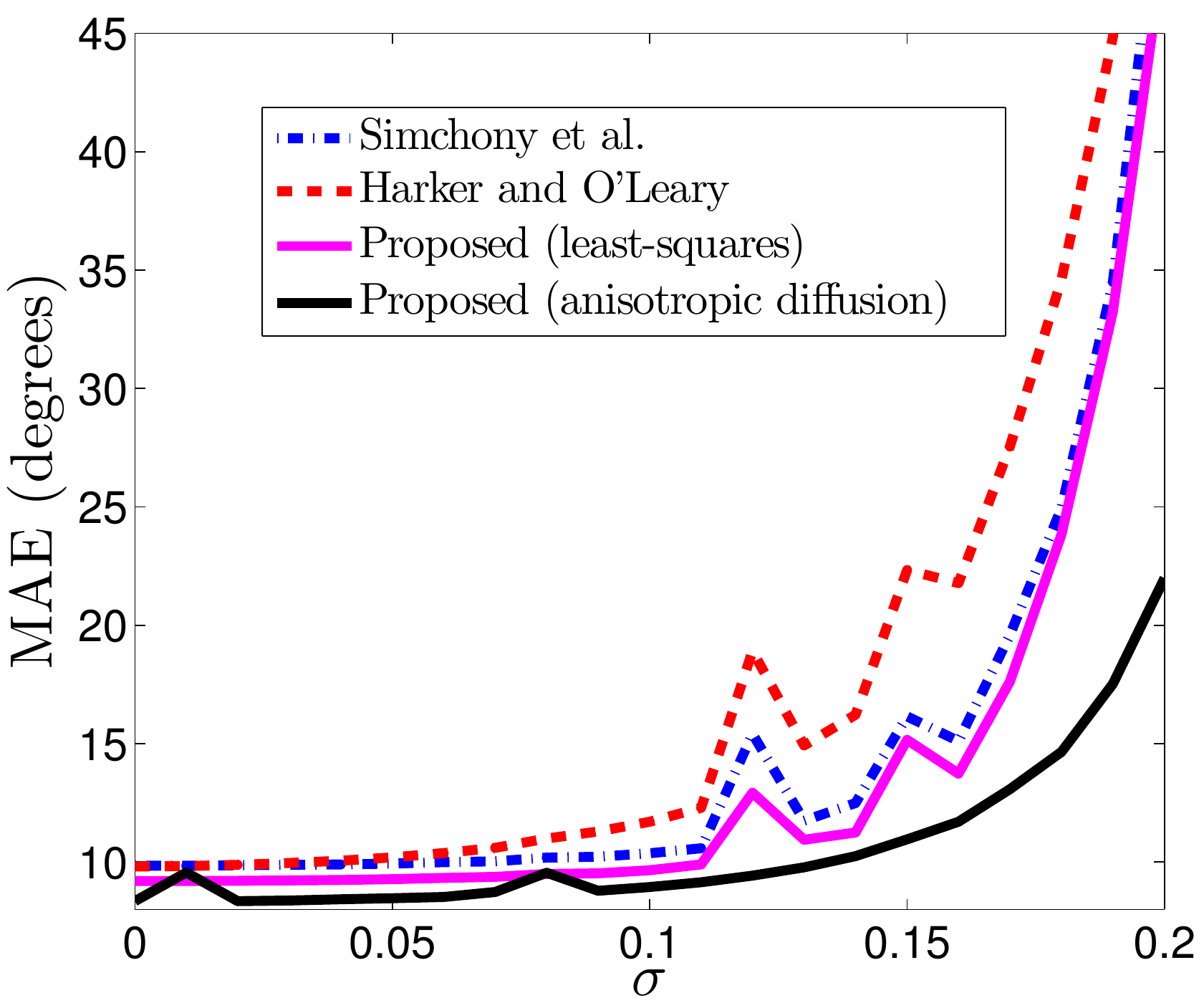}
\end{center}
\caption{Mean angular error (in degrees) as a function of the standard deviation $\sigma \|I\|_\infty$ of the noise which was added to the photometric stereo images. The anisotropic diffusion approach always outperforms least-squares. For the methods~\cite{Harker:2015a,Simchony:1990a}, the gradient field was filled with zeros outside the reconstruction domain, which adds even more bias.}
\label{fig:courbe_cat}
\end{figure}

Figure~\ref{fig:Cat} shows that the 3D-reconstruction obtained by anisotropic diffusion outperforms that obtained by least-square: discontinuities are partially recovered, and robustness to noise is improved (see {Figure}~\ref{fig:courbe_cat}). However, although the diffusion tensor~\eqref{eq:anis_diff} does not require any parameter tuning, the restoration of discontinuities is not as sharp as with the non-convex integrators, and artifacts are visible along the discontinuities.

\begin{figure}[!ht]
\begin{center}
	\begin{tabular}{c}
		\includegraphics[width = 0.95\linewidth,trim={0 0 0 2cm},clip]{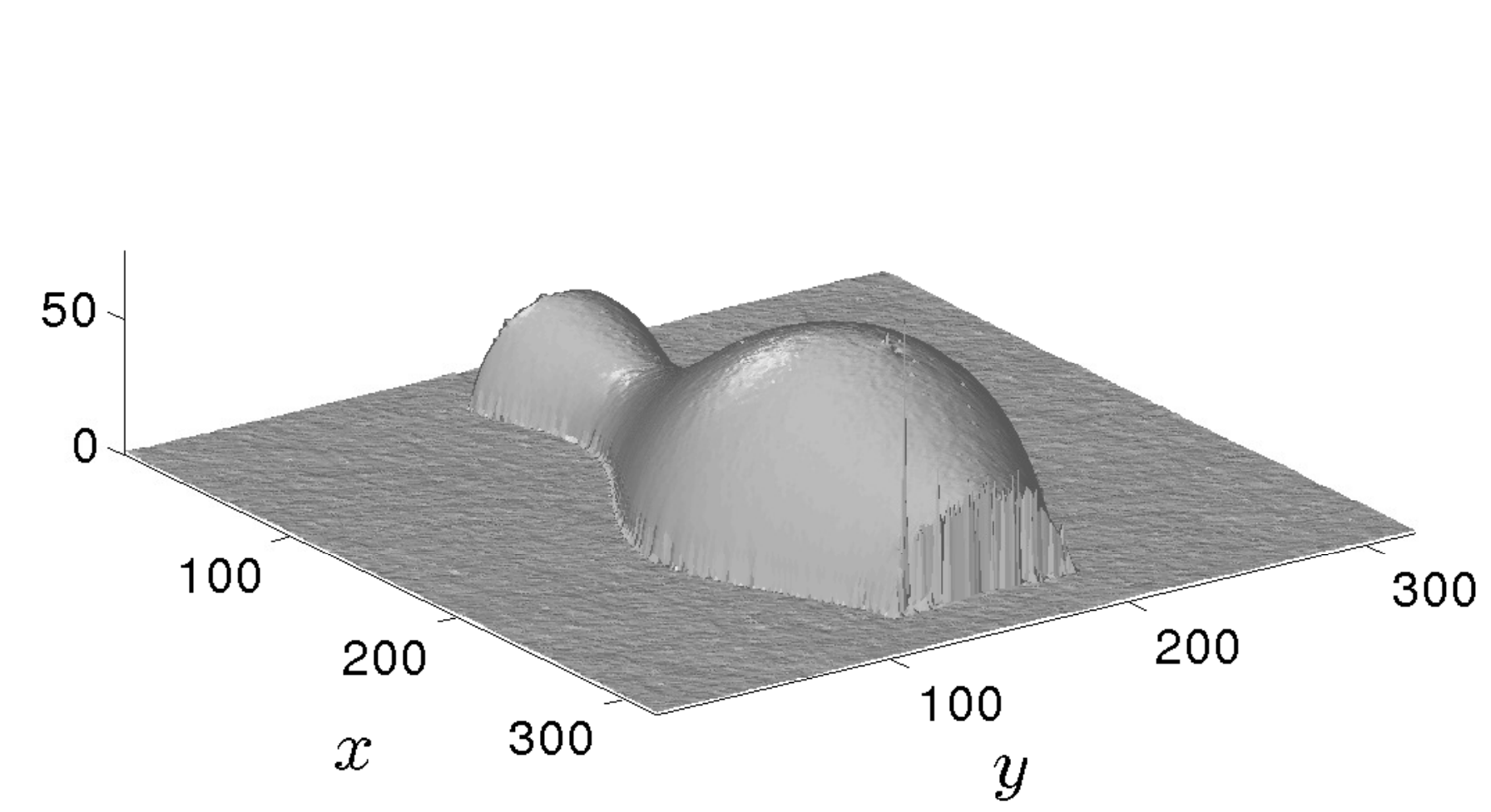} \\
		$\mu = 0.02$ - RMSE $ = 2.38$ \\
		\includegraphics[width = 0.95\linewidth,trim={0 0 0 2cm},clip]{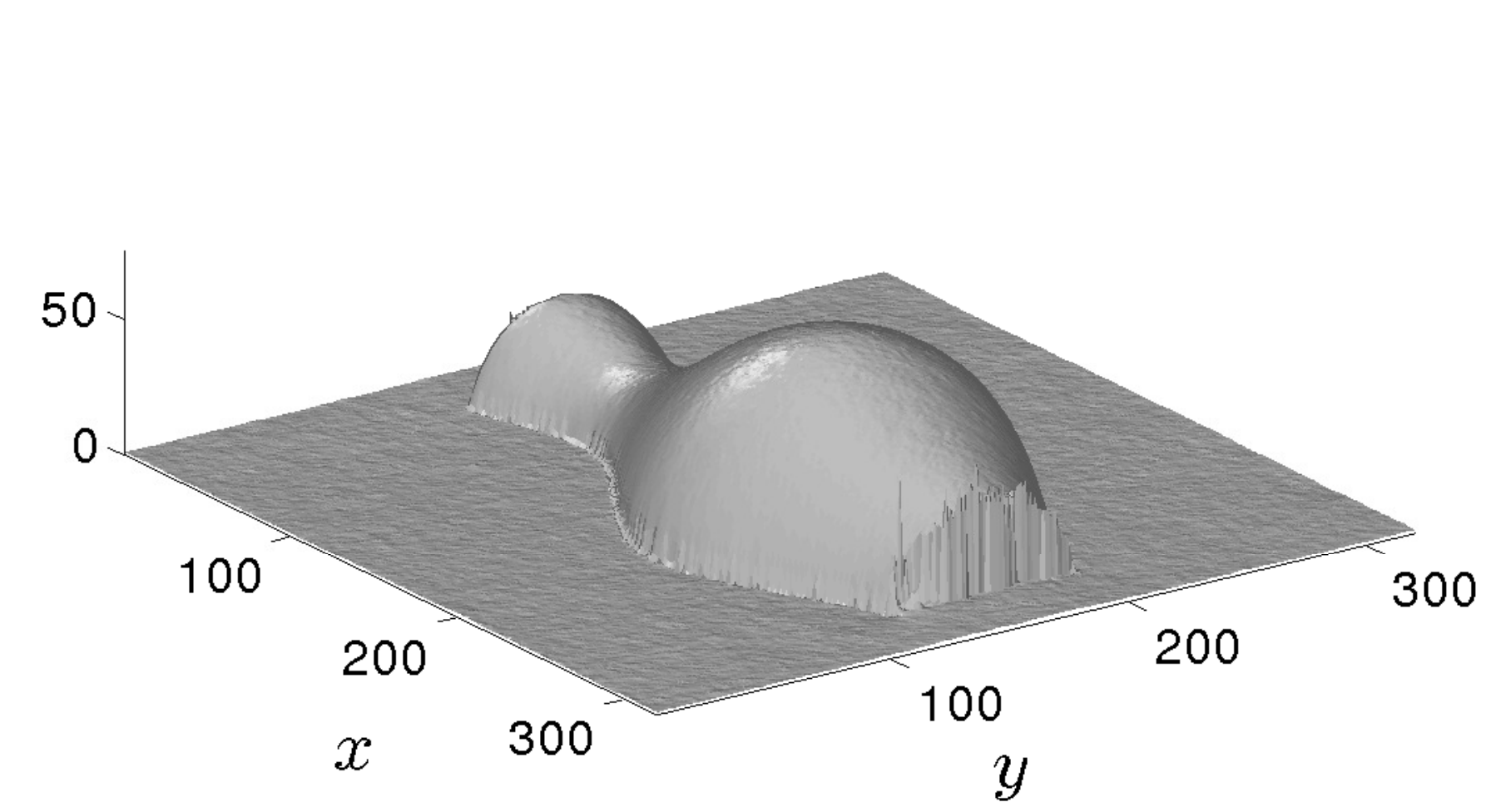} \\
		$\mu = 0.2$ - RMSE $ = 2.19$ \\
		\includegraphics[width = 0.95\linewidth,trim={0 0 0 2cm},clip]{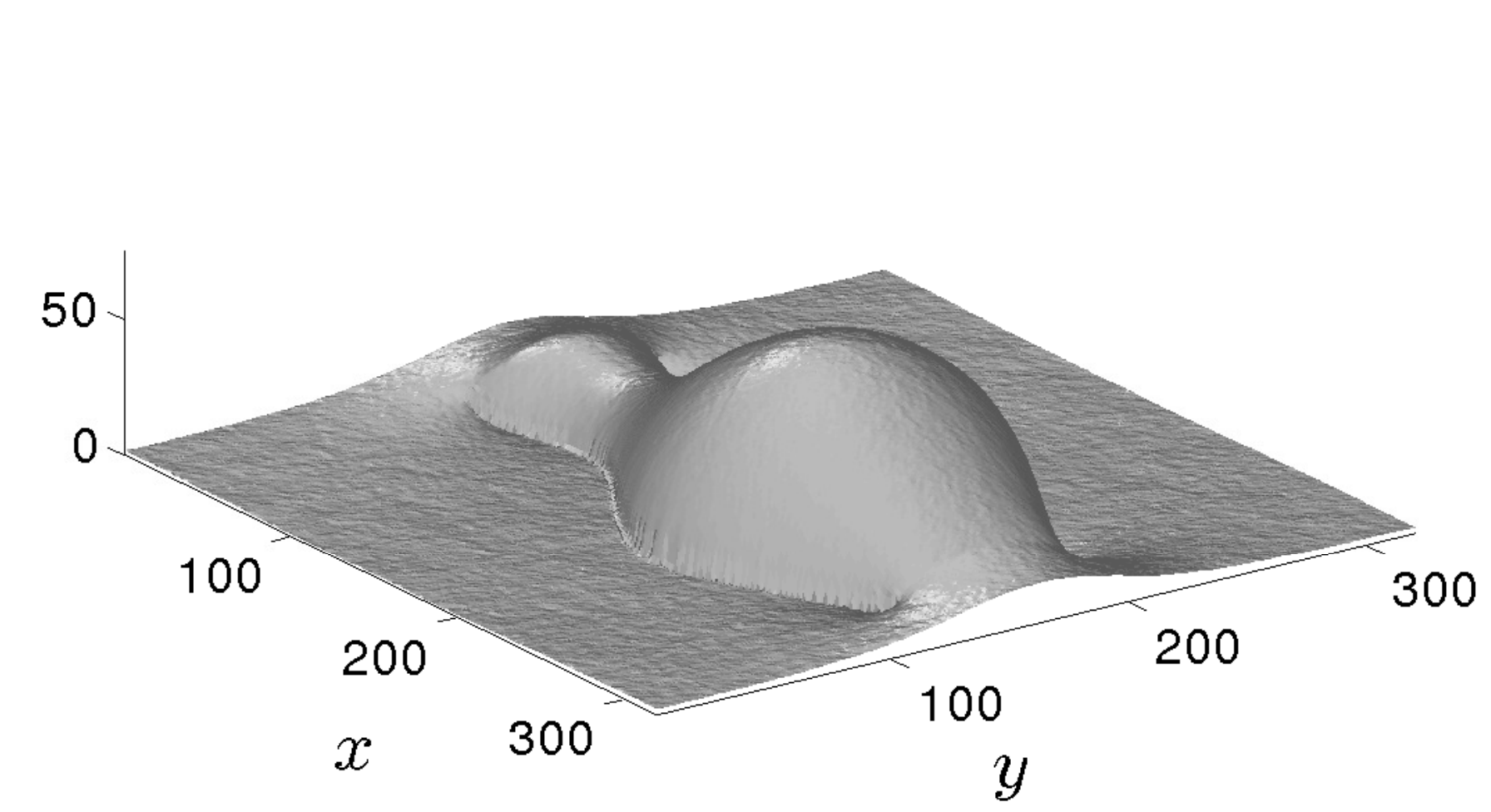} \\
		$\mu = 2$ - RMSE $ = 5.09$
	\end{tabular}
\end{center}
\caption{Integration of the noisy gradient of $\mathcal{S}_\text{vase}$ ($\sigma = 1\%$) by anisotropic diffusion. As long as $\mu$ is small enough, discontinuities are recovered.  Besides, no staircasing artifact is visible. Yet, the restored discontinuities are not perfectly sharp. }
\label{fig:AD}
\end{figure}

Although the parameter-free diffusion tensor~\eqref{eq:anis_diff} seems able to recover discontinuities, this is not always the case. For instance, we did not succeed in recovering the discontinuities of the surface $\mathcal{S}_\text{vase}$. For this dataset, we had to use the tensor~\eqref{eq:anis_diff2}. The results from {Figure}~\ref{fig:AD} show that with an appropriate tuning of $\mu$, discontinuities are recovered and Gibbs phenomena are removed, without staircasing artifact. Yet, as in the experiment of {Figure}~\ref{fig:Cat}, the discontinuities are not very sharp. Such artifacts were also observed by Badri et al.~\cite{Badri2014}, when experimenting with the anisotropic diffusion tensor from Agrawal et al.~\cite{Agrawal:2006a}. Sharper discontinuities could be recovered by using binary weights: this is the spirit of the Mumford-Shah segmentation method, which we explore in the next subsection.

\subsection{Adaptation of the Mumford and Shah Functional}
\label{sec:MS}

Let $z^0:\,\Omega \to \mathbb{R}$ be a noisy image to restore. In order to estimate a denoised image $z$ while perserving the discontinuities of the original image, Mumford and Shah suggested in~\cite{Mumford_Shah} to minimize a quadratic functional only over a subset $\Omega \backslash K$ of $\Omega$, while automatically estimating the discontinuity set $K$ according to some prior. A reasonable prior is that the length of $K$ is ``small", which leads to the following optimization problem:
\begin{align}
	& \underset{z,K}{\min}\,\,\,\, \mu \!\!\! \iint\displaylimits_{(u,v) \in \Omega \backslash K} \| \nabla z(u,v) \|^2 \, \mathrm{d}u\,\mathrm{d}v + \int_K \, \mathrm{d} \sigma \nonumber \\
	& \qquad + \lambda\iint\displaylimits_{(u,v) \in \Omega \backslash K} \left[ z(u,v)-z^0(u,v)\right]^2 \, \mathrm{d}u\,\mathrm{d}v
\label{MS}
\end{align}
where $\lambda$ and $\mu$ are positive constants, and $\int_K \, d \sigma$ is the length of the set $K$. See~\cite{AubertKornprobst} for a detailed introduction to this model and its qualitative properties.

Several approaches have been proposed to numerically minimize the Mumford-Shah functional: finite differences scheme \cite{Chambolle_num}, piecewise constant approximation \cite{Chan_Vese}, primal-dual algorithms~\cite{Pock:2009a}, etc. Another approach consists in using elliptic functionals. An auxiliary function $w:\,\Omega \to \mathbb{R}$ is introduced. This function stands for $1- \chi_K$, where $\chi_K$ is the characteristic function of the set $K$. Ambrosio and Tortorelli have proposed in \cite{Ambrosio_Tortorelli} to consider the following optimization problem:
\begin{align}
	& \underset{z,w}{\min}\,\,\,\, \mu \iint\displaylimits_{(u,v) \in \Omega} w(u,v)^2 \, \| \nabla z(u,v) \|^2 \, \mathrm{d}u\,\mathrm{d}v \nonumber \\
	& \quad\quad+\iint\displaylimits_{(u,v) \in \Omega} \left[ \epsilon \, \| \nabla w(u,v) \|^2 \!+\! \frac{1}{4 \epsilon} \, [w(u,v)\!-\!1]^2 \right] \, \mathrm{d}u\,\mathrm{d}v \nonumber \\
	& \quad\quad+ \lambda \iint\displaylimits_{(u,v) \in \Omega} \left[z(u,v)-z^0(u,v)\right]^2 \, \mathrm{d}u\,\mathrm{d}v
\label{AT}
\end{align}
By using the theory of $\Gamma$-convergence, it is possible to show that \eqref{AT} is a way to solve \eqref{MS} when $\epsilon \to 0$.

We modify the above models, so that they fit our integration problem. Considering $\mathbf{g}$ as basis for least-squares integration everywhere except on the discontinuity set $K$, we obtain the following energy:
\begin{align}
	& \mathcal{E}_{\text{MS}}(z,K) = \mu\! \!\!\!\!\!	\iint\displaylimits_{(u,v) \in \Omega \backslash K} \!\! \| \nabla z(u,v)-\mathbf{g}(u,v) \|^2 \, \mathrm{d}u\,\mathrm{d}v + \int_K \mathrm{d} \sigma \nonumber \\
	& \qquad\qquad+ \!\!\!\!\!\! \iint\displaylimits_{(u,v) \in \Omega\backslash K} \!\!\!\lambda(u,v) \left[ z(u,v)\!-\!z^0(u,v)\right]^2 \mathrm{d}u\,\mathrm{d}v
\label{eq:MSMSMSMS}
\end{align}
for the Mumford-Shah functional, {and the following Ambrosio-Tortorelli approximation:}
\begin{align}
	& \mathcal{E}_{\text{AT}}(z,w) = \mu \iint\displaylimits_{(u,v) \in \Omega} w(u,v)^2 \, \| \nabla z(u,v)-\mathbf{g}(u,v) \|^2 \, \mathrm{d}u\,\mathrm{d}v \nonumber \\
	& \qquad +\iint\displaylimits_{(u,v) \in \Omega}\left[ \epsilon \, \| \nabla w(u,v) \|^2 \!+\! \frac{1}{4 \epsilon} \, [w(u,v)-1]^2 \right] \mathrm{d}u\,\mathrm{d}v \nonumber \\
	& \qquad + \!\!\!\!\iint\displaylimits_{(u,v) \in \Omega}\lambda(u,v) \left[ z(u,v)-z^0(u,v)\right]^2 \, \mathrm{d}u\,\mathrm{d}v
\label{eq:ATcont}
\end{align}
where $w:\,\Omega \to \mathbb{R}$ is a smooth approximation of $1- \chi_K$.

\paragraph{Numerical Solution.} We use the same strategy as in {Section}~\ref{sec:smooth} for discretizing $\nabla z(u,v)$  inside Functional~\eqref{eq:ATcont}, i.e. all the possible first-order discrete approximations of the differential operators are summed. Since discontinuities are usually ``thin" structures, it is possible that a forward discretization contains the  
discontinuity while a backward discretization does not. Hence, the definition of the weights $w$ should be made accordingly to that of $\nabla z$. Thus, we define four fields $w_{u/v}^{+/-}:\,\Omega \to \mathbb{R}$, associated with the finite differences operators $\partial_{u/v}^{+/-}$. This leads to the following discrete {analogue} of Functional~\eqref{eq:ATcont}:
\begin{align}
	& E_{\text{AT}}(\mathbf{z},\mathbf{w}^+_u,\mathbf{w}^-_u,\mathbf{w}^+_v,\mathbf{w}^-_v) = \nonumber \\
	& \frac{\mu}{2} \Bigg( \left\|\mathbf{W}_u^+\left(\mathbf{D}_u^+ \mathbf{z} - \mathbf{p} \right)\right\|^2 + \left\|\mathbf{W}_u^-\left(\mathbf{D}_u^- \mathbf{z} - \mathbf{p} \right)\right\|^2 \nonumber \\
	& \qquad +\left\|\mathbf{W}_v^+\left(\mathbf{D}_v^+ \mathbf{z} - \mathbf{q} \right)\right\|^2 + \left\|\mathbf{W}_v^-\left(\mathbf{D}_v^- \mathbf{z} - \mathbf{q} \right)\right\|^2\Bigg) \nonumber \\
	& + \!\frac{\epsilon}{2} \! \left( \left\|\mathbf{D}_u^+ \mathbf{w}_u^+ \right\|^2 \!\!+\! \left\|\mathbf{D}_u^- \mathbf{w}_u^- \right\|^2 \!\!+\! \left\|\mathbf{D}_v^+ \mathbf{w}_v^+ \right\|^2 \!\!+\! \left\|\mathbf{D}_v^- \mathbf{w}_v^- \right\|^2 \right) \nonumber \\
	& + \!\frac{1}{8\epsilon} \! \left( \left\|\mathbf{w}_u^+ \!-\!\mathbf{1}\right\|^2 \!\!+\! \left\|\mathbf{w}_u^- \!-\!\mathbf{1} \right\|^2 \!\!+\! \left\| \mathbf{w}_v^+ \!-v\mathbf{1}\right\|^2 \!\!+\! \left\|\mathbf{w}_v^- \!-\!\mathbf{1} \right\|^2 \right) \nonumber \\
	& +\left\|{\bm \Lambda} \left( \mathbf{z}-\mathbf{z}^0 \right)\right\|^2
\label{eq:discAT}
\end{align}
where $\mathbf{w}_{u/v}^{+/-} \in \mathbb{R}^{\left|\Omega\right|}$ is a vector containing the values of the discretized field ${w}_{u/v}^{+/-}$, and $\mathbf{W}_{u/v}^{+/-} = \text{Diag}(\mathbf{w}_{u/v}^{+/-})$ is the $|\Omega| \times |\Omega|$ diagonal matrix containing these values. 

\begin{figure*}[!ht]
\begin{center}
	\begin{tabular}{ccc}
		\includegraphics[width = 0.33\linewidth,trim={0 0 0 2.2cm},clip]{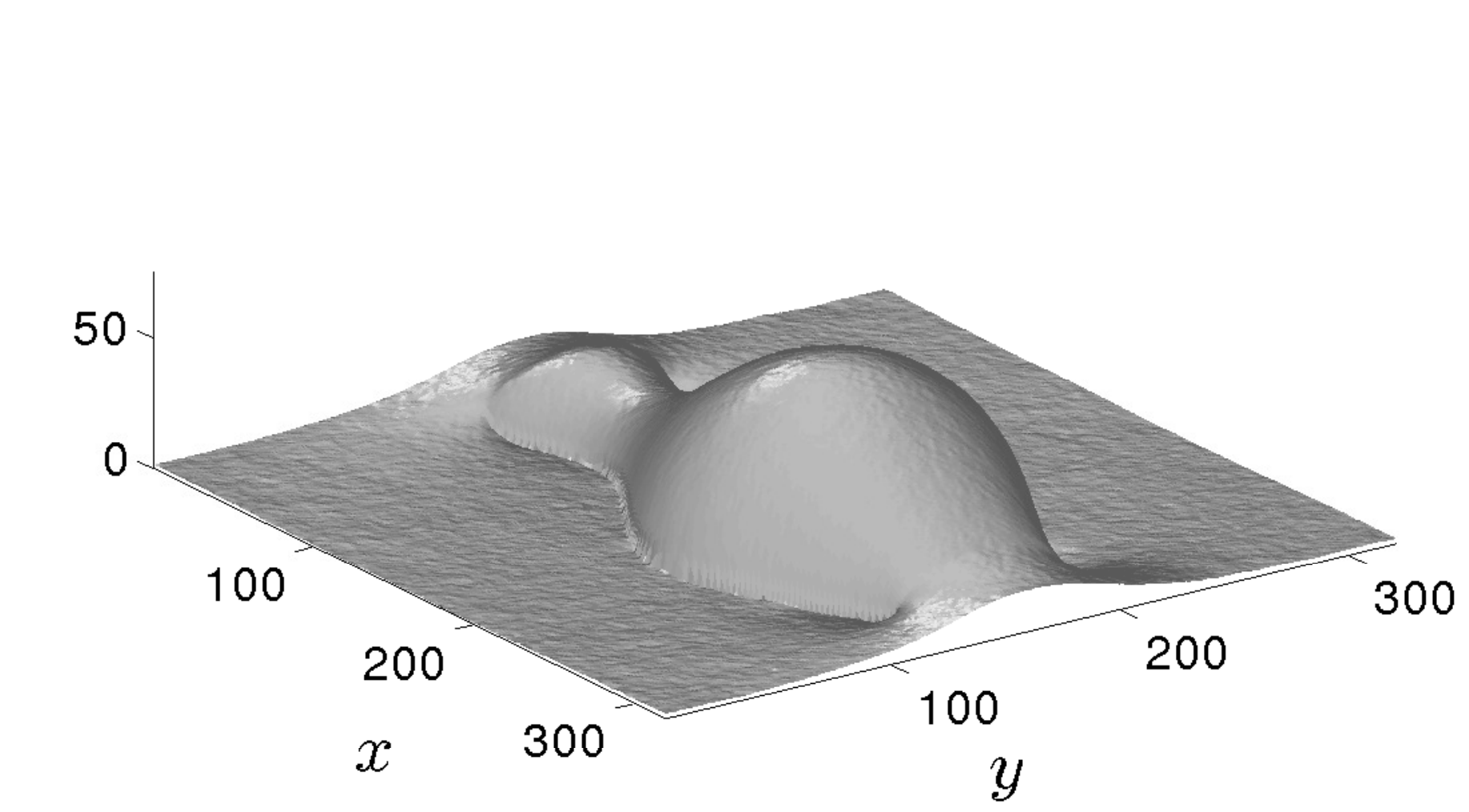} & \!\!\!\!\!\!\!\!\!\!
		\includegraphics[width = 0.33\linewidth,trim={0 0 0 2.2cm},clip]{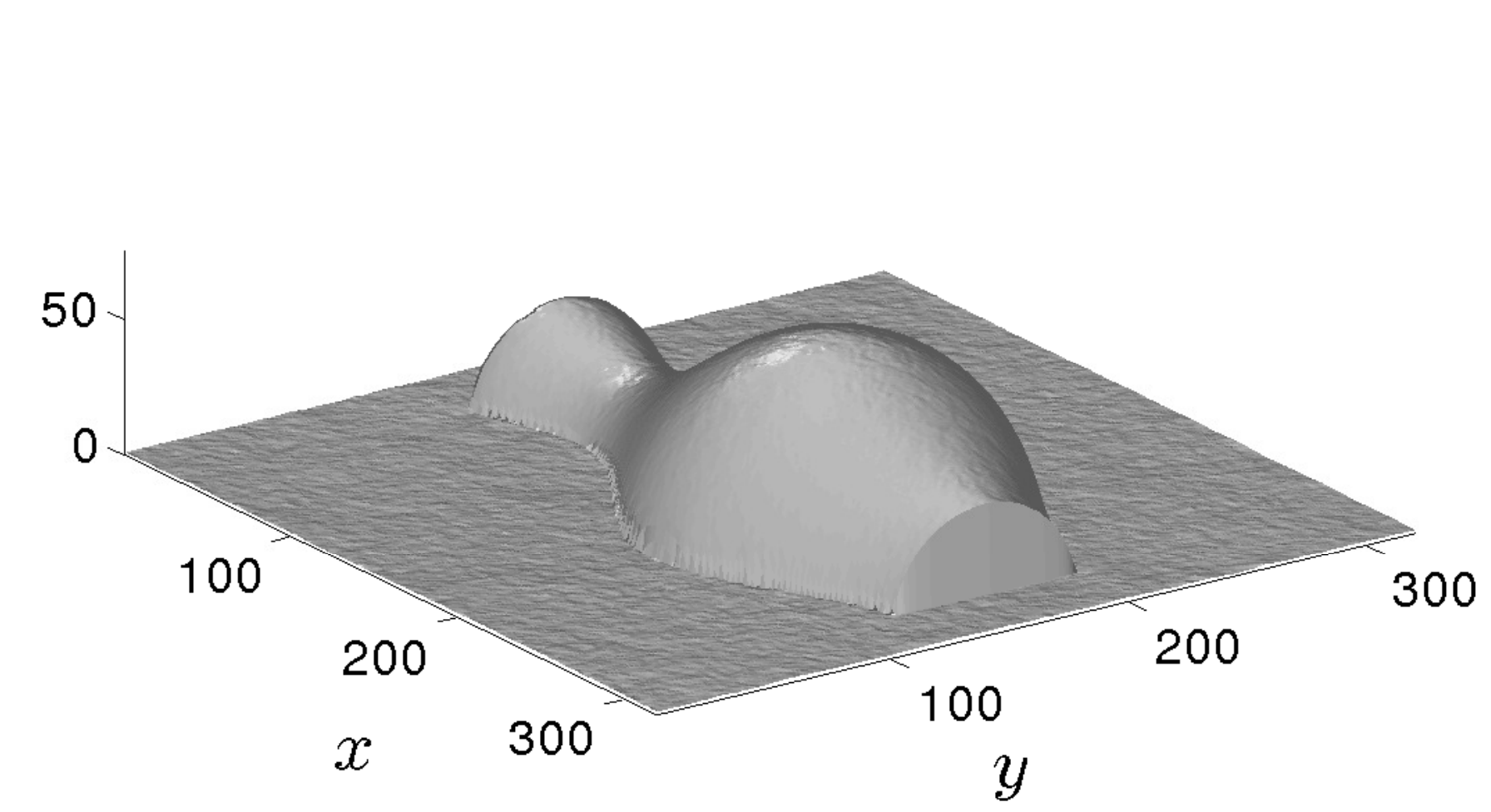} & \!\!\!\!\!\!\!\!\!\!
		\includegraphics[width = 0.33\linewidth,trim={0 0 0 2.2cm},clip]{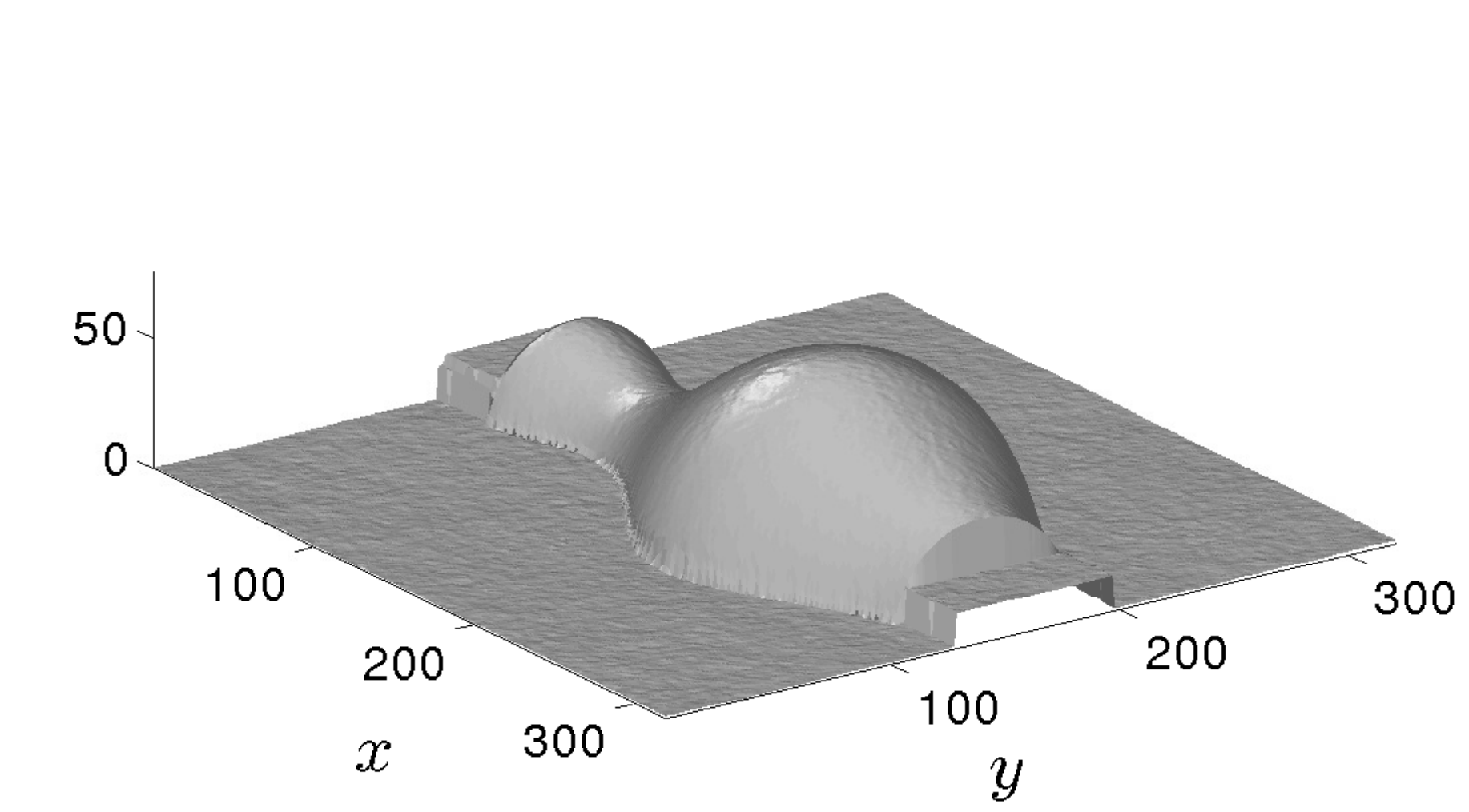} \\				
		$\mu = 1$ - RMSE $=4.94$ &
		$\mu = 45$ - RMSE $=2.37$ &
		$\mu = 100$ - RMSE $=4.14$
	\end{tabular}
\end{center}
\caption{3D-reconstructions from the noisy gradient of $\mathcal{S}_\text{vase}$ ($\sigma = 1\%$), using the Mumford-Shah integrator. If $\mu$ is tuned appropriately, sharp discontinuities can be restored, without staircasing artifacts.}
\label{fig:MS_par}
\end{figure*}

We tackle the nonlinear problem~\eqref{eq:discAT} by an alternating optimization scheme:
\begin{align}
	& \mathbf{z}^{(k+1)} \!= \underset{\mathbf{z} \in \mathbb{R}^{|\Omega|}}{\operatorname{argmin}}~ {E}_{\text{AT}}(\mathbf{z},{\mathbf{w}_u^+}^{(k)}\!\!,\!{\mathbf{w}_u^-}^{(k)}\!\!,\!{\mathbf{w}_v^+}^{(k)}\!\!,\!{\mathbf{w}_v^-}^{(k)}) \label{eq:z_update} \\
	& {\mathbf{w}_u^+}^{(k+\!1)} \!\!\!\!=\! \underset{\mathbf{w} \in \mathbb{R}^{|\Omega|}}{\operatorname{argmin}}\, {E}_{\text{AT}}(\mathbf{z}^{(k\!+1)},\mathbf{w},\!{\mathbf{w}_u^-}^{(k)}\!\!,\!{\mathbf{w}_v^+}^{(k)}\!\!,\!{\mathbf{w}_v^-}^{(k)}\!) \label{eq:wup_update}
\end{align}
and similar straightforward updates for the other indicator functions. We can choose as initial guess, for instance, the smooth solution from {Section}~\ref{sec:smooth} for $\mathbf{z}^{(0)}$, and ${\mathbf{w}_u^+}^{(0)}={\mathbf{w}_u^-}^{(0)}={\mathbf{w}_v^+}^{(0)}={\mathbf{w}_v^-}^{(0)}\equiv \mathbf{1} $.

At each iteration $(k)$, updating the surface and the indicator functions requires solving a series of linear least-squares problems. We achieve this by solving the resulting linear systems (normal equations) by means of the conjugate gradient algorithm. Contrarily to the approaches that we presented so far, the matrices involved in these systems are modified at each iteration. Hence, it is not possible to compute the preconditioner beforehand. 
In our experiments, we did not consider any preconditioning strategy at all. Thus, the proposed scheme could obviously be accelerated. 

\paragraph{Discussion.} Let us now check experimentally, on the same noisy gradient of surface $\mathcal{S}_\text{vase}$ as in previous experiments, whether the Mumford-Shah integrator satisfies the expected properties. In the experiment of {Figure}~\ref{fig:MS_par}, we performed 50 iterations of the proposed alternating optimization scheme, with various choices for the hyper-parameter $\mu$. The $\epsilon$ parameter was set to $\epsilon = 0.1$ (this parameter is not critical: it only has to be ``small enough", in order for the Ambrosio-Tortorelli approximation to converge towards the Mumford-Shah functional). As it was already the case with other non-convex regularizers (see {Subsection}~\ref{sec:nonconvex}), a bad tuning of the parameter leads either to over-smoothing ({low} values of $\mu$) or to staircasing artifacts ({high} values of $\mu$), which indicate the presence of local minima. Yet, by appropriately setting this parameter, we obtain a 3D-reconstruction which is very close to the genuine surface, without staircasing artifact.

\begin{figure}[!ht]
\begin{center}
	\begin{tabular}{c}
		\includegraphics[width = 0.87\linewidth,trim={0 0 0 1.2cm},clip]{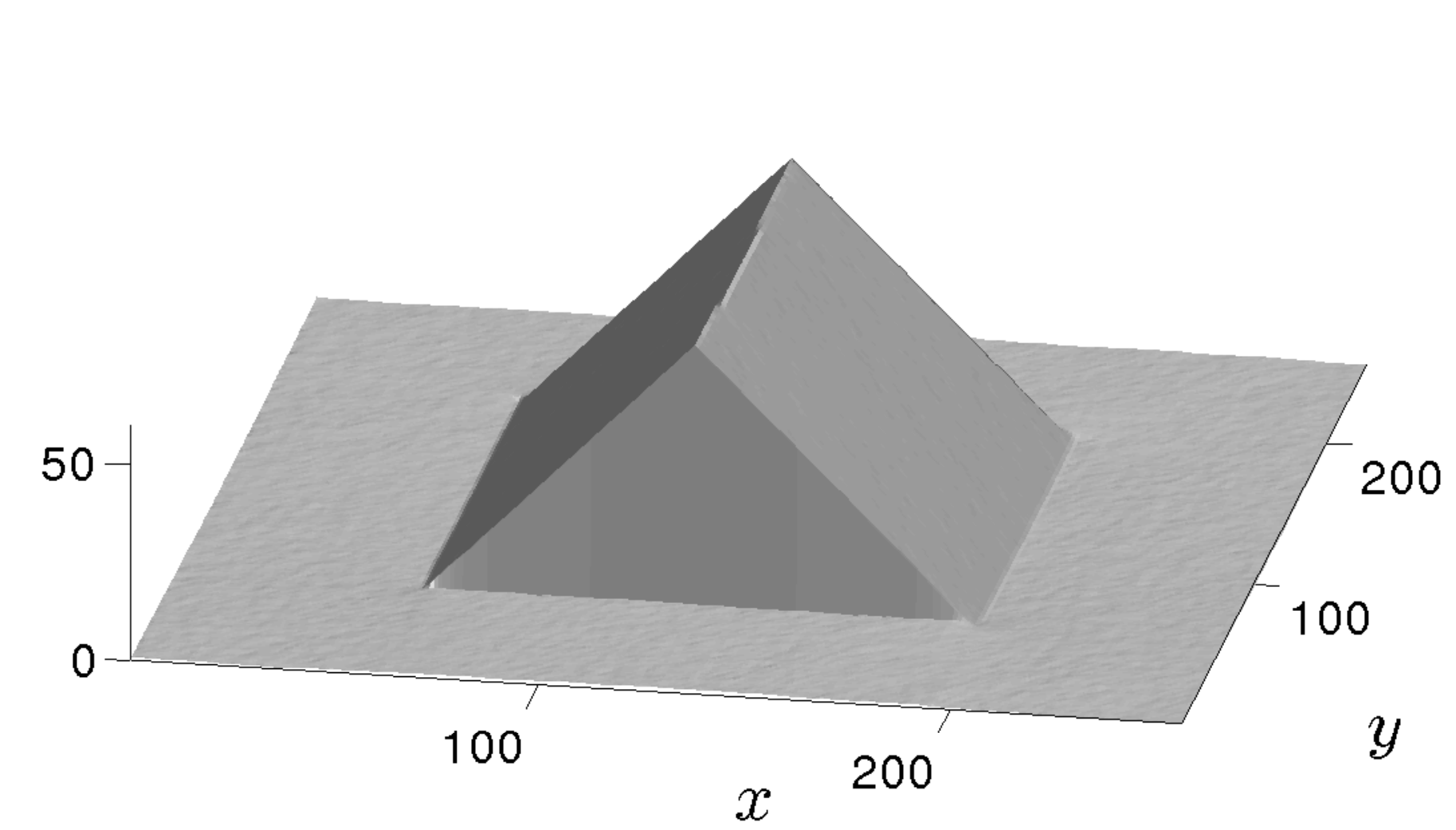} \\
		$z^{(0)} = $ least-squares solution - RMSE $= 0.74$ \\
		\includegraphics[width = 0.87\linewidth,trim={0 0 0 1.2cm},clip]{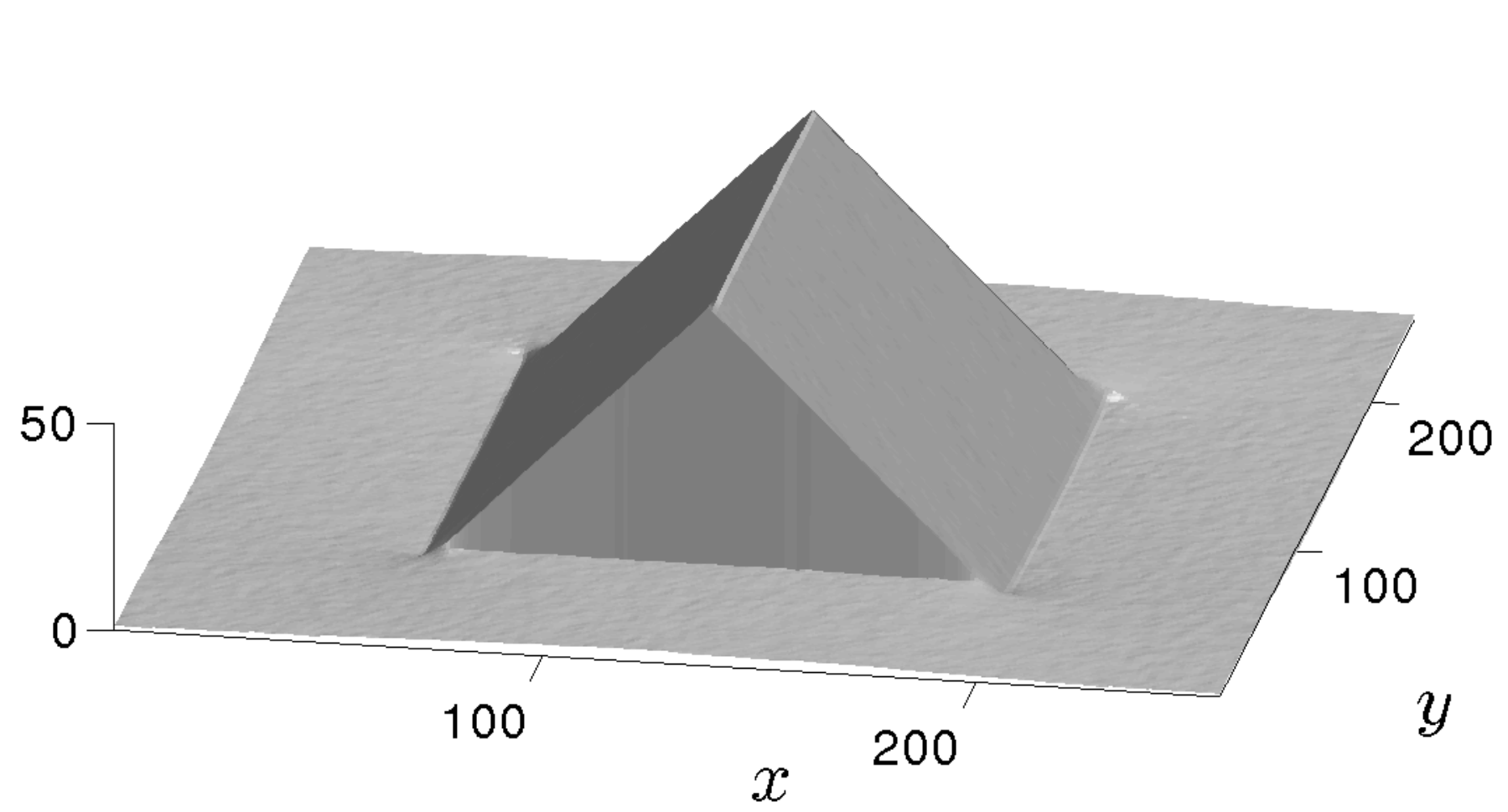} \\
		$z^{(0)} \equiv 0$ - RMSE $= 1.84$
	\end{tabular}
\end{center}
\caption{3D-reconstructions of the ``Canadian tent" surface from its noisy gradient ($\sigma = 1\%$), by the Mumford-Shah integrator ($\mu = 20$), using two different initializations. The initialization matters, but not as much as with the non-convex estimators from {Subsection}~\ref{sec:nonconvex}.}
\label{fig:MS_init}
\end{figure}

The Mumford-Shah functional being non-convex, local minima may {exist}. Yet, as shown in {Figure}~\ref{fig:MS_init}, the choice of the initialization may not be as crucial as with the non-convex estimators from {Subsection}~\ref{sec:nonconvex}. Indeed, the 3D-reconstruction of the ``Canadian tent" surface is similar using as initial guess the least-squares solution or the trivial initialization $z^{(0)} \equiv 0$.

Hence, among {all the variational integration} methods we have studied, the adaptation of the Mumford-Shah model is the approach which provides the most satisfactory 3D-reconstructions {in the presence of sharp features}: it is possible to recover {discontinuities and kinks}, even in the presence of noise, and with limited artifacts. {Nevertheless, local minima may theoretically arise, as well as staircasing if the parameter $\mu$ is not tuned appropriately.}
 
%

\begin{table*}[htbp]
\caption[]{Main features of the five methods of integration proposed in this paper. The quadratic method has all desirable properties, except $\mathcal{P}_{\text{Disc}}$. The others lose $\mathcal{P}_{\text{Fast}}$ but hold $\mathcal{P}_{\text{Disc}}$. {Sharpest features are recovered by using non-convex regularization or the Mumford-Shah approach, yet staircasing artifacts and local minima may appear}. In addition, all discontinuity-preserving methods except TV require tuning at least one hyper-parameter. {Yet, TV is not able to recover discontinuities in the presence of noise.} Overall, {we recommend using: quadratic integration if speed is the most important issue; the Mumford-Shah approach if recovering discontinuities is the most important issue; and anisotropic diffusion if discontinuities are present, but limited.}}
\label{tab:recap}
\begin{center}
{
	\begin{tabular}{|c|c|c|c|c|c|c|c|c|}
		\hline
		Method & $\mathcal{P}_{\text{Fast}}$ & $\mathcal{P}_{\text{Robust}}$ & $\mathcal{P}_{\text{FreeB}}$ & $\mathcal{P}_{\text{Disc}}$ & $\mathcal{P}_{\text{NoRect}}$ & $\mathcal{P}_{\text{NoPar}}$ & Local minima & Staircasing \\
		\hline
		Quadratic & $+++$ & $+$ & $+$ & $-$ & $+$ & $+$ & No & No \\
		Total variation & $+$ & $+$ & $+$ & $+$ & $+$ & $+$ & No & Yes \\
		Non-convex regularization & $-$ & $+$ & $+$ & $+++$ & $+$ & $-$ & Yes & Yes \\
		Anisotropic diffusion & $-$ & $+$ & $+$ & $++$ & $+$ & $--$ & No & No \\
		Mumford-Shah & $-$ & $+$ & $+$ & $+++$ & $+$ & $--$ & Yes & Yes \\
		\hline
	\end{tabular}
}
\end{center}
\end{table*}

\section{Conclusion and Perspectives}
\label{sec:applications}

{We proposed several new variational methods for solving the normal integration problem. These methods were designed to satisfy the largest subset of properties that were identified in a companion survey paper~\cite{Durou:2016a} entitled \emph{Normal Integration: A Survey}.}
 


We first detailed in {Section}~\ref{sec:smooth} a least-squares solution which is fast, robust and parameter-free, while assuming neither a particular shape for the integration domain nor a particular boundary condition. However, discontinuities in the surface can be handled only if the integration domain is first segmented into pieces without discontinuities. {Therefore}, we discussed in {Section}~\ref{sec:nonsmooth} several non-quadratic or non-convex variational formulations {aiming at appropriately handling discontinuities}. {As we have seen, the latter property can be satisfied only if (slow) iterative schemes are used and $/$ or one critical parameter is tuned. Therefore, there is still room for improvement: a fast, parameter-free integrator, able to handle discontinuities remains to be proposed. 

{Table~\ref{tab:recap} summarizes the main features of the five new integration methods proposed in this 
 {article}. Contrarily to Table~1 in~\cite{Durou:2016a}, which recaps the features of state-of-the-art methods, this time we use a more nuanced evaluation than binary features $+/-$.}
Among the new methods, we believe that the {least-squares method discussed in {Section}~\ref{sec:smooth} is the best if speed is the most important criterion, while the Mumford-Shah approach discussed in {Subsection}~\ref{sec:MS} is the most appropriate one for recovering discontinuities and kinks. Inbetween, the anisotropic diffusion approach from Subsection~\ref{sec:PM} represents a good compromise}. 

Future research directions may include accelerating the numerical schemes and proving their convergence when this is not trivial (e.g., for the non-convex integrators). We also believe that introducing additional smoothness terms inside the functionals may be useful for eliminating the artifacts in anisotropic diffusion integration. Quadratic (Tikhonov) smoothness terms were suggested in~\cite{Harker:2015a}: to enforce surface smoothness while preserving the discontinuities, we should rather consider non-quadratic ones. In this view, higher-order functionals (e.g., total generalized variation methods~\cite{Bredies2015}) may reduce not only these artifacts, but also staircasing. Indeed, as shown in {Figure}~\ref{fig:Beethoven}, such artifacts may be visible when performing photometric stereo~\cite{Woodham:1980a} without prior segmentation. Yet, this example also shows that the artifacts are visible only over the background, and do not seem to affect the relevant part.

\begin{figure*}[!ht]
\begin{center}
	\begin{tabular}{cccc}
		\includegraphics[height = 0.18\linewidth]{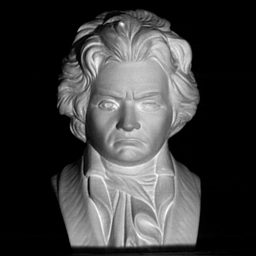} &
		\includegraphics[height = 0.18\linewidth]{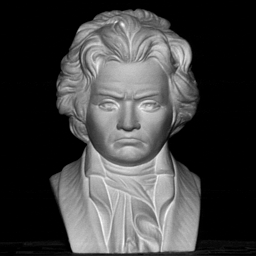} &
		\includegraphics[height = 0.18\linewidth]{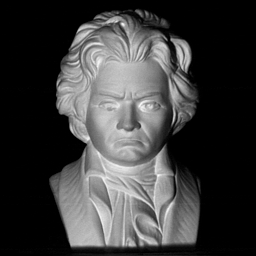} &
		\includegraphics[height = 0.18\linewidth,trim={0 0 0 1.8cm},clip]{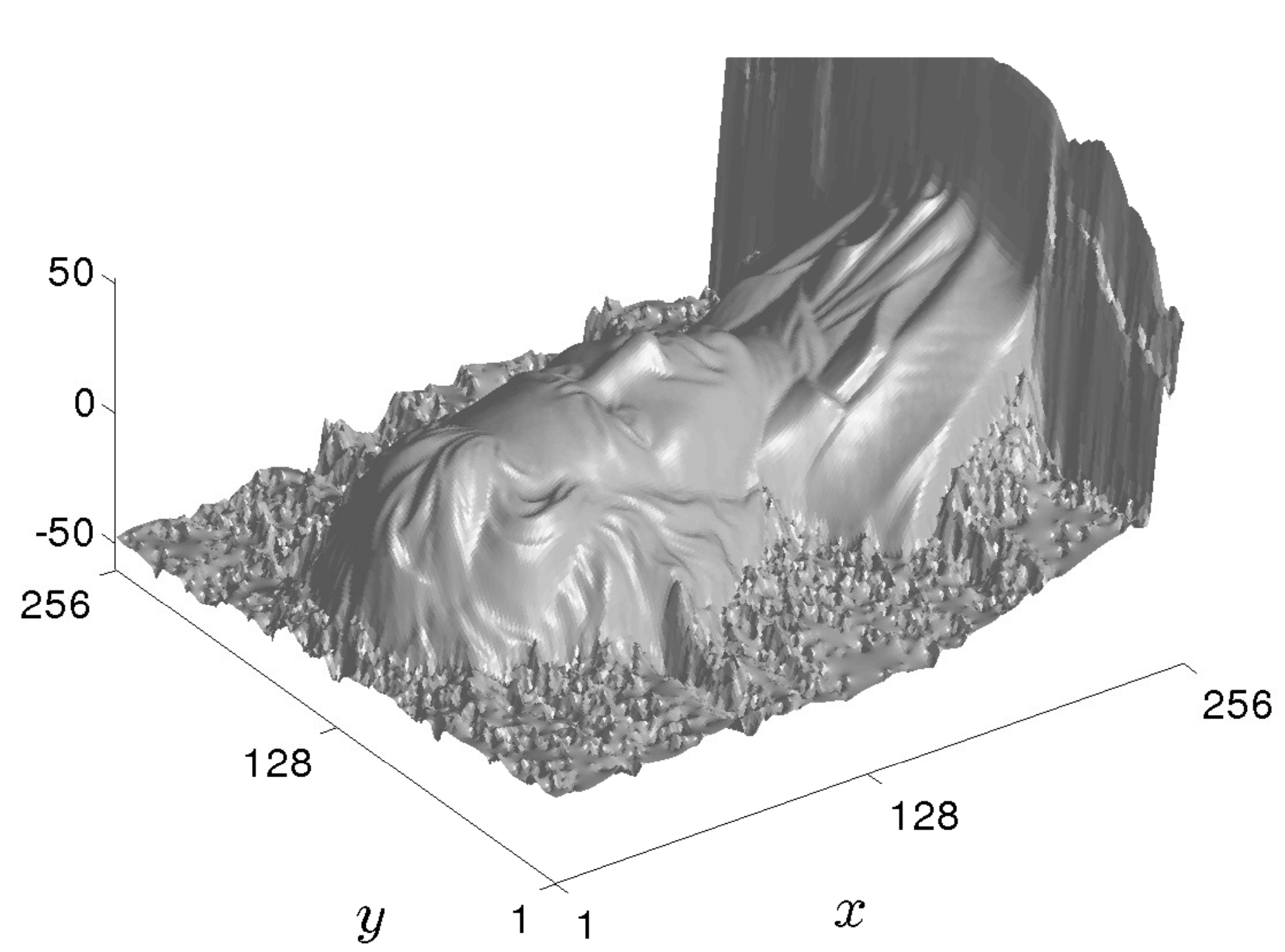} \\
		(a) & (b) & (c) & (d)
	\end{tabular}
	\begin{tabular}{cc}
		\includegraphics[width = 0.48\linewidth,trim={0 0 0 2.5cm},clip]{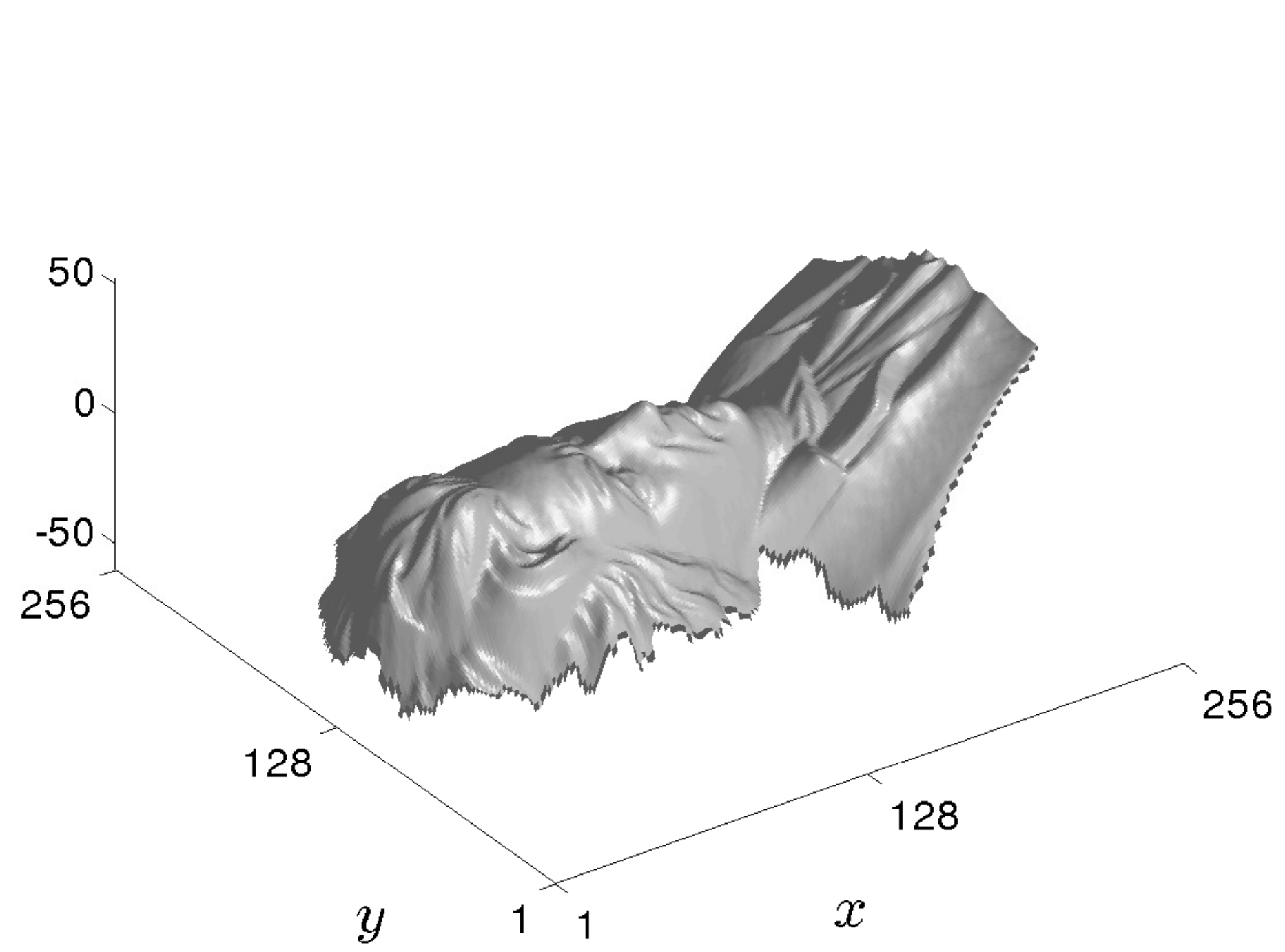} &
		\includegraphics[width = 0.48\linewidth,trim={0 0 0 2.5cm},clip]{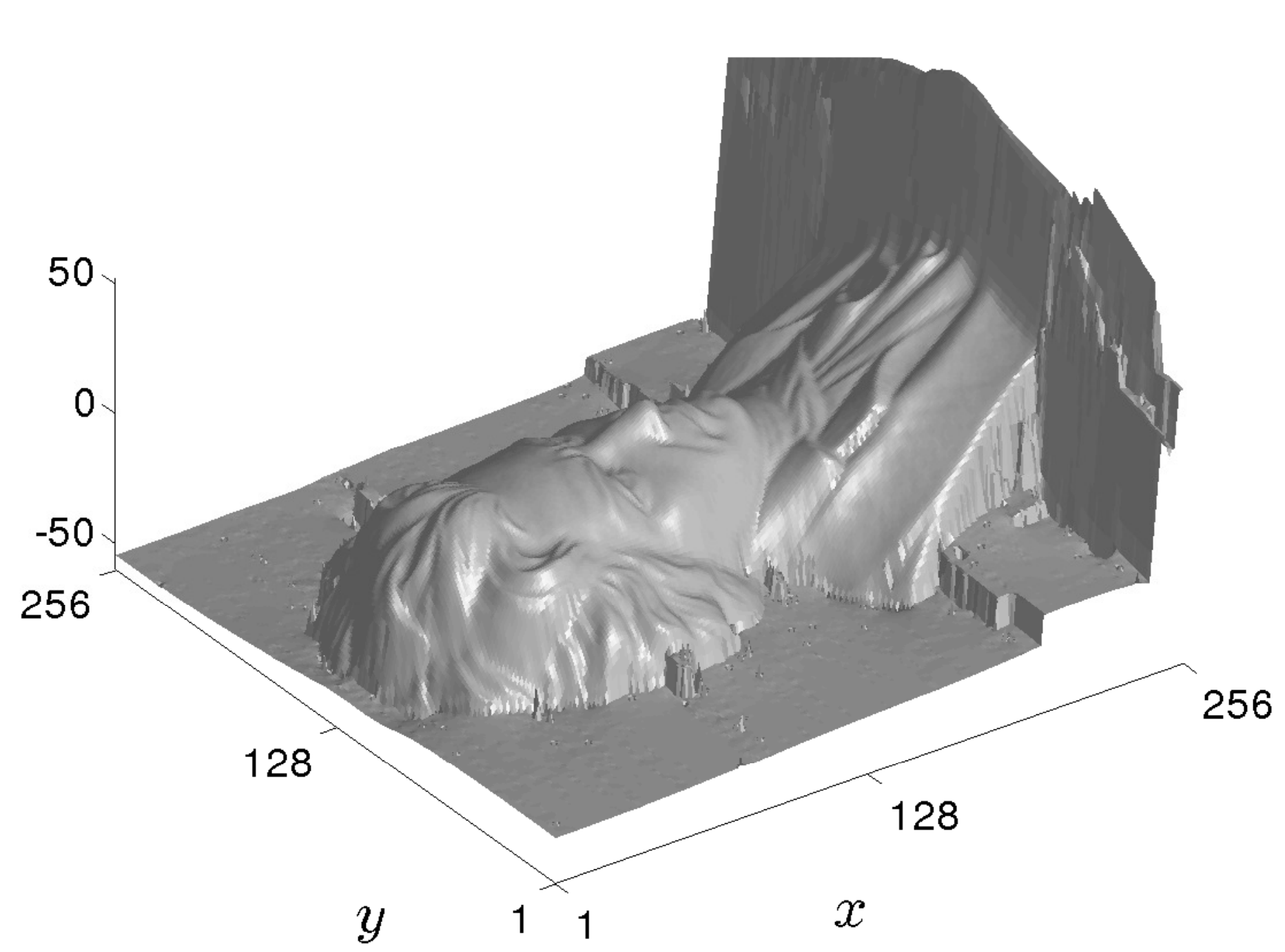} \\
		(e) & (f)
	\end{tabular}
\end{center}
\caption{3D-reconstruction using photometric stereo. (a-c) All (real) input images. (d) 3D-reconstruction by least-squares on the whole grid. (e) 3D-reconstruction by least-squares on the non-rectangular reconstruction domain corresponding to the images of the bust. (f) 3D-reconstruction using the Mumford-Shah approach, on the whole grid. When discontinuities are handled, it is possible to perform photometric stereo without prior segmentation of the object.} 
\label{fig:Beethoven}
\end{figure*}

3D-reconstruction is not the only application where efficient tools for gradient field integration are required. Although the assumption on the noise distribution may differ from one application to another, PDE-based imaging problems such as Laplace image compression~\cite{Peter:2016a} or Poisson image editing~\cite{Perez:2003} also require an efficient integrator. In this view, the ability of our methods to handle control points may be useful. We illustrate in {Figure}~\ref{fig:Peppers} an interesting application. From an RGB image $I$, we selected the points where the norm of the gradient of the luminance {(in the CIE-LAB color space)} was the highest (conserving only $10\%$ of the points). Then, we created a gradient field $\mathbf{g}$ equal to zero everywhere, except on the control points, where it was set to the gradient of the color levels. The prior $z^0$ was set to a null scalar field, except on the control points where we retained the original color data. Eventually, $\lambda$ is set to an arbitrary small value ($\lambda = 10^{-9}$) everywhere, except on the control points ($\lambda = 10$). The integration of each color channel gradient is performed independently, using the Mumford-Shah method to extrapolate the data from the control points to the whole grid. Using this approach, we obtain a nice piecewise-constant approximation of the image, in the spirit of the ``texture-flattening" application presented in~\cite{Perez:2003}. Besides, by selecting the control points in a more optimal way~\cite{Belhachmi:2009a,Hoeltgen:2013a}, this approach could easily be extended to image compression, reaching state-of-the-art lossy compression rates. In fact, existing PDE-based methods can already compete with the compression rate of the well-known JPEG 2000 algorithm~\cite{Peter:2016a}. We believe that the proposed edge-preserving framework may yield even better results.

\begin{figure*}[!ht]
\begin{center}
	\begin{tabular}{ccc}
		\includegraphics[width = 0.3\linewidth]{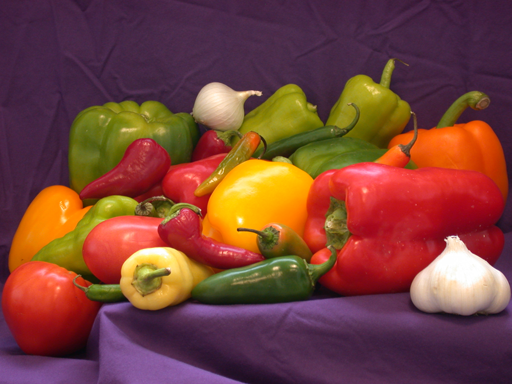} &
		\includegraphics[width = 0.3\linewidth]{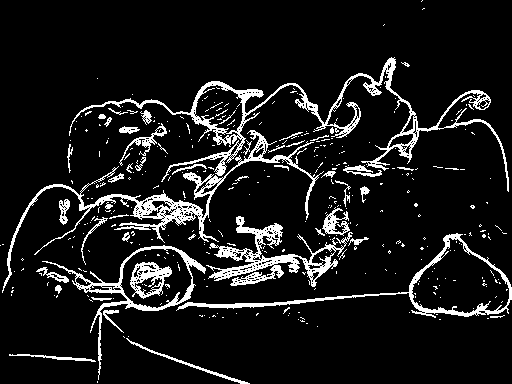} &
		\includegraphics[width = 0.3\linewidth]{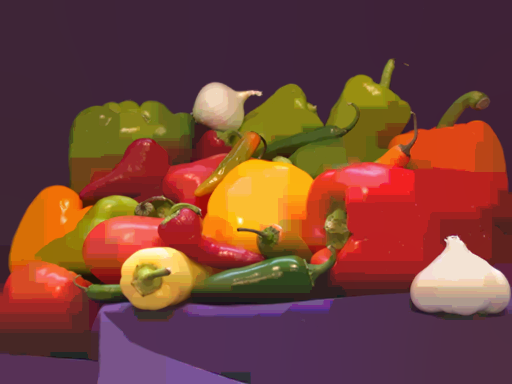} \\
		(a) & (b) & (c)
	\end{tabular}
\end{center}
\caption{Application to image compression$/$image editing. (a) Reference image. (b) Control points (where the RGB-values and their gradients are kept). (c) Restored image obtained by considering the proposed Mumford-Shah integrator as a piecewise-constant interpolation method. A reasonable piecewise constant restoration of the initial image can be obtained from as few as $10\%$ of the initial information.
}
\label{fig:Peppers}
\end{figure*}

Eventually, some of the research directions already mentioned in the conclusion section of our  {survey paper~\cite{Durou:2016a}}
were ignored in this second paper, but they remain of important interest. One of the most appealing examples is multi-view normal field integration~\cite{Chang:2007a}. Indeed, discontinuities represent a difficulty in our case because they are induced by occlusions, yet more information would be obtained near the occluding contours by using additional views.

\bibliographystyle{spmpsci}
\bibliography{biblio2}

\end{document}

%% file: NewFigures/TroisTrois/troistrois.pdf_tex
\begingroup%
  \makeatletter%
  \providecommand\color[2][]{%
    \errmessage{(Inkscape) Color is used for the text in Inkscape, but the package 'color.sty' is not loaded}%
    \renewcommand\color[2][]{}%
  }%
  \providecommand\transparent[1]{%
    \errmessage{(Inkscape) Transparency is used (non-zero) for the text in Inkscape, but the package 'transparent.sty' is not loaded}%
    \renewcommand\transparent[1]{}%
  }%
  \providecommand\rotatebox[2]{#2}%
  \ifx\svgwidth\undefined%
    \setlength{\unitlength}{272bp}%
    \ifx\svgscale\undefined%
      \relax%
    \else%
      \setlength{\unitlength}{\unitlength * \real{\svgscale}}%
    \fi%
  \else%
    \setlength{\unitlength}{\svgwidth}%
  \fi%
  \global\let\svgwidth\undefined%
  \global\let\svgscale\undefined%
  \makeatother%
  \begin{picture}(1,0.99911167)%
    \put(0,0){\includegraphics[width=\unitlength]{troistrois.pdf}}%
    \put(0.27982653,0.75001505){\color[rgb]{0,0,0}\makebox(0,0)[lb]{\smash{$(1,1)$}}}%
    \put(0.221003,0.92648561){\color[rgb]{0,0,0}\makebox(0,0)[lb]{\smash{$v$}}}%
    \put(0.05923829,0.76472093){\color[rgb]{0,0,0}\makebox(0,0)[lb]{\smash{$u$}}}%
    \put(0.75041476,0.75001505){\color[rgb]{0,0,0}\makebox(0,0)[lb]{\smash{$(1,3)$}}}%
    \put(0.51512065,0.75001505){\color[rgb]{0,0,0}\makebox(0,0)[lb]{\smash{$(1,2)$}}}%
    \put(0.27982653,0.51472093){\color[rgb]{0,0,0}\makebox(0,0)[lb]{\smash{$(2,1)$}}}%
    \put(0.75041476,0.51472093){\color[rgb]{0,0,0}\makebox(0,0)[lb]{\smash{$(2,3)$}}}%
    \put(0.51512065,0.51472093){\color[rgb]{0,0,0}\makebox(0,0)[lb]{\smash{$(2,2)$}}}%
    \put(0.27982653,0.27942677){\color[rgb]{0,0,0}\makebox(0,0)[lb]{\smash{$(3,1)$}}}%
    \put(0.51512065,0.27942677){\color[rgb]{0,0,0}\makebox(0,0)[lb]{\smash{$(3,2)$}}}%
  \end{picture}%
\endgroup%

%% file: NewFigures/SchemaDisc/schemaDisc.pdf_tex
\begingroup%
  \makeatletter%
  \providecommand\color[2][]{%
    \errmessage{(Inkscape) Color is used for the text in Inkscape, but the package 'color.sty' is not loaded}%
    \renewcommand\color[2][]{}%
  }%
  \providecommand\transparent[1]{%
    \errmessage{(Inkscape) Transparency is used (non-zero) for the text in Inkscape, but the package 'transparent.sty' is not loaded}%
    \renewcommand\transparent[1]{}%
  }%
  \providecommand\rotatebox[2]{#2}%
  \ifx\svgwidth\undefined%
    \setlength{\unitlength}{337.6bp}%
    \ifx\svgscale\undefined%
      \relax%
    \else%
      \setlength{\unitlength}{\unitlength * \real{\svgscale}}%
    \fi%
  \else%
    \setlength{\unitlength}{\svgwidth}%
  \fi%
  \global\let\svgwidth\undefined%
  \global\let\svgscale\undefined%
  \makeatother%
  \begin{picture}(1,0.52369668)%
    \put(0,0){\includegraphics[width=\unitlength]{schemaDisc.pdf}}%
    \put(0.76066351,0.3921801){\color[rgb]{0,0,0}\makebox(0,0)[lb]{\smash{Ground truth}}}%
    \put(0.76066351,0.32109005){\color[rgb]{0,0,0}\makebox(0,0)[lb]{\smash{Least-squares}}}%
    \put(0.76066351,0.25){\color[rgb]{0,0,0}\makebox(0,0)[lb]{\smash{Sparsity}}}%
  \end{picture}%
\endgroup%